\documentclass[b5paper,10pt]{book}

\usepackage{phdthesis}
\usepackage[
  paperwidth=17cm
  ,paperheight=24cm
  ,inner=2.6cm
  ,outer=2.3cm
  ,top=2.3cm
  ,bottom=2.3cm
]{geometry}

\usepackage[hyphens]{url}
\usepackage{tabularx}
\usepackage{amsmath}
\usepackage{multirow}
\usepackage{booktabs}
\usepackage{amssymb}
\usepackage[square,sort,sectionbib,numbers]{natbib}
\usepackage{wrapfig}
\usepackage{enumerate}
\usepackage{supertabular}
\usepackage{multicol}
\usepackage{setspace}
\usepackage{import}
\usepackage{booktabs} 
\usepackage{graphicx}
\usepackage[hyphens]{url}
\usepackage{microtype}
\usepackage{tabularx,amsmath}
\usepackage{multirow}
\usepackage{color}
\usepackage[usenames,dvipsnames,table]{xcolor}
\usepackage[normalem]{ulem}
\usepackage[english]{babel}
\usepackage{statex}
\usepackage[noend]{algorithmic}
\usepackage{algorithm}
\usepackage{paralist}
\usepackage{flushend}
\usepackage{enumerate}
\usepackage[shortlabels,inline]{enumitem}
\usepackage{setspace}
\usepackage{textcomp}
\usepackage{listings}
\usepackage{keyval}
\usepackage{rotating}
\usepackage{acronym}
\usepackage{bibentry}
\nobibliography*

\usepackage{multirow}
\usepackage{balance}
\usepackage{subfigure}
\usepackage[np, autolanguage]{numprint}
\usepackage[skip=1pt]{caption}
\usepackage{amsthm}
\usepackage[official]{eurosym}
\usepackage{bbm}
\usepackage{pdflscape}
\usepackage{eurosym}
\usepackage{lmodern}
\usepackage{ragged2e}
\usepackage[backref=page]{hyperref}
\hypersetup{%
pdfborder = {0 0 0},
pdftitle = {Your Thesis Title Here},
pdfsubject = {PhD Thesis},
pdfkeywords = {add, some, keywords},
pdfauthor = {Your Name}
}
\usepackage{bookmark}

\renewcommand*{\backref}[1]{} 
\renewcommand*{\backrefalt}[4]{%
\ifcase #1
\or (Cited on page~#2.)  %
\else 
(Cited on pages~#2.)  
\fi
}

\let\svthefootnote\thefootnote
\newcommand\blankfootnote[1]{%
  \let\thefootnote\relax\footnotetext{#1}%
  \let\thefootnote\svthefootnote%
}
\let\svfootnote\footnote
\renewcommand\footnote[2][?]{%
  \if\relax#1\relax%
    \blankfootnote{#2}%
  \else%
    \if?#1\svfootnote{#2}\else\svfootnote[#1]{#2}\fi%
  \fi
}

\acrodef{IR}{information retrieval}
\acrodef{ML}{Machine Learning}
\acrodef{GDPR}{General Data Protection Regulation}
\acrodef{FT}{Feature Tweaking}
\acrodef{RP}{Random Perturbation}
\acrodef{FOCUS}{Flexible Optimizable CoUnterfactual Explanations for Tree EnsembleS}
\acrodef{FACT}{Fairness, Accountability, Confidentiality and Transparency}
\acrodef{FACT-AI}{Fairness, Accountability, Confidentiality and Transparency in Artificial Intelligence}
\acrodef{MLRC}{Machine Learning Reproducibility Challenge}
\acrodef{GNN}{Graph Neural Network}

\newcommand{\OurCompany}{the retailer}
\newcommand{\OurMethod}{MC-BRP}
\newcommand{\OurUniversity}{the University of Amsterdam}
\newcommand{\OurShort}{\textsc{CF-GNN}}
\newcommand{\synone}{\textsc{ba-shapes}}
\newcommand{\synfour}{\textsc{tree-cycles}}
\newcommand{\synfive}{\textsc{tree-grid}}
\newcommand{\gnnexplainer}{\textsc{GNNExplainer}}
\newcommand{\gnnexpshort}{\textsc{GNNExp}}

\newcommand{\cgraph}{subgraph neighborhood}
\newcommand{\baserand}{\textsc{random}}
\newcommand{\basekeep}{\textsc{1hop}}
\newcommand{\baserm}{\textsc{rm-1hop}}

\newcommand{\dubbelop}{$^{\blacktriangle}$}

\newcommand{\dubbelneer}{$^{\blacktriangledown}$}
\newcommand{\notsig}{$^{\circ}$}

\newcommand{\NoExample}{$^\otimes$}

\newcommand{\norm}[1]{\left\lVert#1\right\rVert}

\theoremstyle{definition}
\newtheorem{definition}{Definition}[section]

\theoremstyle{remark}

\DeclareMathOperator{\softmax}{softmax}

\newcommand{\graph}{\mathcal{G}}
\newcommand{\compgraph}{\mathcal{G}_v}
\newcommand{\compadj}{A_v}
\newcommand{\compdeg}{D_v}
\newcommand{\compnode}{X_v}

\newcommand{\loss}{\mathcal{L}}
\newcommand{\losspred}{\mathcal{L}_{pred}}

\newcommand{\lossdist}{\mathcal{L}_{dist}}
\newcommand{\perturbb}{P}
\newcommand{\perturbl}{\hat{P}}

\newlength\myindent
\newcommand\bindent{%
  \begingroup
  \setlength{\itemindent}{\myindent}
  \addtolength{\algorithmicindent}{\myindent}
}
\newcommand\eindent{\endgroup}

\counterwithin{algorithm}{chapter}

\newlength\longest

\begin{document}

% Start the front matter
% Latex takes care of Roman page numbering ( I, II, III, IV, etc) for front matter
% Front matter, acknowledgements, and ToC
\frontmatter
% !TEX root = thesis-main.tex
% This is the first page, consisting of your title and name.

% Title goes here
{\pagestyle{empty}
\newcommand{\printtitle}{%
{
\Huge\bf Explaining Predictions from Machine Learning Models: Algorithms, Users, and Pedagogy \\[0.8cm]
}}

% Some markup followed by your name
\begin{titlepage}
\par\vskip 2cm
\begin{center}
\printtitle
\vfill
{\LARGE\bf Ana Lu\v{c}i\'{c}}		% ACCENT 1/5
%{\LARGE\bf Ana Lucic}
\vskip 2cm
\end{center}
\end{titlepage}

% Skip a page to start on a right page again.
\mbox{}\newpage
\setcounter{page}{1}

% This is the title page (titelblad)
% You also need to send it to the Bureau Pedel
% Make sure you change:
%   the date of your defense
%   your name
%   your place of birth (including country, but ONLY if you were born outside the Netherlands)
\clearpage
\par\vskip 2cm
\begin{center}
\printtitle
\par\vspace {4cm}
{\large \sc Academisch Proefschrift}
\par\vspace {1cm}
{\large ter verkrijging van de graad van doctor aan de \\
Universiteit van Amsterdam\\
op gezag van de Rector Magnificus\\
prof.\ dr.\ G.T.M. ten Dam\\
ten overstaan van een door het College voor Promoties ingestelde \\
commissie, in het openbaar te verdedigen in \\
de Aula der Universiteit\\
op vrijdag 23 september 2022, te 14:00 uur \\ } %IT SHOULD REALLY BE 'te'
\par\vspace {1cm} {\large door}
\par \vspace {1cm}
{\Large Ana Lu\v{c}i\'{c}}		% ACCENT 2/5
%{\Large Ana Lucic}
\par\vspace {1cm}
{\large geboren te Para\'{c}in, Joegoslavi\"{e}} 	% ACCENT 3/5
%{\large geboren te Paracin} 	
%{\large geboren te Para\'{c}in} 
% IN THE PAST THIS HAS TO INCLUDE THE COUNTRY TOO, BUT NOT ANYMORE
%MAKE VERY SURE THIS MATCHES YOUR PASSPORT, THE BEADLE WILL CHECK THIS! 
%{\large geboren te Stad} %IF YOU ARE BORN IN THE NETHERLANDS, OMMIT THE COUNTRY
\end{center}

% The page following the titelblad. This usiallu contains:
%   committee members
%   SIKS logo + text
%   sponsors/ project numbers
%   ISBN
%   copyrights, cover design, printer
\clearpage
\noindent%
\textbf{Promotiecommissie} \\\\
\begin{tabular}{@{}l l l}
Promotor: \\
& Prof.\ dr.\ M.\ de Rijke & Universiteit van Amsterdam \\  %promotor
Co-promotor: \\
& Prof.\ dr.\ H.\ Haned & Universiteit van Amsterdam \\  %co-promotor
Overige leden: \\
& Prof.\ dr.\ F.\ Silvestri & Sapienza University of Rome \\ 
& Prof.\ dr.\ M.\ Lovri\'{c} & McMaster University \\ 
& Prof.\ dr.\ M.\ Welling & Universiteit van Amsterdam \\ 
& Prof.\ dr.\ C.\ S\'{a}nchez Guti\'{e}rrez & Universiteit van Amsterdam \\ 
& Dr.\ F.\ P.\ Santos & Universiteit van Amsterdam \\

\end{tabular}

\bigskip\noindent%
Faculteit der Natuurwetenschappen, Wiskunde en Informatica\\

\vfill

% Sponsors and projects
\noindent
This research was supported by the Netherlands Organisation for Scientific Research under project number\ 652.\-001.\-003, the Partnership on AI, and Ahold Delhaize. 
\bigskip

% Copyrights
\noindent
Copyright \copyright~2022 Ana Lu\v{c}i\'{c}, Amsterdam, The Netherlands\\		% ACCENT 4/5
%Copyright \copyright~2022 Ana Lucic, Amsterdam, The Netherlands\\
Cover by Off Page, Amsterdam\\
Printed by Off Page, Amsterdam\\
\\
ISBN: 978-94-93278-21-9\\

\clearpage

\thispagestyle{empty}
\null\vfill
\centering
\textit{The more I see, the less I know.} \\ \medskip
\qquad \qquad \qquad \qquad \qquad -- {Anthony Kiedis}
\vfill\vfill

\clearpage

}

% !TEX root = thesis-main.tex
{\pagestyle{empty}

{
\begin{center}

\noindent
\textbf{Acknowlegements} \\ \vskip .5cm
\end{center}
}

\noindent
Moving to a foreign country to do a PhD on a foreign topic was not something I had originally planned, but it has nonetheless turned out to be a really rewarding experience. 
This thesis is the culmination of that experience, and there are many people without whom it would not have been possible. 

First and foremost, I want to thank my supervisors who guided me through this whole process. 
Maarten -- thank you for always listening to me and advocating for me.
Hinda -- thank you for your unwavering support and for always being on my team. 

Next, I'd like to thank the committee members who reviewed my thesis. 
Miroslav -- thank you for motivating me to pursue a math major during my undergrad. This decision has fueled everything I've done since then and I'm not sure I would have been confident enough to do it without your encouragement.
Fernando -- thank you for taking over the FACT-AI course we created for the MSc AI program. I'm really looking forward to seeing how it develops over the next few years. 
Clarisa -- thank you for giving me advice about the challenges I encountered during my fellowship project. Your guidance helped me prioritize and carve out the next steps. 
Fabrizio -- thank you for letting me do my internship project unofficially after I was unable to do a formal internship due to the pandemic. This project is my favorite chapter in this thesis and it has inspired me to shift my research into a new direction. 
Max -- thank you for allowing me to pursue this new direction under your guidance.

I want to thank my paranymphs for standing with me as I defend my thesis. 
Maartje and Maurits -- thank you for being my main sources of support within our group and for sharing all of the ups and downs of doing a PhD with me. I also want to thank the many great people I've met while working at the University of Amsterdam:
Alexey, 
Ali, 
Ali, 
Amir, 
Antonis, 
Arezoo, 
Artem, 
Boris, 
Chang, 
Christof, 
Chuan,
Dan,  
David, 
Gabriel, 
Hamid, 
Ilya, 
Jiahuan, 
Jie, 
Jin, 
Julia, 
Julien, 
Katya, 
Ke, 
Maria, 
Marlies, 
Mohammad, 
Mostafa, 
Mozdheh, 
Olivier, 
Pengjie, 
Petra, 
Sam, 
Sami, 
Shaojie,
Spyretta, 
Svitlana, 
Thomas, 
Trond, 
Vera, 
Yangjun, and 
Yifan -- thank you for making Science Park (and especially our mint green container) a fun place to work.  
In particular, I want to thank 
Harrie,
Hosein, 
Nikos, 
Rolf, 
Christophe, 
Bob, and 
Tom for your leadership in the group, and of course for all the borrels, biertjes, and bitterballen. 
Marzieh, 
Mariya, and
Mahsa
 -- thank you for your honesty, friendship, and support over these last few years. 
Ziming -- thank you for letting me babysit your cat for a summer, it was truly one of the highlights of my PhD. 
I also want to thank the many bright students I had: Michael, Stefan, Kim, Puja, Fije -- thank you for teaching me way more than I taught you.

During my PhD, I had the opportunity to do a fellowship at the Partnership on AI. Rebecca -- thank you for leading me through my time at the Partnership and for your trust in me amid all the twists and turns involved in doing real-world research. 
\pagebreak

I've had three lovely roommates while living in Amsterdam, each of which were living with me at very different phases of my PhD. 
Elisabet -- thank you for helping me through the challenges and uncertainties of early PhD life and for showing me every single specialty coffee shop in Amsterdam. 
Maria -- thank you for quarantining with me and supporting me through hairdresser fiascos and global pandemics (not sure what's worse!). Luckily we can always fall back on falafel salads and eggless desserts. 
Camila -- thank you for supporting my leopard print addiction and for always being up for something fun. 
Special shoutout to my Kelowna roommates Danielle and Bree, who only lived with me briefly during my PhD, but have nonetheless contributed to many ``good vibes'' and sometimes questionable memories. \\

I've learned that having a support system outside the research world is incredibly important while pursuing a PhD. 
Sophie, Tamara, Juliette, Jago, and Thomas -- thank you for adopting me at Mystic Garden all those years ago and for the many parties, picnics, pizzas, and pals we've shared since then.
Duffey, Allie, Kristine, Craig, Devra, Jeff, and Aly -- thank you for not forgetting about me even though I only come home once a year. 
Thomas -- thank you for showing me how to be a kinder, softer person and for always listening to the many things I have to say (ook in het Nederlands). You are what I look forward to. 
Bedankt aan de hele familie Jak om me welkom te heten met open armen en open harten. \\

Naravno se moram zahvaliti na\v{s}om rajom. 
Hvala mojim Para\'{c}incima: Deda Uj\v{c}e, Baka Ujna, Ana, Sale, Sanja, Mi\v{s}o, Margarita, Marija i Nikola, i mojim drugim roditeljim iz Kanade: Ksenija, Mi\v{s}o, Srdjan, Ksenija, Stanko, \v{Z}eki I Gogo. Vi ste mi uvijek u srcu.
Najva\v{z}nije, ho\'{c}u da se zahvalim svojim roditeljima, Vesna i Tiho, koji su meni sve mogu\'{c}e u\v{c}inili. Hvala mojoj Baki koja me je ``dva puta rodila'' i cuvala od malenog. Ja vas obozavam, i ovo je za vas. \\ \vskip .3cm

\RaggedLeft{
Ana Lu\v{c}i\'{c} \\		% ACCENT 5/5
%Ana Lucic \\
Amsterdam \\
July 2022}

\clearpage
}			%Add back in once finished
\renewcommand\contentsname{Table of Contents}
{%avoid all upper case headers
\let\MakeUppercase=\relax
\tableofcontents
}

% Start the main matter
% Here, Latex will use Arabic page numbering (1, 2, 3, etc.)
% Add all your chapters:
%  1. introduction
%  2. background/ related work
%  3. experimental methodology
%  4,5,X. technical chapters
%  X+1. conclusion
\mainmatter

% !TEX root = ../thesis-main.tex

\chapter{Introduction}
\label{chapter:introduction}

Machine learning (ML) is the study of algorithms that learn models directly from data \citep{murphy_2012_ml}. 
Such algorithms are typically self-improving -- their parameters are updated iteratively based on the data they receive, thereby \emph{learning} a model that is representative of the data. 
Once an ML model is trained, it is usually evaluated on unseen data in order to test its generalization capabilities. 
The ability to generalize to new situations is one of the most important aspects of ML models, and is perhaps the reason such models are often referred to as ``intelligent'' \citep{thornhill_2021_ai}.
In other words, ML models use information from the past (i.e., historical data) to make predictions about the future (i.e., unseen data).

The field of ML has enjoyed great success in the last decade primarily due to the increased availability of data and computational resources \citep{Amodei2018AI-and-Compute}. 
As ML models have become more prominent in decision-making scenarios \citep{aiindex_2022_stanford}, there has been an increased demand for ensuring such models are
\begin{inparaenum}[(i)]
	\item fair, 
	\item accountable, 
	\item confidential, and 
	\item transparent \citep{olteanu_2021_factsir}.\footnote{Although confidentiality and transparency may seem like contradictory objectives, confidentiality typically refers to preserving the privacy of individuals within a training dataset, while transparency refers to the ML model and the process that went into deploying it.}
%it is possible (and preferable) for ML models to be transparent w.r.t. to the algorithms involved while also confidential w.r.t. preserving the privacy of individuals in a training dataset. 
\end{inparaenum}
%This is in part due to the many examples of problematic unintended consequences of ML systems that have been uncovered in recent years, such as discrimination in policing \citep{propublica} and hiring \citep{dastin_2018_amazon} as a result of algorithmic decision making. 
However, ML models can be difficult to interpret due to their complex architectures and the large numbers of parameters involved, effectively deeming them ``black-boxes'' \citep{castelvecchi_2016_blackbox}. 
In this thesis, we primarily focus on developing methods to increase \emph{transparency}, which we define as mechanisms that provide insight into an ML model. 
This knowledge is typically presented to a user in the form of an \emph{explanation}. 

Recently, the artificial intelligence (AI) research community has embarked on the development of explainable artificial intelligence (XAI): a relatively new subfield of AI where the aim is to explain predictions from complex ML models \citep{guidotti-2018-survey}. 
Explanations can be used to make ML models more accountable to various stakeholders involved in the pipeline by providing insight into not only how the model arrived at its decision, but also how to change or contest the decision, if necessary \citep{wachter_counterfactual_2017}. 
We distinguish between two main types of explanations:
\begin{itemize}[leftmargin=*]
\item \textbf{Behavior-based explanations}: provide insight into how an ML model makes predictions from an algorithmic or mathematical perspective. For example, ranking the most important features \citep{ribeiro-2016-should,lundberg_unified_2017}, identifying influential  \citep{koh-2017-understanding,sharchilev-2018-finding} or prototypical \citep{li_deep_2017,tan_tree_2016} training samples, or generating counterfactual perturbations \citep{wachter_counterfactual_2017,stepin2021survey,verma2020counterfactual}. 
%Most existing XAI methods produce behavior-based explanations.
Behavior-based explanations are important for understanding the internal processes of ML models. 

\pagebreak
\item \textbf{Process-based explanations:} provide insight into the ML modeling pipeline. For example, detailing how the data were collected and preprocessed \citep{gebru_datasheets_2020}, or reporting on how the model was trained and evaluated \citep{mitchell_model_2019}. Process-based explanations are important for ensuring that ML research is conducted in a responsible and reproducible manner. 

\end{itemize}

%Next we define \emph{explainability} as translating transparency insights into something that is understandable by a human \todo{\citep{lucic2021multistakeholder}?}.
%This allows us to disentangle the algorithmic component of generating model insights (i.e., the \emph{interpretation} from the form in which the information is presented to the user (i.e., the explanation). 

\noindent%
This thesis has three parts: the first focuses on algorithms, the second focuses on users, and the third focuses on pedagogy. 
In the first two parts of this thesis, we develop methods for generating behavior-based explanations, which is what the majority of existing XAI methods produce.
\citet{guidotti-2018-survey} develop a taxonomy for classifying XAI methods using four main criteria, we slightly adapt their taxonomy as follows:\footnote{We refer to explanations as ``global'' or ``local'' since this aligns more closely with standard terms recently in the literature. We also introduce the distinction between ``model-specific'' and ``model-agnostic''. }
\begin{itemize}[leftmargin=*]
\item{\textbf{Problem:} the type of explanation we want to generate.}
 \begin{enumerate}[(i)]
% \item{\emph{Model explanations:} interpret black-box model as a whole (globally)}
% \item{\emph{Outcome explanations:} interpret individual black-box predictions (locally)}
 \item{\emph{Global explanations:} interpret ML model behavior in general, i.e., how it makes predictions across data points.}
 \item{\emph{Local explanations:} interpret individual ML model predictions.}
 \item{\emph{Inspection:} interpret model behavior through visual representations (globally or locally).}
 \item{\emph{Transparent design:} model is inherently interpretable (globally or locally).}
 \end{enumerate}
 
\item{\textbf{Model:} the dependence on model class.}
 \begin{enumerate}[(i)]
 \item{\emph{Model-specific:} requires full access to the model's inner workings, where the model can be a neural network, tree ensemble, support vector machine, etc.}
 \item{\emph{Model-agnostic:} treats model as a ``black-box'' and is therefore not dependent on its inner workings.}
 \end{enumerate}

\item{\textbf{Explanator:} the mechanism used to generate explanations, e.g., decision rules, feature attributions, sensitivity analysis, prototype selection, etc.}

\item{\textbf{Data:} the type of data being explained, e.g., tabular, image, text, graph, etc.}
\end{itemize}
 
\noindent%
In Chapters~\ref{chapter:research-focus} and \ref{chapter:research-cfgnn}, we develop and investigate local, model-specific explanation methods.
We focus on local explanations methods because they are a natural precursor to global explanations \citep{vanderlinden_2019_global}. 
%We focus on model-specific explanations because we believe that in order to truly understand what a model is doing, we need to have access to its inner workings.
We focus on model-specific explanations because in many practical scenarios, we have full access to the ML model and can therefore make use of its inner workings when generating explanations. 
The algorithm proposed in Chapter~\ref{chapter:research-focus} is specific to tree ensembles and operates on tabular data, while the algorithm proposed in Chapter~\ref{chapter:research-cfgnn} is specific to graph neural networks (GNNs) and operates on graph data. 

In general, we take the position that treating the model as a ``black-box'' in order to interpret its predictions is a somewhat contradictory statement;
%If we want to understand what is going on inside the ``black-box'', surely we have to open it? 
how can we understand what a model is doing if do not have access to its internal processes?
However, there exist use cases where we need explanations but we do not have full access to the model, such as auditing \citep{raji_actionable_2019}. 
To accommodate such scenarios, we propose a method in Chapter \ref{chapter:research-mcbrp} that is model-agnostic in principle, but we evaluate it in a model-specific manner. 

%Although there are use cases where model-agnostic explanations are more appropriate (i.e., auditing \citep{raji_actionable_2019}), we take the position that treating the model as a ``black-box'' in order to interpret its predictions is a somewhat contradictory statement. 
%The algorithms proposed in Chapters~\ref{chapter:research-focus} and~\ref{chapter:research-mcbrp} are specific to tree ensembles and operate on tabular data, while the algorithm proposed in Chapter~\ref{chapter:research-cfgnn} is specific to graph neural networks (GNNs) and operates on graph data. 

In the third and final part of the thesis, we shift from investigating behavior-based explanations for explaining individual predictions to investigating process-based explanations for explaining the ML modeling pipeline. We do so from a pedagogical point of view. 
We use reproducibility as a mechanism for teaching about responsible AI concepts: fairness, accountability, confidentiality, and transparency in a graduate-level course at the University of Amsterdam. 
In Chapter~\ref{chapter:research-pedagogy}, we report on our experiences and lessons learned after teaching the course over two academic years.

%% The research questions and sub questions
% !TEX root = ../thesis-main.tex

\section{Research Outline and Questions}
\label{section:introduction:rqs}

\acrodef{rq:focus}[\ref{rq:focus}]{Can we generate counterfactual explanations for tree-based models using gradient-based optimization?}
% Short answer: using gradient-based optimization + differentiable approximations of trees

\acrodef{rq:cf-gnn}[\ref{rq:cf-gnn}]{Can we extend our counterfactual explanation method for tree-based models to graph-based models?}
% Short answer: first extend problem formalization to graphs, then apply the same gradient-based optimization technique as in FOCUS

\acrodef{rq:mcbrp}[\ref{rq:mcbrp}]{Given a real-world use case, can we create an explanation method based on this use case and evaluate it in a context-specific manner?}
% Short answer: Yes, we took a use case from Ahold and developed an explanation method around those users' needs. We evaluate the method with a user study that has objective + subjective components.

\acrodef{rq:pedagogy}[\ref{rq:pedagogy}]{How can we teach about responsible AI topics to a technical, research-oriented audience?} 
% Short answer: We create a course centered around a project that involves reproducing FACT-AI algorithms.

This thesis focuses on explaining for ML models in three different contexts: algorithms (Chapters~\ref{chapter:research-focus} and~\ref{chapter:research-cfgnn}), users (Chapter~\ref{chapter:research-mcbrp}), and pedagogy (Chapter~\ref{chapter:research-pedagogy}). 
Below, we describe the main research questions for each chapter. 

\subsection{Algorithms}
In the first part of the thesis, we focus on the algorithmic component of generating explanations for ML predictions. 
Based on existing work from the philosophy, cognitive science and social psychology disciplines, \citet{miller-2017-explanations} identifies various aspects of explainability that AI researchers should pay attention to when generating explanations for ML models. 
One of the main recommendations is that explanations should be \emph{contrastive}: they should allow the user to compare and contrast between the original instance (i.e., data point) and a counterfactual case.\footnote{There has recently been discussion on the differences between \emph{contrastive} and \emph{counterfactual} explanations \citep{stepin2021survey,karimi2020survey}. For the purposes of this thesis, we use these terms interchangeably.} 
\emph{Counterfactual explanations} are defined as the minimal perturbations to the input data that such that the prediction changes \citep{wachter_counterfactual_2017}. 
Counterfactual explanations have been identified as explanations that can ``\textit{provide information to the data subject that is both easily digestible and practically useful for understanding the reasons for a decision, challenging them, and altering future behavior for a better result}'' \citep{wachter_counterfactual_2017}.

In the first chapter in this part of the thesis, we develop and evaluate a method for generating counterfactual explanations specific to tree ensembles. 
We focus on tree ensembles because the majority of existing research on counterfactual explanations is either 
\begin{inparaenum}[(i)]
	\item model-agnostic or
	\item specific to deep learning \citep{karimi2020survey}, 
\end{inparaenum}
even though tree ensembles perform well on tabular data and are widely used in industrial settings \citep{shwartz-ziv2021tabular}. 
In this chapter, we aim to answer the following research question:
\begin{enumerate}[label=\textbf{RQ\arabic*},ref={RQ\arabic*},resume,leftmargin=*]
	\item \acl{rq:focus}\label{rq:focus}
\end{enumerate}

\noindent%
To answer \textbf{RQ1}, we propose a method that introduces differentiable approximations into the optimization framework, which allows us to use standard gradient-based optimization techniques to generate counterfactual explanations. 
We find that, unlike existing approaches, our method 
\begin{inparaenum}[(i)]
	\item produces counterfactual examples for all instances in a dataset, 
	\item produces counterfactual examples that are closer to the original instances compared to existing approaches, and
	\item can handle larger model sizes.
\end{inparaenum}

Although there exist many methods for generating counterfactual explanations for tabular, text, and image data (see surveys by \citet{verma2020counterfactual}, \citet{karimi_recourse_2020}, and \citet{stepin2021survey}), there are relatively few for graph data.\footnote{At the time of writing \citep{lucic2021cfgnnexplainer}, there were no existing methods for generating counterfactual explanations for graph data.}
In the second chapter in this part of the thesis, we investigate the following research question:

\begin{enumerate}[label=\textbf{RQ\arabic*},ref={RQ\arabic*},resume,leftmargin=*]
	\item \acl{rq:cf-gnn}\label{rq:cf-gnn}
\end{enumerate}
To answer \textbf{RQ2}, we formalize the problem of counterfactual explanations for graph neural networks (GNNs) and propose a method for generating them by iteratively removing entries in the adjacency matrix, which corresponds to removing edges in the graph. 
We also propose an experimental setup for evaluating counterfactual explanations for GNNs and find that our explanation method is able to generate counterfactual explanations that are more minimal and more accurate in comparison to the baselines.

\subsection{Users}
Given that explanations are created in order to be consumed by users, it is a natural next step to think about incorporating a more user-centric perspective into the development of XAI methods \citep{ehsan_2021_expanding}. 
In this part of the thesis, we investigate generating explanations that are situated and evaluated within a particular deployment context. 
We ask the following research question: 

\begin{enumerate}[label=\textbf{RQ\arabic*},ref={RQ\arabic*},resume,leftmargin=*]
	\item \acl{rq:mcbrp}\label{rq:mcbrp}
\end{enumerate}

\noindent
To answer \textbf{RQ3}, we first identify a use case where users want to see explanations for ML models: explaining errors in forecasting predictions. 
We propose a method for generating explanations based on the specifics of this use case and design a user study to evaluate how interpretable and actionable our explanation are. 
We also test users' subjective attitudes towards the explanations generated by our method.

\subsection{Pedagogy}
In the final part of this thesis, we transition from explaining ML predictions to explaining best practices for conducting ML research. 
We ask the following research question:

\begin{enumerate}[label=\textbf{RQ\arabic*},ref={RQ\arabic*},resume,leftmargin=*]
	\item \acl{rq:pedagogy}\label{rq:pedagogy}
\end{enumerate}
\noindent
To answer \textbf{RQ4}, we design a graduate-level course about fairness, accountability, confidentiality, and transparency in AI. 
We describe how we structured the course around a reproducibility project and report in detail on the insights gained from teaching the course over two academic years. 
We emphasize that conducting research in a responsible and reproducible manner is not only important for individual ML researchers, but is also essential for scientific progress in general.

%% Lists the main contributions of the thesis
% !TEX root = ../thesis-main.tex

\section{Main Contributions}
\label{section:introduction:contributions}

In this section, we summarize the main contributions of this thesis.

\subsection*{Theoretical contributions}
\begin{enumerate}
	\item A formalization of the counterfactual explanation problem for GNNs (Chapter~\ref{chapter:research-cfgnn}). 
	\item An experimental setup for evaluating counterfactual explanations for GNNs (Chapter~\ref{chapter:research-cfgnn}). 
	\item A user study framework for evaluating the effectiveness of contrastive explanations (Chapter~\ref{chapter:research-mcbrp}). 
	\item An analysis on the difference in attitudes towards explanations between different types of stakeholders (Chapter~\ref{chapter:research-mcbrp}). 
\end{enumerate}

\subsection*{Algorithmic contributions}
\begin{enumerate}[resume]
	\item Flexible Optimizable CoUnterfactual Explanations for Tree EnsembleS (FOCUS): an algorithm for generating counterfactual explanations for tree ensembles (Chapter~\ref{chapter:research-focus}). 
	\item CF-GNNExplainer: an algorithm for generating counterfactual explanations for GNNs (Chapter~\ref{chapter:research-cfgnn}). 
	\item Monte Carlo Bounds for Reasonable Predictions (MC-BRP): an algorithm for generating explanations about errors in forecasting predictions (Chapter~\ref{chapter:research-mcbrp}). 
\end{enumerate}

\subsection*{Pedagogical contributions}
\begin{enumerate}[resume]
	\item A teaching setup for a course about responsible AI with a focus on reproducibility (Chapter~\ref{chapter:research-pedagogy}), including a set of guidelines for implementing similar courses in the future based on our insights. 
\end{enumerate}

%% Overview of the thesis; what is described in which chapter
% !TEX root = ../thesis-main.tex

\section{Thesis Overview}
\label{section:introduction:overview}

This thesis is organized into three parts, each part can be read independently. 

The first part focuses on proposing new algorithms for explaining predictions from ML models. 
Specifically, we propose methods for generating counterfactual explanations for tree-based models (Chapter~\ref{chapter:research-focus}), and for graph-based models (Chapter~\ref{chapter:research-cfgnn}). 
These methods can be applied on any tree- or graph-based model, respectively.

The second part focuses on the interaction between ML explanations and the users who consume them. 
We propose a method for explaining errors in forecasting predictions (Chapter~\ref{chapter:research-mcbrp}). 
To evaluate our method, we propose a user study with both objective and subjective components, where we contrast and compare the results between two types of users: researchers and practitioners.

In the third part of the thesis, we shift our focus from translating knowledge about individual predictions to transferring knowledge to the next generation of researchers. 
We propose a course setup for teaching about responsible AI topics to a graduate-level audience and reflect on our learnings from past implementations of the course at the University of Amsterdam (Chapter~\ref{chapter:research-pedagogy}).

%% Describes the papers from which the chapters originate
% !TEX root = ../thesis-main.tex

\section{Origins}
\label{section:introduction:origins}

Below we list the publications that are the origins of each chapter.

\begin{enumerate}[label=\textbf{Chapter~\arabic*},align=left]
\setcounter{enumi}{0}

\item is based in part on the following paper:
\begin{itemize}
\item \bibentry{lucic2021multistakeholder}.
\end{itemize}
AL designed the framework during a research fellowship at the Partnership on AI, based on discussions with all authors. All authors contributed to providing feedback, AL did most of the writing. 

\item is based on the following paper:
\begin{itemize}
\item \bibentry{lucic2020focus}.
\end{itemize}
AL and HO designed the method. AL ran the experiments. All authors contributed to the writing, AL did most of the writing.

\item is based on the following paper:
\begin{itemize}
\item \bibentry{lucic2021cfgnnexplainer}.
\end{itemize}
AL designed the method and ran the experiments. All authors contributed to the writing, AL did most of the writing.

\item is based on the following paper:
\begin{itemize}
\item \bibentry{lucic_2020_why}.
\end{itemize}
AL designed the method and ran the experiments. All authors contributed to the writing, AL did most of the writing.

\item is based on the following paper:
\begin{itemize}
\item \bibentry{lucic2022reproducibility}.
\end{itemize}
AL and MB designed the course and implemented it together with MdR. All authors contributed to the writing, AL did most of the writing.

\end{enumerate}

%\todo{\noindent%
%We note that parts of Chapter~\ref{chapter:introduction} are based on the following paper:
%\begin{itemize}
%\end{itemize}}

\noindent%
The writing of this thesis also benefited from work on the following publications:

\begin{itemize}
\item \bibentry{lucic2019boosting}. 
\item \bibentry{lucic_2022_sigir}.
\item \bibentry{lucic_2022_acl}.
\item \bibentry{lucic_2022_pai}.

\item \bibentry{debie2021_trust}.
\item \bibentry{neely2022song}. 
\item \bibentry{neely2021order}. 
\item \bibentry{karunagaran_2022_toolsheets}.

\end{itemize}
%\input{02-background/main}
%\input{03-methodology/main}

%\if0
\part{Algorithms}
% !TEX root = ../thesis-main.tex

\chapter{Counterfactual Explanations for Tree Ensembles}
\label{chapter:research-focus}

\footnote[]{This chapter was published at the AAAI Conference on Artificial Intelligence (AAAI 2022) under the title ``FOCUS: Flexible Optimizable Counterfactual Explanations for Tree Ensembles'' \citep{lucic2020focus}.}
\acresetall

In the first part of this thesis, we explore creating algorithms for explaining predictions from various types of machine learning (ML) models. 
In this chapter, we address the following research question:

\medskip
\noindent
\textbf{\ref{rq:focus}:} \acl{rq:focus}
\medskip

\noindent
Existing methods for generating counterfactual explanations for tree-based models are either based on heuristics \citep{tolomei_interpretable_2017} or on integer linear programming techniques \citep{kanamori_dace_2020}. 
The former do not necessarily converge to an optimal solution, while the latter can be extremely computationally intensive.

The answer to \textbf{\ref{rq:focus}} is yes: we can achieve this by generating probabilistic approximations of tree-based models, which are differentiable and can therefore be used within a standard gradient-based optimization framework. 
Our experimental results show that our proposed algorithm can generate minimal counterfactual explanations in a more efficient and reliable manner in comparison to the baselines. 

%!TEX root = ../main.tex

\section{Introduction}
\label{section:focus-intro}
As ML models are prominently applied and their outcomes have a substantial effect on the general population, there is an increased demand for understanding what contributes to their predictions \citep{doshi-2017-towards}. 
For an individual who is affected by the predictions of these models, it would be useful to have an \emph{actionable} explanation -- one that provides insight into how these decisions can be \emph{changed}. 
The General Data Protection Regulation (GDPR) is an example of recently enforced regulation in Europe which gives an individual the right to an explanation for algorithmic decisions, making the interpretability problem a crucial one for organizations that wish to adopt more data-driven decision-making processes \citep{gdpr}. 

Counterfactual explanations are a natural solution to this problem since they frame the explanation in terms of what input (feature) changes are required to change the output (prediction). 
For instance, a user may be denied a loan based on the prediction of an ML model used by their bank. 
A counterfactual explanation could be: ``\textit{Had your income been \euro  $1000$ higher, you would have been approved for the loan}.''
We focus on finding \emph{optimal} counterfactual explanations: the \emph{minimal} changes to the input required to change the outcome. 

Counterfactual explanations are based on \emph{counterfactual examples}: generated instances that are close to an existing instance but have an alternative prediction. 
The difference between the original instance and the counterfactual example is the counterfactual explanation. 
\citet{wachter_counterfactual_2017} propose framing the problem as an optimization task, but their work assumes that the underlying machine learning models are differentiable, which excludes an important class of widely applied and highly effective non-differentiable models: tree ensembles. 
We propose a method that relaxes this assumption and builds upon the work of \citeauthor{wachter_counterfactual_2017} by introducing differentiable approximations of tree ensembles that can be used in such an optimization framework. 
Alternative non-optimization approaches for generating counterfactual explanations for tree ensembles involve an extensive search over many possible paths in the ensemble that could lead to an alternative prediction \citep{tolomei_interpretable_2017}. 

Given a trained tree-based model $f$, we probabilistically approximate $f$ by replacing each split in each tree with a sigmoid function centred at the splitting threshold. If $f$ is an ensemble of trees, then we also replace the maximum operator with a softmax. 
This approximation allows us to generate a counterfactual example $\bar{x}$ for an instance $x$ based on the minimal perturbation of $x$ such that the prediction changes: $y_{x} \neq y_{\bar{x}}$, where $y_{x}$ and $y_{\bar{x}}$ are the labels $f$ assigns to $x$ and $\bar{x}$, respectively. 
This leads us to our main research question in this chapter:
\begin{quote}
\emph{Are counterfactual examples generated by our method closer to the original input instances than those generated by existing heuristic methods?}
\end{quote}
Our main findings are that our proposed method is
\begin{enumerate*}[label=(\roman*)]
\item a more \emph{effective} counterfactual explanation method for tree ensembles than previous approaches since it manages to produce counterfactual examples that are closer to the original input instances than existing approaches; 
\item a more \emph{efficient} counterfactual explanation method for tree ensembles since it is able to handle larger models than existing approaches; and
\item a more \emph{reliable} counterfactual explanation method for tree ensembles since it is able to generate counterfactual explanations for all instances in a dataset, unlike existing approaches specific to tree ensembles. 
\end{enumerate*}

In the following sections, we examine existing work related to ours (Section~\ref{section:focus-related-work}) and formalize the counterfactual explanation problem (Section~\ref{section:focus-problem-definition}). 
We then describe the details of our method, Flexible Optimizable CoUnterfactual Explanations for Tree EnsembleS (FOCUS), in Section~\ref{section:focus-method}. 
In Section~\ref{section:focus-exp-setup}, we explain the experimental setup, followed by the experimental results in Sections~\ref{section:experiment1} and~\ref{section:experiment2}. 
We analyze our findings in Section~\ref{section:case-study} and conclude in Section~\ref{section:focus-conclusion}. 
%!TEX root = ../main.tex

\section{Related Work}
\label{section:focus-related-work}
Based on the taxonomy described in Chapter~\ref{chapter:introduction}, our setting in this chapter is a \emph{local explanation} problem for \emph{tree ensembles}. 
We use \emph{sensitivity analysis}, specifically counterfactual perturbations, on \emph{tabular} data to generate our explanations. 
Our work is related to counterfactual explanations in general (Section~\ref{section:focus-cf}), algorithmic recourse (Section~\ref{section:focus-recourse}), adversarial examples (Section~\ref{section:focus-adversarial}), and differentiable tree-based models (Section~\ref{section:focus-diff-trees}).

\subsection{Counterfactual Explanations}
\label{section:focus-cf}
Counterfactual examples have been used in a variety of ML areas, such as reinforcement learning \citep{madumal_explainable_2019}, deep learning \citep{alaa_deep_2017}, and XAI. 
Previous XAI methods for generating counterfactual examples are either model-agnostic \citep{poyiadzi_face_2020, karimi_model-agnostic_2019, laugel_inverse_2017, van_looveren_interpretable_2020,  mothilal_explaining_2020} or model-specific \citep{wachter_counterfactual_2017, grath_interpretable_2018, tolomei_interpretable_2017, kanamori_dace_2020, russell_efficient_2019, dhurandhar_explanations_2018}. 
Model-agnostic approaches treat the original model as a ``black-box'' and only assume query access to the model, whereas model-specific approaches typically do not make this assumption and can therefore make use of its inner workings (see Chapter~\ref{chapter:introduction}). 

Our work is a model-specific approach for generating counterfactual examples through optimization. 
Previous model-specific work for generating counterfactual examples through optimization has solely been conducted on differentiable models \citep{wachter_counterfactual_2017, grath_interpretable_2018, dhurandhar_explanations_2018}. 

\subsection{Algorithmic Recourse}
\label{section:focus-recourse}
Algorithmic recourse is a line of research that is closely related to counterfactual explanations, except that methods for algorithmic recourse include the additional restriction that the resulting explanation must be \emph{actionable} \citep{ustun_actionable_2019, joshi_towards_2019, karimi_recourse_2020, karimi_imperfect_causal_2020}. 
This is done by selecting a subset of the features to which perturbations can be applied in order to avoid explanations that suggest impossible or unrealistic changes to the feature values (i.e., change \textit{age} from \numprint{50} $\to$ \numprint{25}). 
Although this work has produced impressive theoretical results, it is unclear how realistic they are in practice, especially for complex ML models such as tree ensembles. 
Existing algorithmic recourse methods cannot solve our task because they 
\begin{enumerate*}[label=(\roman*)]
	\item are either restricted to solely linear \citep{ustun_actionable_2019} or  differentiable \citep{joshi_towards_2019} models, or
	\item  require access to causal information \citep{karimi_recourse_2020, karimi_imperfect_causal_2020}, which is rarely available in real world settings. 
\end{enumerate*}

\subsection{Adversarial Examples}
\label{section:focus-adversarial}
Adversarial examples are a type of counterfactual example with the additional constraint that the minimal perturbation results in an alternative prediction that is \emph{incorrect}. 
There are a variety of methods for generating adversarial examples \citep{goodfellow_explaining_2015,szegedy_intriguing_2014,su_one_2019,brown_adversarial_2018}; a more complete overview can be found in the work of \cite{biggio_wild_2018}. 
The main difference between adversarial examples and counterfactual examples is in the intent: adversarial examples are meant to \emph{fool} the model, whereas counterfactual examples are meant to \emph{explain} the model.

\subsection{Differentiable Tree-based Models}
\label{section:focus-diff-trees}
Part of our contribution involves constructing differentiable versions of tree ensembles by replacing each splitting threshold with a sigmoid function. 
This can be seen as using a (small) neural network to obtain a smooth approximation of each tree. 
Neural decision trees \citep{balestriero_neural_2017, yang_deep_2018} are also differentiable versions of trees, which use a full neural network instead of a simple sigmoid. 
However, these do not optimize for approximating an already trained model. Therefore, unlike our method, they are not an obvious choice for finding counterfactual examples for an existing model. 
Soft decision trees~\citep{hinton_distilling_2014} are another example of differentiable trees, which instead approximate a neural network with a decision tree. 
This can be seen as the inverse of our task.

%!TEX root = ../main.tex

\section{Problem Formulation}
\label{section:focus-problem-definition}

A \emph{counterfactual explanation} for an instance $x$ and a model $f$, $\Delta_{x}$, is a minimal perturbation of $x$ that changes the prediction of $f$. 
$f$ is a probabilistic classifier, where $f(y\mid x)$ is the probability of $x$ belonging to class $y$ according to $f$.
The prediction of $f$ for $x$ is the most probable class label $y_x = \arg\max_{y} f(y \mid x)$, and
a perturbation $\bar{x}$ is a counterfactual example for $x$ if, and only if, $y_x \not = y_{\bar{x}}$, that is:
\begin{align}
\arg\max_{y} f(y \mid x)
\not =
\arg\max_{y'} f(y' \mid \bar{x}).
\label{eq:cfexample}
\end{align}
In addition to changing the prediction, the distance between $x$ and $\bar{x}$ should also be minimized. 
We therefore define an \emph{optimal counterfactual example} $\bar{x}^*$ as: 
\begin{equation}
 \bar{x}^* := \arg\min_{\bar{x}} d(x, \bar{x}) 
 \text{ such that }
y_x \not = y_{\bar{x}},
\label{eq:optimalcondition}
\end{equation}
\noindent
where $d(x, \bar{x})$ is a differentiable distance function. 
The corresponding \emph{optimal counterfactual explanation} $\Delta^*_{x}$ is:
\begin{align}
\Delta^*_{x} = \bar{x}^* - x.
\end{align} 
%\begin{align}
%\begin{split}
% \bar{x}^*&{} := \arg\min_{\bar{x}} d(x, \bar{x})  \\
%& \text{such that }
%\arg\max_{y} f(y \mid x)
%\not =
%\arg\max_{y} f(y \mid \bar{x}).
%\end{split}
%\label{eq:optimalcondition}
%\end{align}
%\todo{It would be nice to better place this in the field, i.e. cite people who agree/disagree with this definition.}
This definition aligns with previous ML work on counterfactual explanations \citep{laugel_inverse_2017, karimi_model-agnostic_2019, tolomei_interpretable_2017}. 
We note that this notion of \emph{optimality} is purely from an algorithmic perspective and does not necessarily translate to optimal changes in the real world, since the latter are dependent on the context in which they are applied. 
It should be noted that if the loss space is non-convex, it is possible that more than one optimal counterfactual explanation exists.

Minimizing the distance between $x$ and $\bar{x}$ should ensure that $\bar{x}$ is as close to the decision boundary as possible. 
This distance indicates the effort it takes to apply the perturbation in practice, and an optimal counterfactual explanation shows how a prediction can be changed with the least amount of effort.
An optimal explanation provides the user with interpretable and potentially actionable feedback related to understanding the predictions of model $f$.

\citet{wachter_counterfactual_2017} recognized that counterfactual examples can be found through gradient descent if the task is cast as an optimization problem.
Specifically, they use a loss consisting of two components: 
\begin{enumerate*}[label=(\roman*)]
%	\item a prediction loss to change the prediction of $f$: $\mathcal{L}_{pred}(y_x, \bar{x} \mid f)$, and
	\item a prediction loss to change the prediction of $f$: $\mathcal{L}_{pred}(x, \bar{x} \mid f)$, and
	\item a distance loss to minimize the distance $d$: $\mathcal{L}_{dist}(x, \bar{x} \mid d)$.
\end{enumerate*}
The complete loss is a linear combination of these two parts, with a weight $\beta \in \mathbb{R}_{>0}$:
\begin{align}
\label{eq:mainloss}
%\mathcal{L}(x, \bar{x} \mid f, d) = \mathcal{L}_{pred}(y_x, \bar{x} \mid f) + \beta \mathcal{L}_{dist}(x, \bar{x} \mid d), 
\mathcal{L}(x, \bar{x} \mid f, d) = \mathcal{L}_{pred}(x, \bar{x} \mid f) + \beta \mathcal{L}_{dist}(x, \bar{x} \mid d).
\end{align}
%where $y_x = \arg\max_{y} f(y \mid x)$ is the original predicted class according to $f$. 
The assumption here is that an optimal counterfactual example $\bar{x}^*$ can be found by minimizing the overall loss:
\begin{align}
\bar{x}^* = \arg\min_{\bar{x}} \mathcal{L}(x, \bar{x} \mid f, d).
\end{align}
\citet{wachter_counterfactual_2017} propose a prediction loss $\mathcal{L}_{pred}$ based on the mean-squared-error. 
A clear limitation of this approach is that it assumes $f$ is differentiable.
This excludes many commonly used ML models, including tree-based models, which we focus on in this work.

\section{Method: FOCUS}
\label{section:focus-method}
To mimic many real-world scenarios, we assume there exists a trained model $f$ that we need to explain. The goal here is not to create a new, inherently interpretable tree-based model, but rather to explain a model that already exists.

\subsection{Loss Function Definitions}

We use a hinge-loss since we assume a classification task:
\begin{align}
\mathcal{L}_{pred}(x, \bar{x} \mid f) = {\mathbbm{1}}\left[\arg\max_{y} f(y \mid x) = \arg\max_{y'} f(y' \mid \bar{x})\right] \cdot  f(y' \mid \bar{x}).
\end{align}
Allowing for flexibility in the choice of distance function allows us to tailor the explanations to the end-users' needs. We make the preferred notion of \emph{minimality} explicit through the choice of distance function. 
Given a differentiable distance function $d$, the distance loss is: 
\begin{align}
\mathcal{L}_{dist}(x, \bar{x}) = d(x, \bar{x}). 
\end{align}
Building off of \citet{wachter_counterfactual_2017}, we propose incorporating differentiable approximations of non-differentiable models to use in the gradient-based optimization framework. 
Since the approximation $\tilde{f}$ is derived from the original model $f$, it should match $f$ closely: $\tilde{f}(y \mid x) \approx f(y \mid x)$. 
We define the approximate prediction loss as follows:
\begin{align}
\widetilde{\mathcal{L}}_{pred}(x, \bar{x} \mid f, \tilde{f}) = \mathbbm{1}\left[\arg\max_{y} f(y \mid x) = \arg\max_{y'} f(y' \mid \bar{x})\right] \cdot  \tilde{f}(y' \mid \bar{x}).
\end{align}
This loss is based both on the original model $f$ and the approximation $\tilde{f}$:
the loss is active as long as the prediction according to $f$ has not changed, but its gradient is based on the differentiable $\tilde{f}$. 
This prediction loss encourages the perturbation to have a different prediction than the original instance by penalizing an unchanged instance. 
The approximation of the complete loss becomes:
\begin{equation}
\widetilde{\mathcal{L}}(x, \bar{x} \mid f, \tilde{f}, d) =\widetilde{\mathcal{L}}_{pred}(x, \bar{x} \mid f, \tilde{f}) + \beta \cdot \mathcal{L}_{dist}(x, \bar{x} \mid d).
\label{eq:approxloss}
\end{equation}
Since we assume that it approximates the complete loss, 
\begin{align}
\widetilde{\mathcal{L}}(x, \bar{x} \mid f, \tilde{f}, d) \approx \mathcal{L}(x, \bar{x} \mid f, d),
\end{align}
we also assume that an optimal counterfactual example can be found by minimizing it:
\begin{align}
\bar{x}^* \approx \arg\min_{\bar{x}} \, \widetilde{\mathcal{L}}(x, \bar{x} \mid f, \tilde{f}, d).
\label{eq:xbar}
\end{align}

\begin{figure}[t]
\centering
\includegraphics[scale=0.7]{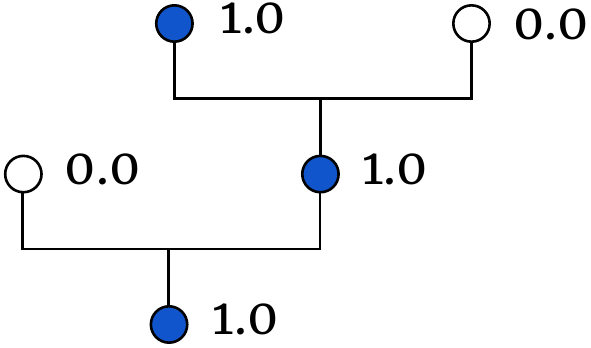} 
\quad
\includegraphics[scale=0.7]{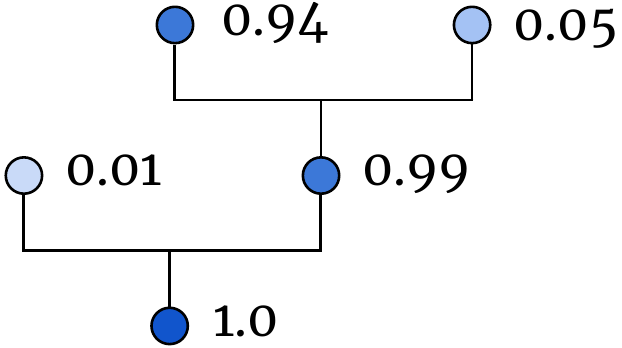}
\vspace{2mm}
\caption{
Left: A decision tree $\mathcal{T}$ and node activations for a single instance. Right: a differentiable approximation of the same tree $\widetilde{\mathcal{T}}$ and activations for the same instance.
}
\label{fig:exampletrees}
\end{figure}

\subsection{Tree-based Models}
To obtain the differentiable approximation $\tilde{f}$ of $f$, we construct a probabilistic approximation of the original tree ensemble $f$.
Tree ensembles are based on decision trees; a single decision tree $\mathcal{T}$ uses a binary-tree structure to make predictions about an instance $x$ based on its features.
Figure~\ref{fig:exampletrees} shows a simple decision tree consisting of five nodes.
A node $j$ is activated if its parent node $p_j$ is activated and feature $x_{f_j}$ is on the correct side of the threshold $\theta_j$; which side is the correct side depends on whether $j$ is a \emph{left} or \emph{right} child. 
The root note is an exception, it is always activated.
Let $t_j(x)$ indicate if node $j$ is activated:
\begin{equation}
\mbox{}\hspace*{-2mm}t_j(x) =
    \begin{cases}
   1, & \text{if $j$ is the root}, \\
   t_{p_j}(x) \cdot  \mathbbm{1}[x_{f_j} > \theta_j], & \text{if $j$ is a left child}, \\
    t_{p_j}(x) \cdot  \mathbbm{1}[x_{f_j} \leq \theta_j], &\text{if $j$ is a right child}.
    \end{cases}
\end{equation}
$\forall x, \,  t_0(x) = 1$.
Nodes that have no children are called \emph{leaf nodes}; an instance $x$ always ends up in a single leaf node.
Every leaf node $j$ has its own predicted distribution $\mathcal{T}(y \mid j)$; the prediction of the full tree is given by its activated leaf node. 
Let $\mathcal{T}_{\textit{leaf}}$ be the set of leaf nodes in $\mathcal{T}$, then:
\begin{equation}
(j \in \mathcal{T}_{\textit{leaf}} \land t_j(x) = 1) \rightarrow \mathcal{T}(y \mid x) = \mathcal{T}(y \mid j).
\end{equation}
Alternatively, we can reformulate this as a sum over leaves:
\begin{equation}
\mathcal{T}(y \mid x) = \sum_{j \in \mathcal{T}_\mathit{leaf}}  t_j(x) \cdot \mathcal{T}(y \mid j).
\end{equation}
Generally, tree ensembles are deterministic; let $f$ be an ensemble of $M$ many trees with weights $\omega_m \in \mathbb{R}$, then:
\begin{equation}
f(y \mid x)
=  \arg\max_{y'} \sum_{m=1}^M \omega_m \cdot \mathcal{T}_m(y' \mid x).
\end{equation}

\subsection{Approximations of Tree-based Models}
If $f$ is not differentiable, we are unable to calculate its gradient with respect to the input $x$. 
However, the non-differentiable operations in our formulation are 
\begin{enumerate*}[label=(\roman*)]
	\item the indicator function, and
	\item a maximum operation, 
\end{enumerate*}
both of which can be approximated by differentiable functions.
First, we introduce the $\widetilde{t}_j(x)$ function that \emph{approximates the activation of node} $j$: $\widetilde{t}_j(x) \approx t_j(x)$, using a sigmoid function with parameter $\sigma \in \mathbb{R}_{>0}$:
$
\textit{sig}(z) = (1 + \exp(\sigma \cdot z))^{-1}
$
and
\begin{align}
\widetilde{t}_j(x) &{} =
    \begin{cases}
    1, & \text{if $j$ is the root}, \\
   \widetilde{t}_{p_j}(x) \cdot \textit{sig}(\theta_j {-} x_{f_j}), & \text{if $j$ is left child}, \\
   \widetilde{t}_{p_j}(x) \cdot  \textit{sig}( x_{f_j} {-} \theta_j), & \text{if $j$ is right child}.
    \end{cases}
\label{eq:sigma}    
\end{align}
As $\sigma$ increases, $\widetilde{t}_j$ approximates $t_j$ more closely.
Next, we introduce a \emph{tree approximation}:
\begin{equation}
\widetilde{\mathcal{T}}(y \mid x) = \sum_{j \in \mathcal{T}_\mathit{leaf}}  \widetilde{t}_j(x) \cdot \mathcal{T}(y \mid j).
\end{equation}
The approximation $\widetilde{\mathcal{T}}$ uses the same tree structure and thresholds as $\mathcal{T}$.
However, its activations are no longer deterministic but instead are dependent on the distance between the feature values $x_{f_j}$ and the thresholds $\theta_j$.
Lastly, we replace the maximum operation of $f$ by a softmax with temperature $\tau\in\mathbb{R}_{>0}$, resulting in:
\begin{align}
\tilde{f}(y \mid x)
= \frac{
\exp\left(\tau \cdot \sum_{m=1}^M \omega_m \cdot \widetilde{\mathcal{T}}_m(y \mid x)\right)
}{
\sum_{y'} \exp\left(\tau \cdot \sum_{m=1}^M \omega_m \cdot \widetilde{\mathcal{T}}_m(y' \mid x)\right)
}.
\label{eq:tau}
\end{align}
The approximation $\tilde{f}$ is based on the original model $f$ and the parameters $\sigma$ and $\tau$.
This approximation is applicable to any tree-based model, and 
how well $\tilde{f}$ approximates $f$ depends on the choice of $\sigma$ and $\tau$.
The approximation is potentially perfect since
\begin{align}
\lim_{\sigma,\tau\rightarrow\infty}
\tilde{f}(y \mid x) = f(y \mid x).
\end{align}
\subsection{Our Method: FOCUS}
We call our method FOCUS: Flexible Optimizable CounterfactUal Explanations for Tree EnsembleS. 
It takes as input an instance $x$, a tree-based classifier $f$, and two hyperparameters: $\sigma$ and $\tau$, which we use to create the approximation $\tilde{f}$. 
Following Equation~\ref{eq:xbar}, FOCUS outputs the optimal counterfactual example $\bar{x}^*$, from which we derive the optimal counterfactual explanation $\Delta^*_{x} = \bar{x}^* - x$. 
%\mdr{Now introduce the FOCUS acronym and say what the method is, including what the inputs and outputs of the method are.}

\subsection{Effects of Hyperparameters}
Increasing $\sigma$ in $\tilde{f}$ eventually leads to exact approximations of the indicator functions, while increasing $\tau$ in $\tilde{f}$ leads to a completely unimodal softmax distribution. 
It should be noted that our approximation $\tilde{f}$ is not intended to replace the original model $f$ but rather to create a differentiable version of $f$ from which we can generate counterfactual examples through optimization. 
In practice, the original model $f$ would still be used to make predictions and the approximation would solely be used to generate counterfactual examples. 

%!TEX root = ../main.tex

\section{Experimental Setup}
\label{section:focus-exp-setup}

We consider \numprint{42} experimental settings to find the best counterfactual explanations using FOCUS. 
We jointly tune the hyperparameters of FOCUS ($\sigma, \tau, \beta, \alpha$) using Adam~\citep{kingma_adam:_2017} for \numprint{1000} iterations. 
We choose the hyperparameters that produce
\begin{enumerate*}[label=(\roman*)]
	\item a valid counterfactual example for every instance in the dataset, and
	\item the smallest mean distance between corresponding pairs ($x$, $\bar{x}$).
\end{enumerate*}

We evaluate FOCUS on four binary classification datasets and three types of tree-based models for each dataset. 
We compare against two baselines that generate counterfactual examples for tree ensembles based on the inner workings of the model: Feature Tweaking (FT) by \citet{tolomei_interpretable_2017} and Distribution-Aware Counterfactual Explanations (DACE) by \citet{kanamori_dace_2020}.

\subsection{Datasets}
\label{section:focus-datasets}
We evaluate FOCUS on four binary classification tasks using the following datasets:
\begin{itemize}
	\item The \textsc{Wine Quality} dataset \citep{wine_2009} has \numprint{4898} instances and \numprint{11} features. The task is about predicting the quality of white wine on a 0--10 scale. We adapt this to a binary classification setting by labelling the wine as ``high quality'' if the quality is $\geq$ \numprint{7}.

	\item The \textsc{HELOC} dataset \citep{fico_2017} has \numprint{10459} instances and \numprint{23} features. The task is from the Explainable Machine Learning Challenge at NeurIPS 2017, where the task is to predict whether or not a customer will default on their loan. 
	
	\item The \textsc{COMPAS} dataset \citep{compas-dataset-2017} has \numprint{6172} instances and \numprint{6} features. It is used for detecting bias in ML systems, where the task is predicting whether or not a criminal defendant will reoffend upon release. 

	\item The \textsc{Shopping} dataset \citep{shoppingdataset} has \numprint{12330} instances and \numprint{9} features. The task entails predicting whether or not an online website visit results in a purchase. 
\end{itemize}
We scale all features such that their values are in the range $\left[0, 1\right]$ and remove categorical features.

\subsection{Models}
We train three types of tree-based models on \numprint{70}\% of each dataset: decision trees (DTs), random forests (RFs), and adaptive boosting (AB) with DTs as the base learners. 
We use the remaining \numprint{30}\% to find counterfactual examples for this test set. 
In total we have \numprint{12} models (\numprint{4} datasets $\times$ \numprint{3} tree-based models).

\subsection{Distance Functions}
\label{section:distance-metrics}
In our experiments, we generate different types of counterfactual explanations using different types of distance functions. 
We note that the flexibility of FOCUS allows for the use of any differentiable distance function. 
Euclidean distance measures the geometric displacement: 
\begin{align}
\label{eq:euclidean}
d_\mathit{Euclidean}(x, \bar{x}) = \sqrt{\sum_{i} (x_i - \bar{x}_i)^2}.
\end{align}
Cosine distance measures the angle by which $\bar{x}$ deviates from $x$ -- whether $\bar{x}$ preserves the relationship between features in $x$:
\begin{align}
\label{eq:cosine}
d_\mathit{Cosine}(x, \bar{x}) = 1 - \frac{\sum_{i} \left( x_i \cdot \bar{x}_i \right)}{\norm{x} \norm{\bar{x}}}.
\end{align}
Manhattan distance (i.e., $L1$-norm) measures per feature differences, minimizing the number of features perturbed and therefore inducing sparsity:
\begin{align}
\label{eq:manhatten}
d_\mathit{Manhattan}(x, \bar{x}) = \sum_{i} |x_i - \bar{x}_i|.
\end{align}
When comparing against DACE \citep{kanamori_dace_2020}, we use the Mahalanobis distance, since this is the distance function used in their novel cost function (see Equation~\ref{eq:daceloss}):
\begin{align}
\label{eq:mahal}
d_\mathit{Mahalanobis}(x, \bar{x}|C) = \sqrt{(x - \bar{x})C^{-1}(x - \bar{x})}.
\end{align}
$C$ is the covariance matrix of $x$ and $\bar{x}$, which allows us to account for correlations between features. 
When all features are uncorrelated, the Mahalanobis distance is equal to the Euclidean distance.

\subsection{Evaluation Metrics}
\label{section:focus-evalmetrics}
We evaluate the counterfactual examples produced by FOCUS based on how close they are to the original input using three metrics, in terms of four distance functions (see Section~\ref{section:distance-metrics}).
The first evaluation metric is distance from the original input averaged over all examples, $d_\mathit{mean}$. 
Let $X$ be the set of $N$ original instances and $\bar{X}$ be the corresponding set of $N$ generated counterfactual examples.
The \emph{mean distance} is defined as:
\begin{equation}
\label{eq:mean-dist}
d_\mathit{mean}(X, \bar{X}) = \frac{1}{N}\sum_{n=1}^{N}d(x^{(n)}, \bar{x}^{(n)}).
\end{equation}
The second evaluation metric is mean relative distance from the original input, $d_\mathit{Rmean}$. 
This metric helps us interpret individual improvements over the baselines; if $d_\mathit{Rmean} < 1$, FOCUS's counterfactual examples are on average closer to the original input compared to the baseline. 
Let $\bar{X}$ be the set of counterfactual examples produced by FOCUS and let $\bar{X}'$ be the set of counterfactual examples produced by a baseline. 
Then the \emph{mean relative distance} is defined as:
\begin{equation}
\label{eq:mean-rel-dist}
d_\mathit{Rmean}(\bar{X}, \bar{X}') = \frac{1}{N}\sum_{n=1}^{N} \frac{d(x^{(n)}, \bar{x}^{(n)})}{d(x^{(n)}, {\bar{x}}^{'(n)})}.
\end{equation}
The third evaluation metric is the proportion of FOCUS's counterfactual examples that are closer to the original input in comparison to the baselines. 
For $d$ we consider Euclidean, Cosine, Manhattan, and Mahalanobis distance.

\if0
\subsection{Evaluation Metrics}
\label{section:evaluation}
We evaluate the counterfactual examples produced by FOCUS based on how close they are to the original input using three metrics. 
Mean distance, $d_\mathit{mean}$, measures the distance from the original input, averaged over all examples. 
Mean relative distance, $d_\mathit{Rmean}$, measures pointwise ratios of distance to the original input. 
This helps us interpret individual improvements over the baselines; if $d_\mathit{Rmean} < 1$, FOCUS's counterfactual examples are on average closer to the original input compared to the baseline. 
We also evaluate the proportion of FOCUS's counterfactual examples that are closer to the original input compared to the baselines ($\mathit{\%_{closer}}$). 
We test the metrics in terms of four distance functions: Euclidean, Cosine, Manhattan and Mahalanobis. 
\fi

%!TEX root = ../main.tex

\section{Experiment 1: FOCUS vs. FT}
\label{section:experiment1}

We compare FOCUS to the Feature Tweaking (FT) method by \citet{tolomei_interpretable_2017} in terms of the evaluation metrics in Section~\ref{section:focus-evalmetrics}. 
We consider \numprint{36} experimental settings (\numprint{4} datasets $\times$ \numprint{3} tree-based models $\times$ \numprint{3} distance functions) when comparing FOCUS to FT. 
The results are listed in Table~\ref{table:distances}. 

\subsection{Baseline: Feature Tweaking}
\label{section:baselineft}
FT identifies the leaf nodes where the prediction of the leaf nodes do not match the original prediction $y_x$: it recognizes the set of leaves that if activated, $t_j(\bar{x}) = 1$, would change the prediction of a tree $\mathcal{T}$:
\begin{equation}
\mathcal{T}_\textit{change} = \left\{ j \mid j \in   \mathcal{T}_\textit{leaf} \land y_x \not = \arg \max_y T(y\mid j) \right\}.
\end{equation}
For every $\mathcal{T}$ in $f$, FT generates a perturbed example per node in $\mathcal{T}_\textit{change}$ so that it is activated with at least an $\epsilon$ difference per threshold, and then selects the most optimal example (i.e., the one closest to the original instance).
For every feature threshold $\theta_j$ involved, the corresponding feature is perturbed accordingly: $\bar{x}_{f_j} = \theta_j \pm \epsilon$.
The result is a perturbed example that was changed minimally to activate a leaf node in $\mathcal{T}_\textit{change}$. 
In our experiments, we test $\epsilon \in \{0.001, 0.005, 0.01, 0.1\}$, and choose the $\epsilon$ that minimizes the mean distance to the original input, while maximizing the number of counterfactual examples generated.

\begin{figure}[t]
\centering
\includegraphics[scale=0.23]{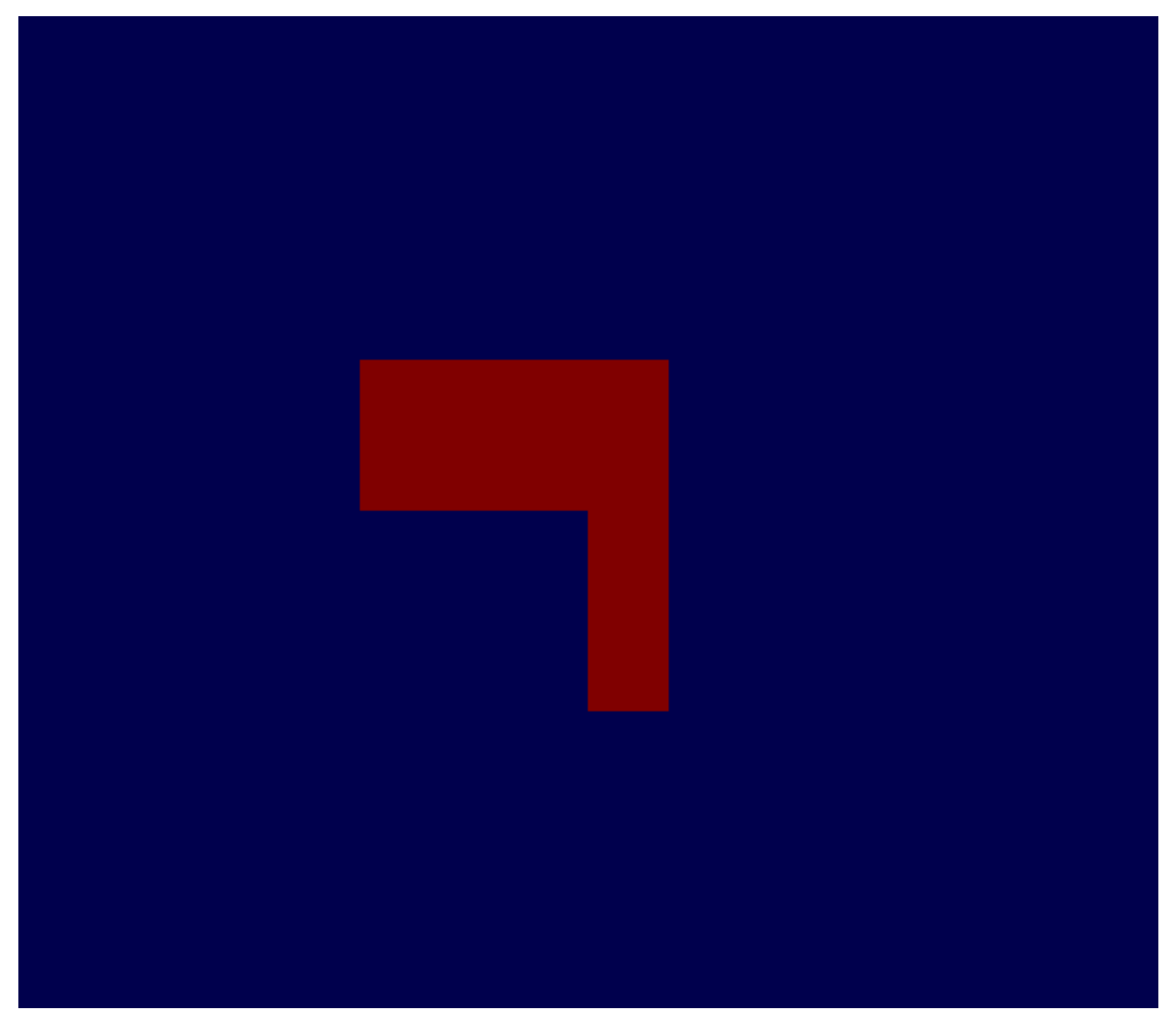} 
\includegraphics[scale=0.23]{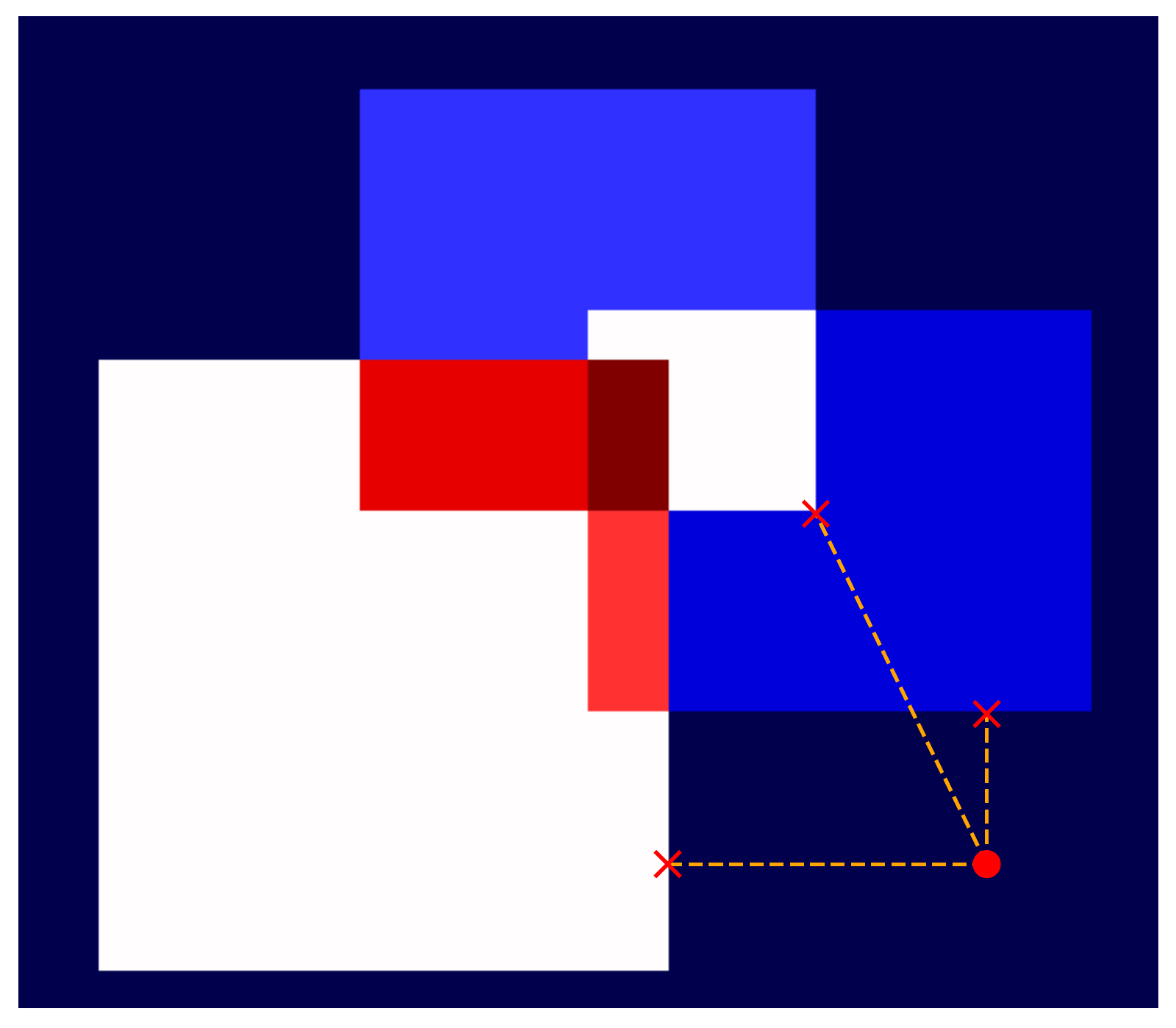} 
\includegraphics[scale=0.23]{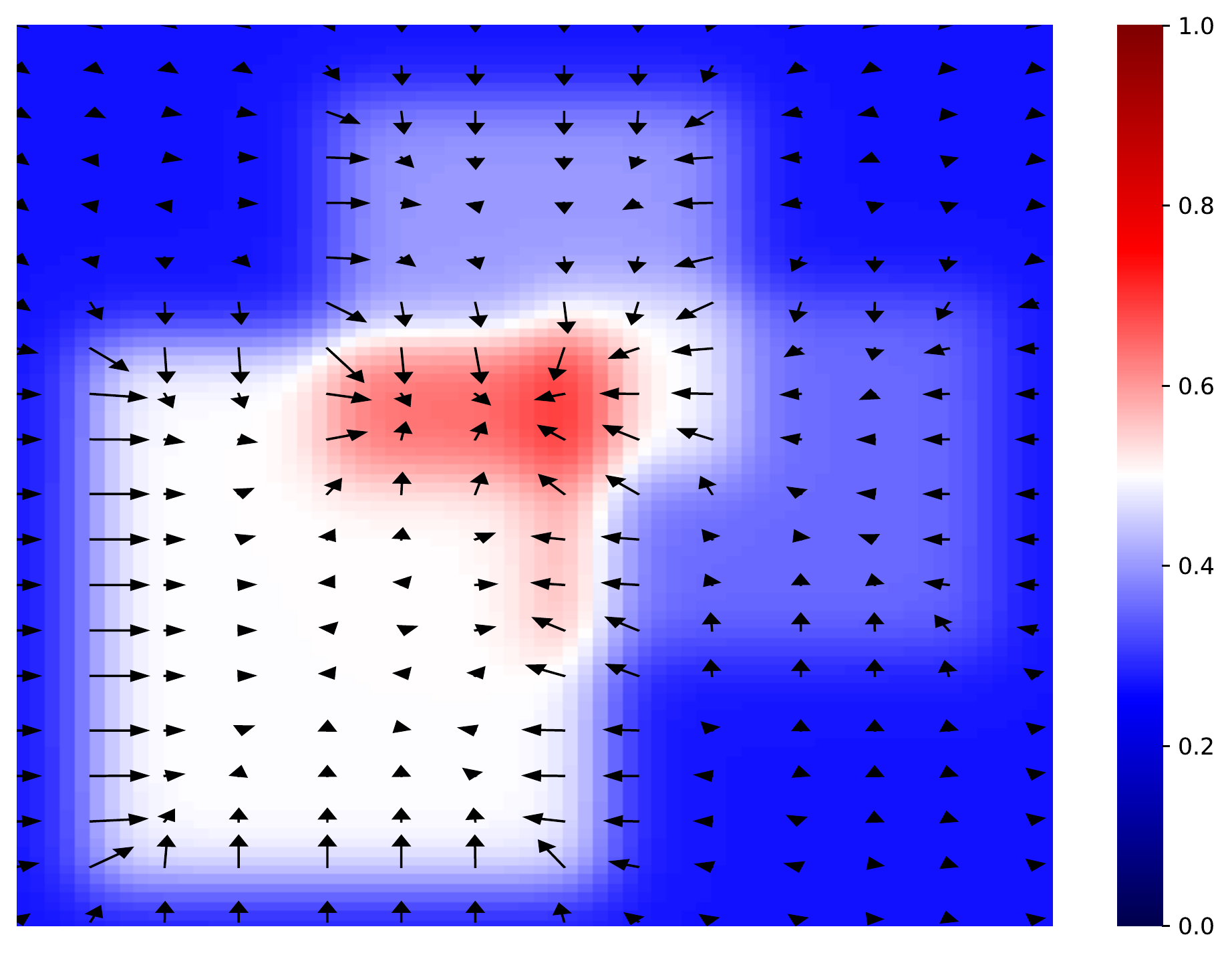}
\caption{
An example of how the FT baseline method (explained in Section~\ref{section:baselineft}) and our method handle an adaptive boosting ensemble with three trees.
Left: decision boundary of the ensemble.
Middle: three positive leaves that form the decision boundary, an example instance, and the perturbed examples suggested by \acs{FT}. 
Right: approximated loss $\widetilde{\mathcal{L}}_{pred}$ and its gradient w.r.t. $\bar{x}$. 
The \acs{FT} perturbed examples do not change the prediction of the forest, whereas the gradient of the differentiable approximation leads toward the true decision boundary.
}
\label{fig:approxensemble}
\end{figure}

The main problem with FT is that the perturbed examples are not necessarily counterfactual examples, since changing the prediction of a single tree $\mathcal{T}$ does not guarantee a change in the prediction of the full ensemble $f$.
Figure~\ref{fig:approxensemble} shows all three perturbed examples generated by FT for a single instance. 
In this case, none of the generated examples change the model prediction and therefore none are valid counterfactual examples. 

Figure~\ref{fig:approxensemble} shows how FOCUS and FT handle an adaptive boosting ensemble using a two-feature ensemble with three trees. 
On the left is the decision boundary for a standard tree ensemble; the middle visualizes the positive leaf nodes that form the decision boundary; on the right is the approximated loss $\widetilde{\mathcal{L}}_{pred}$ and its gradient w.r.t. $\bar{x}$.
The gradients push features close to thresholds harder and in the direction of the decision boundary if $\widetilde{\mathcal{L}}$ is convex.

\subsection{Results}
In terms of $d_\mathit{mean}$, FOCUS outperforms FT in \numprint{20} settings while FT outperforms FOCUS in \numprint{8} settings. The difference in $d_\mathit{mean}$ is not significant in the remaining \numprint{8} settings. 
In general, FOCUS outperforms FT in settings using Euclidean and Cosine distance because in each iteration, FOCUS perturbs many of the features by a small amount. 
Since FT perturbs only the features associated with an individual leaf, we expected that it would perform better for Manhattan distance but our results show that this is not the case. 
There is no clear winner between FT and FOCUS for Manhattan distance. 

\begin{landscape}
\begin{table}
\caption{Evaluation results for Experiment 1 comparing FOCUS and FT counterfactual examples. Significant improvements and losses over the baseline (FT) are denoted by \dubbelneer\ and \dubbelop, respectively ($p < 0.05$, two-tailed t-test,); 
\notsig{} denotes no significant difference;
\NoExample{} denotes settings where the baseline cannot find a counterfactual example for every instance.}
%!TEX root = ../main.tex

\begin{tabular}{llrrrrrrrrrr}
\toprule
  &                   &                  & \multicolumn{3}{c}{\textbf{Euclidean}}                                                              & \multicolumn{3}{c}{\textbf{Cosine}}                                                                 & \multicolumn{3}{c}{\textbf{Manhattan}}                                                              \\
  \cmidrule(r){4-6}\cmidrule(r){7-9} \cmidrule(r){10-12}
 \textbf{Dataset}                 &     \textbf{Metric}                & \textbf{Method}                 & \multicolumn{1}{c}{\textbf{DT}} & \multicolumn{1}{c}{\textbf{RF}} & \multicolumn{1}{c}{\textbf{AB}} & \multicolumn{1}{c}{\textbf{DT}} & \multicolumn{1}{c}{\textbf{RF}} & \multicolumn{1}{c}{\textbf{AB}} & \multicolumn{1}{c}{\textbf{DT}} & \multicolumn{1}{c}{\textbf{RF}} & \multicolumn{1}{c}{\textbf{AB}} \\
 \midrule
\textit{}        &    $d_{mean}$                & FT               & 0.269 & \textbf{0.174}  & 0.267\rlap\NoExample  & 0.030 & 0.017  & 0.034\rlap\NoExample  & 0.269 & \textbf{0.223} & 0.382\rlap\NoExample  \\
\textsc{Wine}    &                    & FOCUS              & \textbf{0.268}\rlap{\notsig} & 0.188\rlap{\dubbelop}   & \textbf{0.188}\rlap{\dubbelneer}  & \textbf{0.003}\rlap{\dubbelneer} & \textbf{0.008}\rlap{\dubbelneer}  & \textbf{0.014}\rlap{\dubbelneer}  & \textbf{0.268}\rlap{\notsig} & 0.312\rlap{\dubbelop} & \textbf{0.360}\rlap{\dubbelneer}  \\
\cmidrule{2-12}
\textsc{Quality}        &       $d_{Rmean}$             & FOCUS/FT           & 0.990 & 1.256  & 0.649  & 0.066 & 0.821  & 0.312  & 0.990 & 1.977 & 0.924  \\
%\cmidrule{2-12}
\textit{}        &      $\mathit{\%_{closer}}$              & FOCUS \textless FT                           & 100\% & 21.0\% & 87.5\% & 100\% & 80.8\% & 95.1\% & 100\% & 5.4\% & 58.6\%                           \\

\midrule

                 &      $d_{mean}$              & FT               & \textbf{0.120}  & 0.210  & 0.185  & 0.003  & 0.008  & 0.007  & \textbf{0.135}  & \textbf{0.278}  & \textbf{0.198}  \\
  \textsc{HELOC}               &                    & FOCUS              & 0.133\rlap{\dubbelop}  & \textbf{0.186}\rlap{\dubbelneer}  & \textbf{0.136}\rlap{\dubbelneer}  & \textbf{0.001}\rlap{\dubbelneer}  & \textbf{0.002}\rlap{\dubbelneer}  & \textbf{0.001}\rlap{\dubbelneer}  & 0.152\rlap{\dubbelop}  & 0.284\rlap{\notsig}  & 0.203\rlap{\notsig}  \\
\cmidrule{2-12}

                 &     $d_{Rmean}$               & FOCUS/FT           & 1.169  & 0.942  & 0.907  & 0.303  & 0.285  & 0.421  & 1.252  & 1.144  & 1.364  \\ 
%\cmidrule{2-12}
                 &     $\mathit{\%_{closer}}$               & FOCUS \textless FT & 16.6\% & 57.9\% & 71.9\% & 91.6\% & 91.5\% & 92.9\% & 51.3\% & 43.6\% & 24.2\%          \\
 
 \midrule

                 &      	$d_{mean}$              & FT               & \textbf{0.082} & \textbf{0.075} & 0.081 & 0.013 & 0.014 & 0.015 & \textbf{0.086} & \textbf{0.078} & \textbf{0.085} \\
  \textsc{COMPAS}               &              & FOCUS              & 0.092\rlap{\dubbelop} & 0.079\rlap{\notsig} & \textbf{0.076}\rlap{\dubbelneer} & \textbf{0.008}\rlap{\dubbelneer} & \textbf{0.011}\rlap{\dubbelneer} & \textbf{0.007}\rlap{\dubbelneer} & 0.093\rlap{\dubbelop} & 0.085\rlap{\notsig} & 0.090\rlap{\notsig} \\
\cmidrule{2-12}
                 &      $d_{Rmean}$              & FOCUS/FT           & 1.162 & 1.150 & 1.062 & 0.473 & 0.965 & 0.539 & 1.182 & 1.236 & 1.155 \\ 
%\cmidrule{2-12}
                 &      $\mathit{\%_{closer}}$              & FOCUS \textless FT & 29.4\% & 22.6\% & 44.8\% & 82.7\% & 68.0\% & 84.8\% & 65.8\% & 36.2\% & 66.9\% \\
\midrule

                 &      	$d_{mean}$              & FT                & \textbf{0.119}  & 0.028  & 0.126\rlap\NoExample  & \textbf{0.050}  & 0.027  & 0.131\rlap\NoExample  & \textbf{0.121}  & 0.030  & 0.142\rlap\NoExample  \\
   \textsc{Shopping}              &              & FOCUS              & 0.142\rlap{\dubbelop}  & \textbf{0.025}\rlap{\dubbelneer}  & \textbf{0.028}\rlap{\dubbelneer}  & 0.055\rlap{\dubbelop}  & \textbf{0.013}\rlap{\dubbelneer}  & \textbf{0.006}\rlap{\dubbelneer}  & 0.128\rlap{\notsig}  & \textbf{0.026}\rlap{\dubbelneer}  & \textbf{0.046}\rlap{\dubbelneer}  \\ 
\cmidrule{2-12}
                 &      $d_{Rmean}$              & FOCUS/FT           & 1.051  & 1.053  & 0.218  & 0.795  & 0.482  & 0.074  & 0.944  & 0.796  & 0.312  \\ 
%\cmidrule{2-12}
                 &    $\mathit{\%_{closer}}$                & FOCUS \textless FT & 40.2\% & 36.1\% & 99.6\% & 44.4\% & 86.1\% & 99.5\% & 55.8\% & 81.9\% & 97.1\% \\

 \bottomrule                
\end{tabular}

\label{table:distances}
\end{table}
\end{landscape}

\noindent
We also see that FOCUS usually outperforms FT in settings using random forests and adaptive boosting, while the opposite is true for decision trees. 

Overall, we find that FOCUS is effective and efficient for finding counterfactual explanations for tree-based models.
Unlike the FT baseline, FOCUS finds valid counterfactual explanations for \emph{every} instance across all settings. 
In the majority of tested settings, FOCUS's explanations are substantial improvements in terms of distance to the original inputs, across all three metrics.

%!TEX root = ../main.tex

\section{Experiment 2: FOCUS vs. DACE}
\label{section:experiment2}
The flexibility of FOCUS allows us to plug in our choice of differentiable distance function. To compare against DACE~\citep{kanamori_dace_2020}, we use the Mahalanobis distance for both 
\begin{inparaenum}[(i)]
\item generation of FOCUS explanations, and
\item evaluation in comparison to DACE, since this is the distance function used in the DACE loss function (see Equation~\ref{eq:daceloss} in Section~\ref{section:baselinedace}). 
\end{inparaenum}

We found two main limitations of DACE:
\begin{inparaenum}[(i)]
	\item in all of our settings, it can only generate counterfactual examples for a subset of the test set, and
	\item it is limited by the size of the tree-based model. 
\end{inparaenum}
All hyperparameter settings are listed in the Appendix to this chapter.

\subsection{Baseline: DACE}
\label{section:baselinedace}
DACE generates counterfactual examples that account for the underlying data distribution through a novel cost function using Mahalanobis distance and a local outlier factor (LOF):
\begin{align}
\label{eq:daceloss}
d_\mathit{DACE}(x, \bar{x}|X, C) = {d_\mathit{Mahalanobis}}^2(x, \bar{x}|C) + \lambda q_k(x, \bar{x}|X), 
\end{align}
where $C$ is the covariance matrix, $q_k$ is the $k$-LOF \citep{breunig_lof_2020}, $X$ is the training set, and $\lambda$ is the trade-off parameter. 
The $k$-LOF measures the degree to which an instance is an outlier in the context of its $k$-nearest neighbors.\footnote{We use $k=1$ in our experiments, since this is the value supported in the original code.}
To generate counterfactual examples, DACE formulates the task as a mixed-integer linear optimization problem and uses the CPLEX Optimizer\footnote{\url{http://www.ibm.com/analytics/cplex-optimizer}} to solve it. 
We refer the reader to the original paper for a more detailed overview of this cost function. 
The $q_k$ term in the loss function penalizes counterfactual examples that are outliers, and therefore decreasing $\lambda$ results in a greater number of counterfactual examples. 
In our experiments, we test $\lambda \in \{0.001, 0.01, 0.1, 0.5, 1.0\}$, and choose the $\lambda$ that minimizes the mean distance to the original input, while maximizing the number of counterfactual examples generated.

We were only able to run DACE on \numprint{6} out of our \numprint{12} models because the problem size is too large (i.e., there are too many model parameters for DACE) for the remaining \numprint{6} models when using the free Python API of CPLEX (the optimizer used in DACE). 
Specifically, we were unable to run DACE on the following settings:
\begin{itemize}
\setlength\itemsep{0.4em}
	\item Wine Quality AB (100 trees, max depth 4)
	\item Wine Quality RF (500 trees, max depth 4)
	\item HELOC RF (500 trees, max depth 4)
	\item HELOC AB (100 trees, max depth 8)
	\item COMPAS RF (500 trees, max depth 4)
	\item Shopping RF (500 trees, max depth 8).
\end{itemize}
Therefore, when comparing against DACE, we have \numprint{6} experimental settings (\numprint{6} models $\times$ \numprint{1} distance function).
We note that these are not unreasonable model sizes, and that unlike DACE, FOCUS can be applied to all \numprint{12} models (see Table~\ref{table:distances}).

\subsection{Results}
Table 2 shows the results for the \numprint{6} settings we could run DACE on. 
We were only able to run DACE on \numprint{6} out of our \numprint{12} models because the problem size is too large (i.e., DACE has too many model parameters) for the remaining \numprint{6} models when using the free Python API of CPLEX (the optimizer used in DACE). 
Therefore, when comparing against DACE, we have \numprint{6} experimental settings (\numprint{6} models $\times$ \numprint{1} distance function).

We found that DACE can only generate counterfactual examples for a small subset of the test set, regardless of the $\lambda$-value, as opposed to FOCUS, which can generate counterfactual examples for the entire test set in all cases. 
To compute $d_{mean}$, $d_{Rmean}$, and $\mathit{\%_{closer}}$, we compare FOCUS and DACE only on the instances for which DACE was able to generate a counterfactual example. 
We find that FOCUS significantly outperforms DACE in \numprint{5} out of \numprint{6} settings in terms of all three evaluation metrics, indicating that FOCUS explanations are indeed more minimal than those produced by DACE. 
FOCUS is also more reliable since 
\begin{inparaenum}[(i)]
\item it is not restricted by model size, and
\item it can generate counterfactual examples for all instances in the test set. 
\end{inparaenum}

\begin{table*}[h!]
\centering
\caption{Evaluation results for Experiment 2 comparing FOCUS and DACE counterfactual examples in terms of Mahalanobis distance. Significant improvements over the baseline are denoted by \dubbelneer\ ($p < 0.05$, two-tailed t-test,). 
\notsig{} denotes no significant difference.}
\begin{tabular}{ll@{}rrrrrr}
\toprule
                &                            & \multicolumn{1}{c}{\textsc{Wine}} & \multicolumn{1}{c}{\textsc{HELOC}} & \multicolumn{2}{c}{\textsc{COMPAS}}                               & \multicolumn{2}{c}{\textsc{Shopping}}                             \\
                \cmidrule(r){3-3}\cmidrule(r){4-4}\cmidrule(r){5-6}\cmidrule{7-8}
\textbf{Metric} & \textbf{Method}            & \multicolumn{1}{c}{\textbf{DT}}   & \multicolumn{1}{c}{\textbf{DT}}    & \multicolumn{1}{c}{\textbf{DT}} & \multicolumn{1}{c}{\textbf{AB}} & \multicolumn{1}{c}{\textbf{DT}} & \multicolumn{1}{c}{\textbf{AB}} \\ \midrule

$d_{mean}$           & DACE              & 1.325                             & {1.427}                              & 0.814                           & 1.570                           & 0.050                           & 3.230                           \\
                & FOCUS             & \textbf{0.542}\rlap{\dubbelneer}                             & \textbf{0.810}\rlap{\dubbelneer}                              & \textbf{0.776}\rlap{\notsig}                           & \textbf{0.636}\rlap{\dubbelneer}                           & \textbf{0.023}\rlap{\dubbelneer}                           & \textbf{0.303}\rlap{\dubbelneer}                           \\ \midrule
$d_{Rmean}$          & FOCUS /           & \multirow{2}{*}{0.420}            & \multirow{2}{*}{0.622}             & \multirow{2}{*}{1.18}          & \multirow{2}{*}{0.372}          & \multirow{2}{*}{0.449}          & \multirow{2}{*}{0.380}          \\
                & DACE              &                                   &                                    &                                 &                                 &                                 &                                 \\ \midrule
$\mathit{\%_{closer}}$          & FOCUS \textless{} & \multirow{2}{*}{100\%}            & \multirow{2}{*}{94.5\%}             & \multirow{2}{*}{29.9\%}          & \multirow{2}{*}{96.1\%}          & \multirow{2}{*}{99.4\%}          & \multirow{2}{*}{90.8\%}          \\
                & DACE              &                                   &                                    &                                 &                                 &                                 &                                 \\ 
\midrule
 \textit{\# CFs}        & DACE              & 241                               & 1342                               & 842                             & 700                             & 362                             & 448                             \\
  \textit{found}              & FOCUS             & 1470                              & 3138                               & 1852                            & 1852                            & 3699                            & 3699                            \\ \midrule 
                
   \textit{\# obs      in}	& \textit{dataset}	& 1470                              & 3138                               & 1852                            & 1852                            & 3699                            & 3699                            \\ \bottomrule
\end{tabular}
\label{table:experiment2}
\end{table*}
%!TEX root = ../main.tex

\section{Discussion and Analysis}
\label{section:case-study}

\begin{figure}[t]
\begin{center}
\includegraphics[scale=0.45]{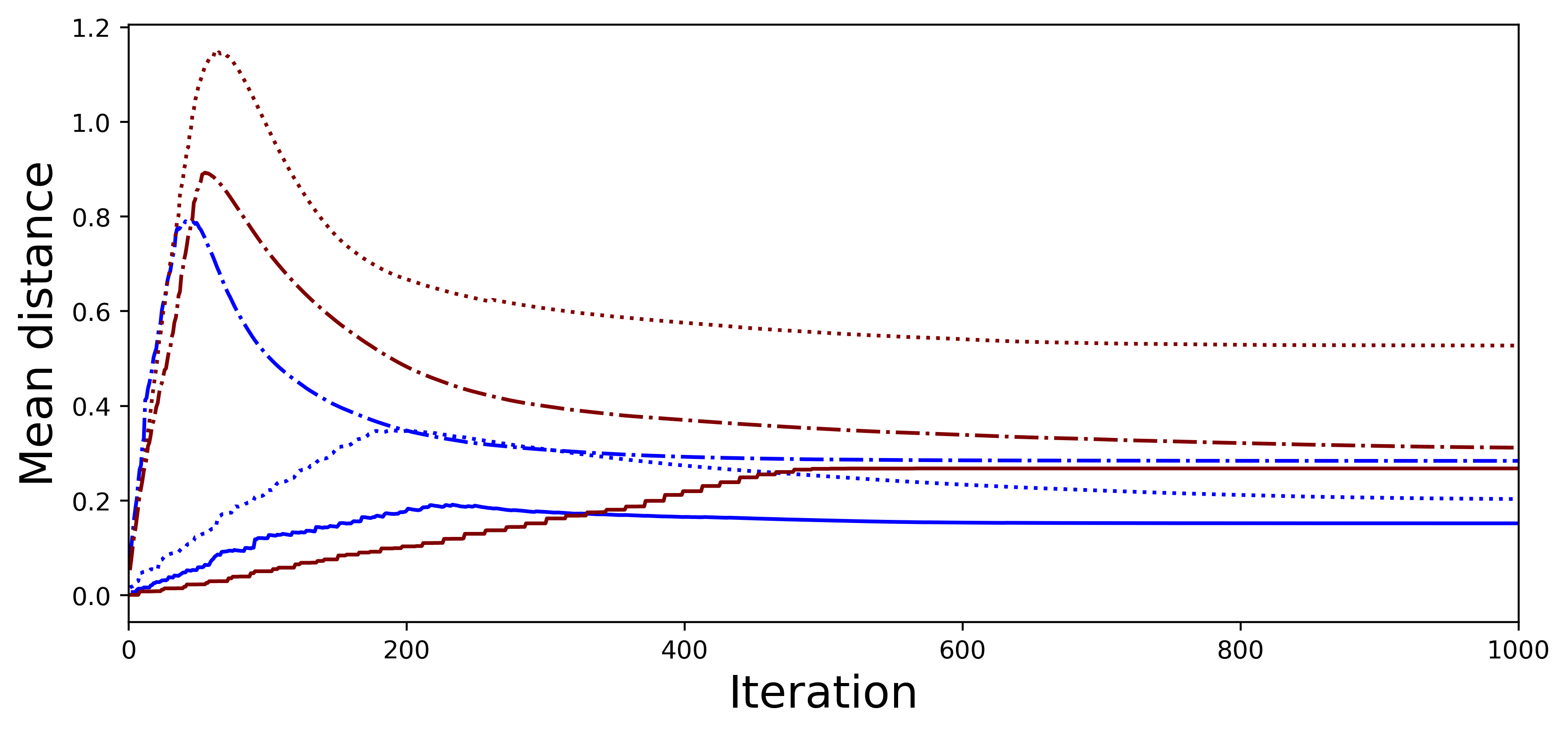} 
\includegraphics[scale=0.45]{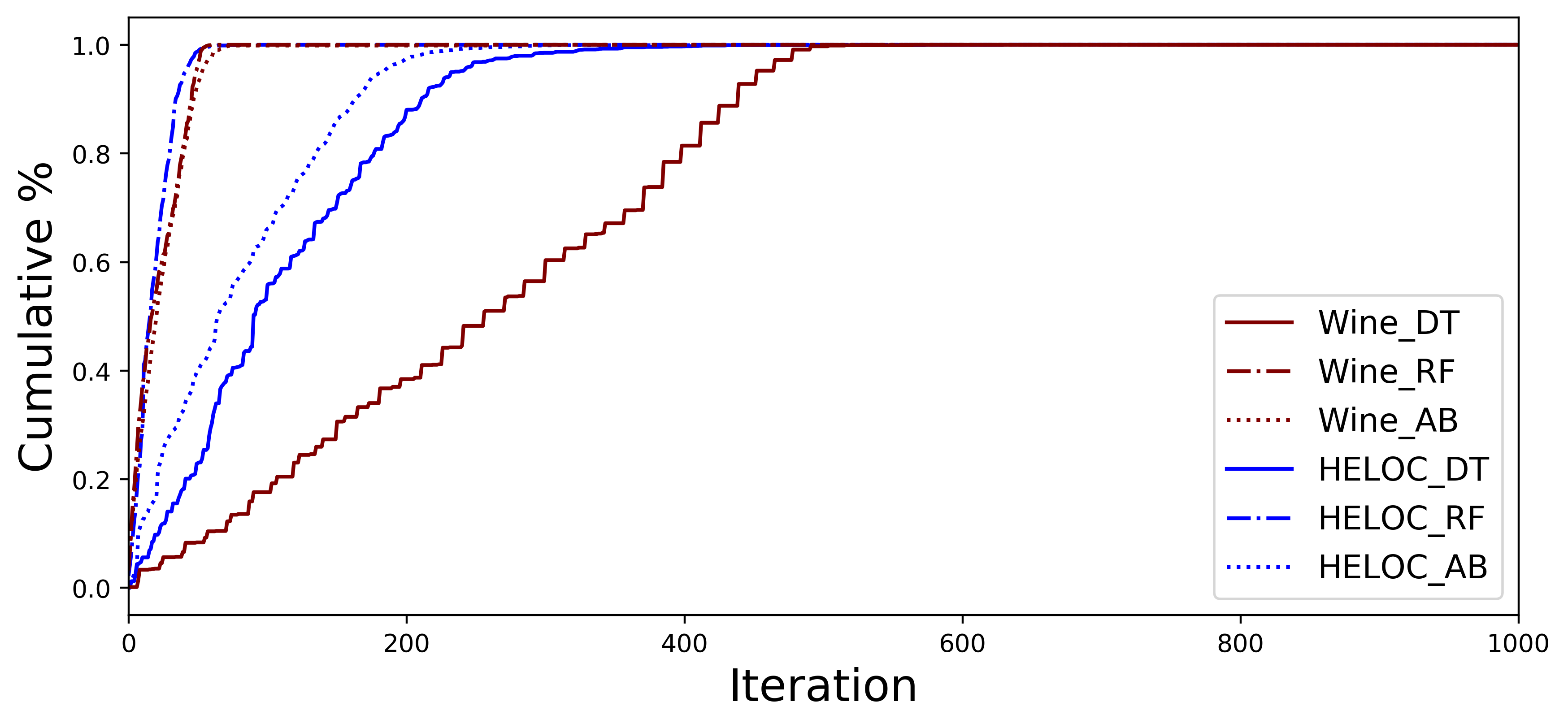} 
\end{center}
\caption{Mean distance (top) and cumulative \% (bottom) of counterfactual examples in each iteration of FOCUS for Manhattan explanations.}
\label{fig:distances}
\end{figure}

Figure~\ref{fig:distances} shows the mean Manhattan distance of the perturbed examples in each iteration of FOCUS, along with the proportion of perturbations resulting in valid counterfactual examples found for two datasets (we omit the others due to space considerations). These trends are indicative of all settings: the mean distance increases until a counterfactual example has been found for every $x$, after which the mean distance starts to decrease. This seems to be a result of the hinge-loss in FOCUS, which first prioritizes finding a valid counterfactual example (see Equation~\ref{eq:cfexample}), then decreasing the distance between $x$ and $\bar{x}$. 

\begin{figure*}[h!]
\centering
\includegraphics[scale=0.26]{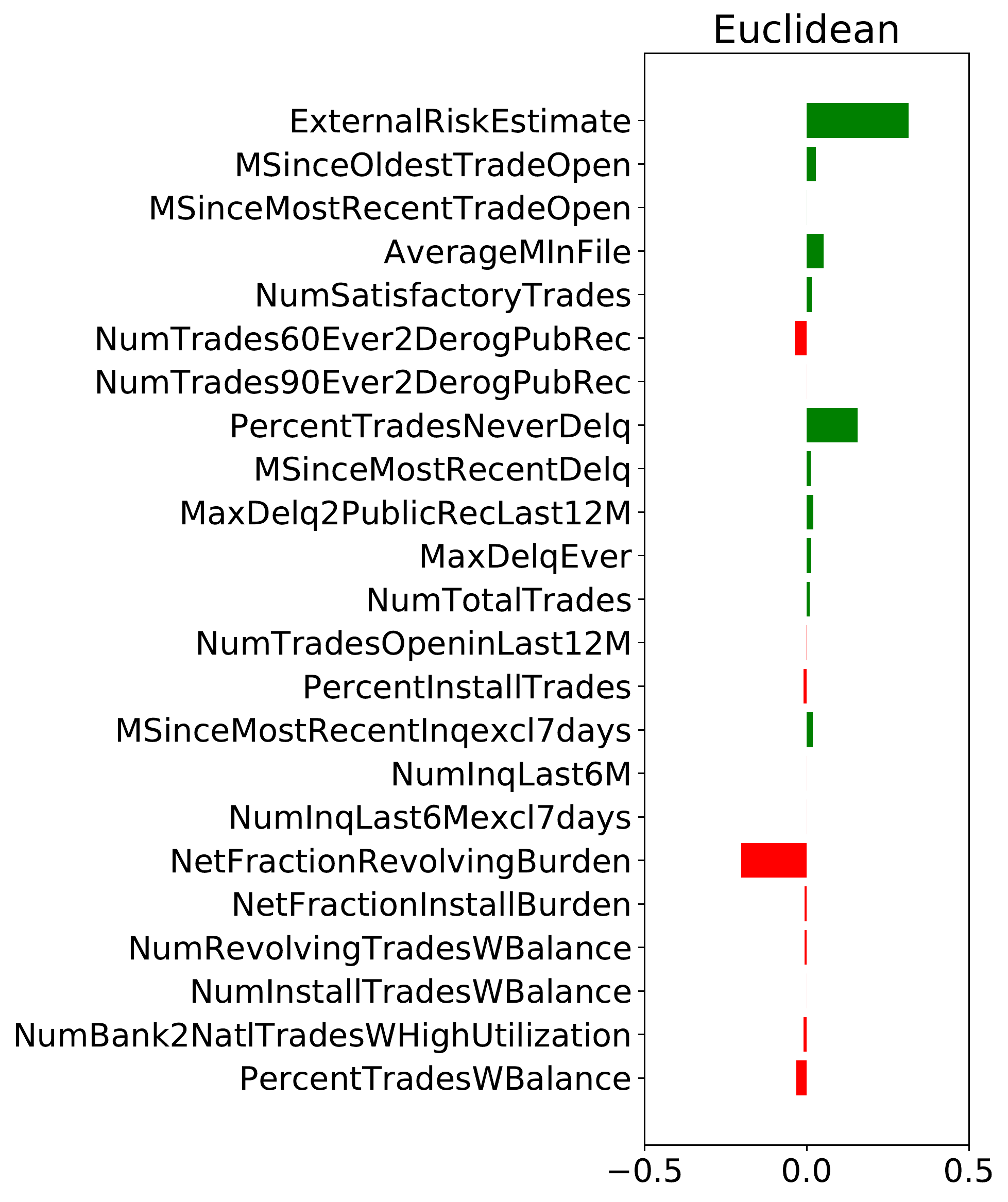}
\includegraphics[scale=0.26]{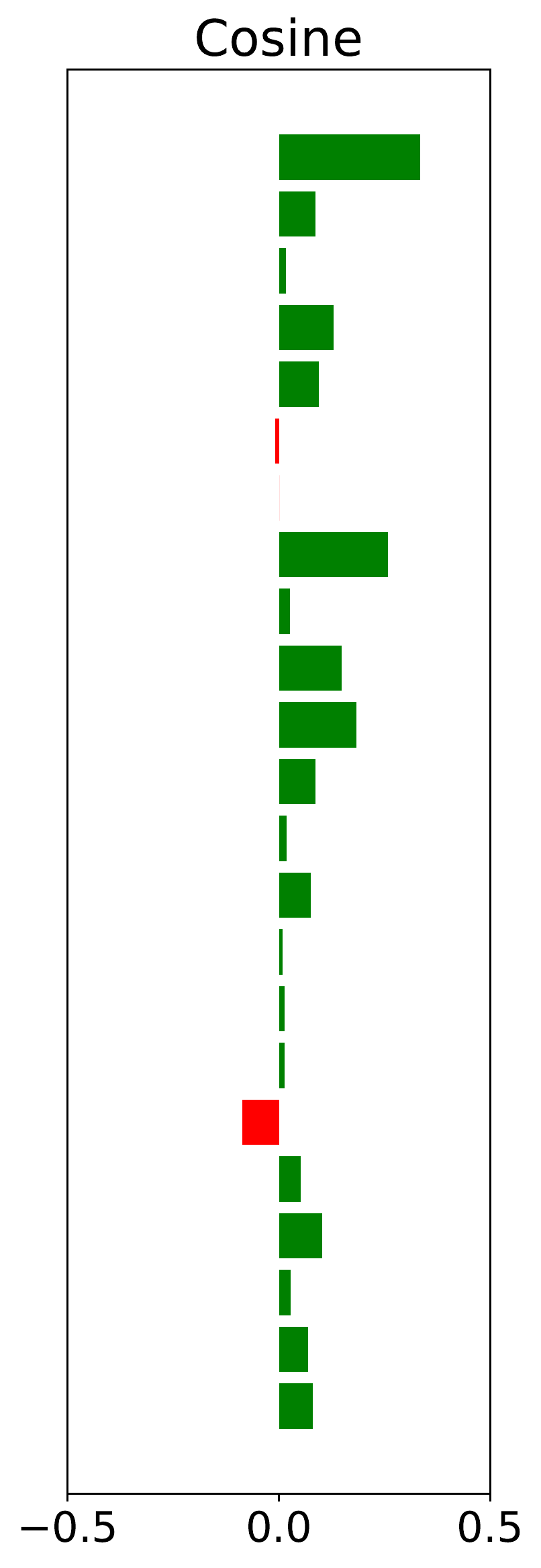} 
\includegraphics[scale=0.26]{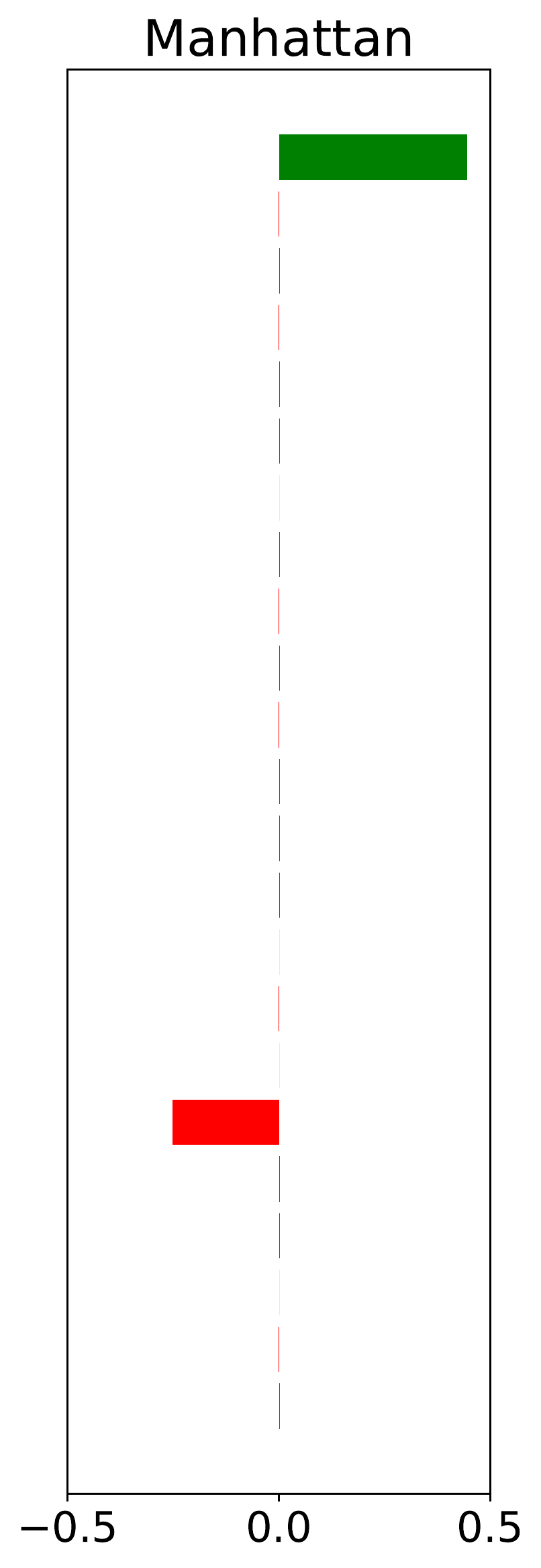} 
\includegraphics[scale=0.26]{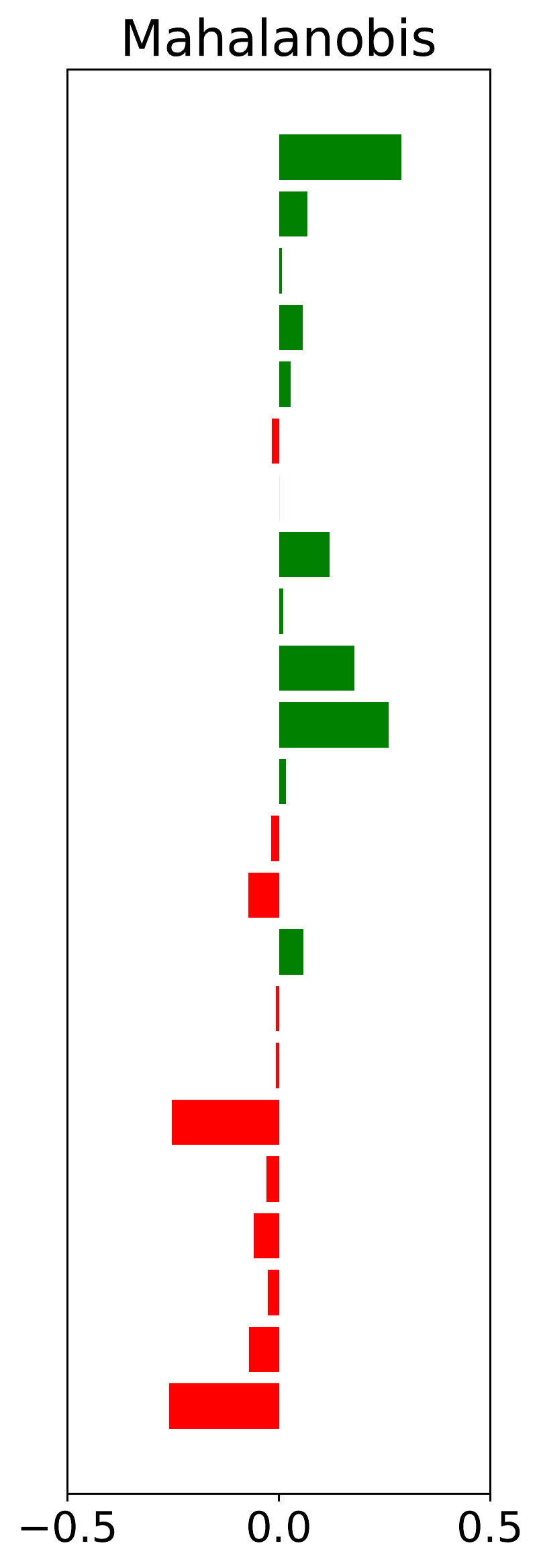} 
\caption{FOCUS explanations for the same model and same $x$ from the \textsc{HELOC} dataset based on different distance functions. 
Green and red indicate increases and decreases in feature values, respectively. 
Perturbation values are based on normalized feature values. 
Left: Euclidean explanation perturbs several features, but only slightly. 
Middle Left: Cosine explanation perturbs almost all of the features. 
Middle Right: Manhattan explanation perturbs two features substantially.
Right: Mahalanobis explanation perturbs almost all of the features. 
}
\label{fig:perturb-examples}
\end{figure*}

\subsection{Case Study: Credit Risk}
As a practical example, we investigate what FOCUS explanations look like for individuals in the \textsc{HELOC} dataset. 
Here, the task is to predict whether or not an individual will default on their loan. 
This has consequences for loan approval: individuals who are predicted as defaulting will be denied a loan. 
For these individuals, we want to understand how they can change their profile such that they are approved. 
Given an individual who has been denied a loan from a bank, a counterfactual explanation could be:
\begin{quote}
\textit{Your loan application has been denied. In order to have your loan application approved, you need to 
\begin{inparaenum}[(i)]
	\item increase your ExternalRiskEstimate score by 62, and
	\item decrease your NetFractionRevolvingBurden by 58.
\end{inparaenum}}
\end{quote}

\noindent%
Figure~\ref{fig:perturb-examples} shows four counterfactual explanations generated using different distance functions for the same individual and same model. 
We see that the Manhattan explanation only requires a few changes to the individual's profile, but the changes are large.
In contrast, the individual changes in the Euclidean explanation are smaller but there are more of them. 
In settings where there are significant dependencies between features, the Cosine explanations may be preferred since they are based on perturbations that try to preserve the relationship between features. 
For instance, in the \textsc{Wine Quality} dataset, it would be difficult to change the amount of citric acid without affecting the pH level. 
The Mahalanobis explanations would be useful when it is important to take into account not only correlations between features, but also the training data distribution. 
This flexibility allows users to choose what kind of explanation is best suited for their problem. 

Different distance functions can result in different \emph{magnitudes} of feature perturbations as well as different \emph{directions}. For example, the Cosine explanation suggests increasing \textit{PercentTradesWBalance}, while the Mahalanobis explanations suggests decreasing it. 
This is because the loss space of the underlying RF model is highly non-convex, and therefore there is more than one way to obtain an alternative prediction. When using complex models such as tree ensembles, there are no monotonicity guarantees. In this case, both options result in valid counterfactual examples. 

We examine the Manhattan explanation in more detail. 
We see that FOCUS suggests two main changes: 
\begin{inparaenum}[(i)]
	\item increasing the \textit{ExternalRiskEstimate}, and 
	\item decreasing the \textit{NetFractionRevolvingBurden}. 
\end{inparaenum}
We obtain the definitions and expected trends from the data dictionary created by the authors of the dataset. 
The \emph{ExternalRiskEstimate} is a ``consolidated version of risk markers'' (i.e., a credit score). 
A higher score is better: as one's \emph{ExternalRiskEstimate} increases, the probability of default decreases. 
The \textit{NetFractionRevolvingBurden} is the ``revolving balance divided by the credit limit'' (i.e., utilization). 
A lower value is better: as one's \emph{NetFractionRevolvingBurden} increases, the probability of default increases. 
We find that the changes suggested by FOCUS are fairly consistent with the expected trends in the data dictionary, as opposed to suggesting nonsensical changes such as increasing one's utilization to decrease the probability of default.

Decreasing one's utilization is heavily dependent on the specific situation: an individual who only supports themselves might have more control over their spending in comparison to someone who has multiple dependents. 
An individual can decrease their utilization in two ways: 
\begin{inparaenum}[(i)]
	\item decreasing their spending, or
	\item increasing their credit limit (or a combination of the two).
\end{inparaenum}
We can postulate that (i) is more ``actionable'' than (ii), since (ii) is usually a decision made by a financial institution. 
However, the degree to which an individual can actually change their spending habits is completely dependent on their specific situation: an individual who only supports themselves might have more control over their spending than someone who has multiple dependents. 
In either case, we argue that deciding what is (not) actionable is not a decision for the developer to make, but for the individual who is affected by the decision. 
Counterfactual examples should be used as part of a human-in-the-loop system and not as a final solution. 

The individual should know that utilization is an important component of the model, even if it is not necessarily ``actionable'' for them. 
We also note that it is unclear how exactly an individual would change their credit score without further insight into how the score was calculated (i.e., how the risk markers were consolidated).
It should be noted that this is not a shortcoming of FOCUS, but rather of using features that are uninterpretable on their own, such as credit scores.
Although FOCUS explanations cannot tell a user precisely how to increase their credit score, it is still important for the individual to know that their credit score is an important factor in determining their probability of getting a loan, as this empowers them to ask questions about how the score was calculated (i.e., how the risk markers were consolidated).

%!TEX root = ../main.tex

\section{Conclusion}
\label{section:focus-conclusion}
In this chapter, we propose an explanation method for tree-based classifiers, FOCUS, which casts the problem of finding counterfactual examples as a gradient-based optimization task and provides a differentiable approximation of tree-based models to be used in the optimization framework. 

Given an input instance $x$, FOCUS generates an optimal counterfactual example based on the minimal perturbation to the input instance $x$ which results in an alternative prediction from a model $f$. 
Unlike previous methods that assume the underlying classification model is differentiable, we propose a solution for when $f$ is a non-differentiable, tree-based model that provides a differentiable approximation of $f$, which can be used to find counterfactual examples using gradient-based optimization techniques.  

In the majority of experiments, examples generated by FOCUS are significantly closer to the original instances in terms of three different evaluation metrics compared to those generated by the baselines. 
FOCUS is able to generate valid counterfactual examples for all instances across all datasets, and the resulting explanations are flexible depending on the distance function.

This answers \textbf{\ref{rq:focus}}: we can generate counterfactual explanations for tree-based models using gradient-based optimization if we include differentiable approximations of tree-based models within the optimization framework. 
In the following chapter, we will investigate how to extend our method to accommodate different types of data such as graphs.

\section*{Reproducibility}
To facilitate the reproducibility of this work, our code is available at \url{https://github.com/a-lucic/focus}.

% !TEX root = ../thesis-main.tex

\acresetall

\chapter{Counterfactual Explanations for \aclp{GNN}}
%\chapter{Counterfactual Explanations for Graphs}
\label{chapter:research-cfgnn}

\footnote[]{This chapter was published at the International Conference on Artificial Intelligence and Statistics (AISTATS 2022) under the title ``CF-GNNExplainer: Counterfactual Explanations for Graph Neural Networks'' \citep{lucic2021cfgnnexplainer}.}
\acresetall

%\todo{Add intro paragraph that connects this chapter to its RQ from chapter 1: How can we extend our explanation method for tree-based models to graph-based models? first extend problem formalization to graphs, then apply the same gradient-based optimization technique as in FOCUS}

In the previous chapter, we developed a method for generating counterfactual explanations specific to tree-based models using gradient-based optimization techniques. 
In this chapter, we address the following research question:

\medskip
\noindent
\textbf{\ref{rq:cf-gnn}:} \acl{rq:cf-gnn}
\medskip

\noindent
Most existing methods for explaining predictions from graph neural networks (GNNs) are based on retrieving a subgraph of the original graph that is most relevant for the prediction. 
This differs from the counterfactual explanation problem where the task is to find the minimal perturbation to the original graph such that the prediction changes. 
The method we propose in this chapter is one of the first methods for generating counterfactual explanations for GNNs. 

The answer to \textbf{\ref{rq:cf-gnn}} is yes: we first extend the counterfactual explanation problem formalization to the graph data setting, then apply the same gradient-based optimization techniques as in the previous chapter. 
Our experimental results show that our algorithm can reliably generate minimal and accurate counterfactual explanations for GNNs.

%!TEX root = ../main.tex

\section{Introduction}
\label{section:cfgnn-introduction}
Advances in machine learning (ML) have led to breakthroughs in several areas of science and engineering,  ranging from computer vision, to natural language processing, to conversational assistants. 
Parallel to the increased performance of ML systems, there is an increasing call for the ``understandability'' of ML models ~\citep{goebel-2018-explainable}. 
Understanding \emph{why} an ML model returns a certain output in response to a given input is important for a variety of reasons such as model debugging, aiding decison-making, or fulfilling legal requirements \citep{gdpr}. 
Having certified methods for interpreting ML predictions will help enable their use across a variety of applications~\citep{miller-2017-explanations}.

Explainable artificial intelligence (XAI) refers to the set of techniques ``\textit{focused on exposing complex AI models to humans in a systematic and interpretable manner}''~\citep{samekexplainable}. A large body of work on XAI has emerged in recent years~\citep{guidotti-2018-survey,bodria2021benchmarking}. Counterfactual explanations are used to explain predictions of individual instances in the form: ``If X had been different, Y would not have occurred''~\citep{stepin2021survey,karimi_model-agnostic_2019,schut_generating_2021}. 
Counterfactual explanations are based on counterfactual examples: modified versions of the input sample that result in an alternative output (i.e., prediction). 
If the proposed modifications are also \emph{actionable}, this is referred to as achieving recourse \citep{ustun_actionable_2019,karimi2020survey}. 

To motivate our problem, consider an ML application for computational biology: drug discovery is a task that involves generating new molecules that can be used for medicinal purposes \citep{stokes_deep_2020,xie2021mars}. 
Given a candidate molecule, a GNN can predict if this molecule has a certain property that would make it effective in treating a particular disease \citep{wieder_compact_2020,guo2021fewshot,nguyen2020metalearning}.
%If the GNN predicts it does not have this desirable property, counterfactual explanations can help identify the minimal change required in order for the molecule to have the desirable property. 
%This could help us not only generate a new molecule that has this property, but also understand the molecular structures that contribute to this property.
If the GNN predicts it does not have this desirable property, counterfactual explanations can help identify the minimal change required such that the molecule is predicted to have this property. 
This could help not only inform the design of a new molecule that has this property, but also understand the molecular structures that contribute to this property.

Although GNNs have shown state-of-the-art results on tasks involving graph data \citep{zitnik_modeling_2018,deac_drug-drug_2019}, existing methods for explaining the predictions of GNNs have primarily focused on generating subgraphs that are relevant for a particular prediction~\citep{yuan2020explainability,baldassarre_explainability_2019,duval2021graphsvx,lin_causal_2021,luo_parameterized_2020,pope_explainability_2019,schlichtkrull_interpreting_2020,vu2020pgmexplainer,ying_gnnexplainer_2019,yuan_subgraph_2021}. 
However, \emph{none of these methods are able to identify the minimal subgraph automatically} -- they all require the user to specify the size of the subgraph, $S$, in advance. 
We show that even if we adapt existing methods to the counterfactual explanation problem, and try varying values for $S$, such methods are not able to produce valid, accurate counterfactual explanations, and are therefore not well-suited to solve the counterfactual explanation problem. 
To address this gap, we propose CF-GNNExplainer, a method for generating counterfactual explanations for GNNs. 

Similar to other counterfactual methods for tabular or image data proposed in the literature~\citep{verma2020counterfactual, karimi2020survey}, CF-GNNExplainer works by perturbing input data at the instance-level. 
Unlike previous methods, CF-GNNExplainer can generate counterfactual explanations for graph data. 
In particular, our method iteratively removes edges from the original adjacency matrix based on matrix sparsification techniques, keeping track of the perturbation that leads to a change in prediction, and returning the perturbation with the smallest change w.r.t.\ the number of edges. 

\begin{figure}[t]
    \centering
    \includegraphics[width=\columnwidth]{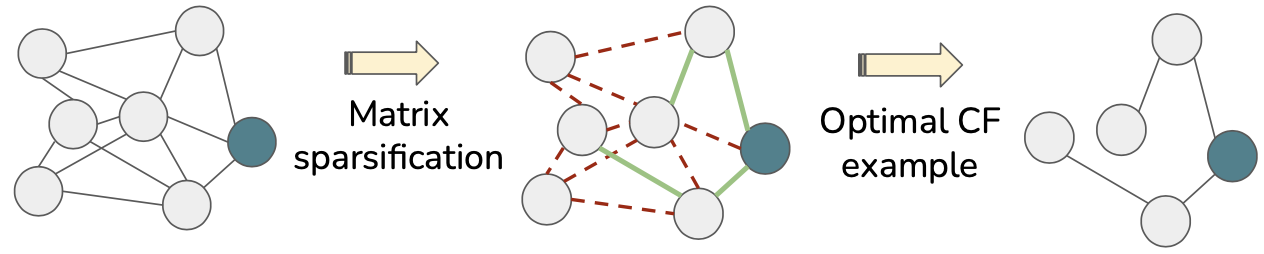}
    \caption{Intuition of counterfactual example generation by CF-GNNExplainer.}
    \label{fig:visual}
\end{figure}

\pagebreak

We evaluate CF-GNNExplainer on three public datasets for GNN explanations and measure its effectiveness using four metrics: fidelity, explanation size, sparsity, and accuracy. We find that CF-GNNExplainer is able to generate counterfactual examples with at least 94\% accuracy, while removing fewer than 3 edges on average. 
We make the following contributions:
\begin{enumerate}[(1)]
    \item We formalize the problem of generating counterfactual explanations for GNNs (Section~\ref{section:problem-formulation}). 
    \item We propose CF-GNNExplainer, a novel method for explaining predictions from GNNs (Section~\ref{section:cfgnn-method}). 
    \item We propose an experimental setup for holistically evaluating counterfactual explanations for GNNs (Section~\ref{section:cfgnn-experimental-setup}).
\end{enumerate}

%!TEX root = ../main.tex

\section{Related Work}
\label{section:focus-relatedwork}
Based on the taxonomy described in Chapter~\ref{chapter:introduction}, our setting in this chapter is a \emph{local explanation} problem for \emph{neural networks}, specifically GNNs. 
We use \emph{sensitivity analysis}, specifically counterfactual perturbations, on \emph{graph} data to generate our explanations. 
Since our work is a counterfactual XAI approach for GNNs, it is related to GNN explainability (Section~\ref{section:rw-gnnxai}) as well as counterfactual explanations (Section~\ref{section:rw-cfexp}). It is also related to adversarial attack methods (Section~\ref{section:rw-adversarial}).

\subsection{GNN Explainability}
\label{section:rw-gnnxai}
Several GNN XAI approaches have been proposed -- a recent survey of the most relevant work is presented by~\citet{yuan2020explainability}.
However, unlike our work, {\em none} of the methods in this survey generate counterfactual explanations. 

The majority of existing GNN XAI methods provide an explanation in the form of a subgraph of the original graph that is deemed to be important for the prediction \citep{yuan2020explainability,baldassarre_explainability_2019,duval2021graphsvx,lin_causal_2021,luo_parameterized_2020,pope_explainability_2019,schlichtkrull_interpreting_2020,vu2020pgmexplainer,ying_gnnexplainer_2019,yuan_subgraph_2021}. We refer to these as \emph{subgraph-generating methods}. 
Such methods are analogous to popular XAI methods such as LIME \citep{ribeiro-2016-should} or SHAP \citep{lundberg_unified_2017}, which identify relevant features for a particular prediction for tabular, image, or text data. 
All of these methods require the user to specify the size of the explanation, $S$, in advance: the number of features (or edges) to keep. 
In contrast, CF-GNNExplainer generates counterfactual explanations, which can find the size of the explanation without requiring input from the user. 
Although both types of techniques are meant for explaining GNN predictions, they are solving fundamentally different problems: counterfactual explanations generate the minimal perturbation such that the prediction changes, while subgraph-retrieving methods identify a relevant (and not necessarily minimal) subgraph that matches the original prediction. 

\pagebreak

The work by \citet{kang_explaine_2019} also generates counterfactual examples for GNNs, but they focus on a different task: link prediction. 
Other GNN XAI methods identify important node features~\citep{huang_graphlime_2020} or similar examples~\citep{faber_contrastive_2020}. 
The works of \citet{yuan_xgnn_2020} and \citet{schnake_xai_2020} generate model-level (i.e., global) explanations for GNNs, which differs from our work since we produce instance-level (i.e., local) explanations. 

\subsection{Counterfactual Explanations}
\label{section:rw-cfexp}
There exists a substantial body of work on counterfactual explanations for tabular, image, and text data \citep{verma2020counterfactual,karimi2020survey,stepin2021survey}.
Some methods treat the underlying classification model as a black-box \citep{laugel_inverse_2017, guidotti_local_2018,lucic_2020_why}, whereas others make use of the model's inner workings \citep{tolomei_interpretable_2017, wachter_counterfactual_2017, ustun_actionable_2019, kanamori_dace_2020, lucic2020focus}.
All of these methods are based on perturbing feature values to generate counterfactual examples -- they are not equipped to handle graph data with relationships (i.e., edges) between instances (i.e., nodes). In contrast, CF-GNNExplainer provides counterfactual examples specifically for graph data.

\subsection{Adversarial Attacks}
\label{section:rw-adversarial}
Counterfactual examples are also related to adversarial attacks \citep{sun2020adversarial}: they both represent instances obtained from minimal perturbations to the input, which induce changes in the prediction made by the learned model. 
One difference between the two is in the intent: adversarial examples are meant to fool the model, while counterfactual examples are meant to explain the prediction \citep{freiesleben_intriguing_2021,lucic2020focus}. 
In the context of graph data, adversarial attack methods typically make minimal perturbations to the \emph{overall graph} with the intention of degrading overall model performance, as opposed to attacking individual nodes. 
In contrast, we are interested in generating counterfactual examples for individual nodes, as opposed to identifying perturbations to the overall graph. 
We confirm that the counterfactual examples produced by CF-GNNExplainer are informative and not adversarial by measuring the accuracy of our method (see Section~\ref{section:cfgnn-metrics}).

%!TEX root = ../main.tex

\section{Background}
\label{section:background}
In this section, we provide background information on GNNs (Section~\ref{section:gnns-general}) and matrix sparsification (Section~\ref{section:matrix-sparsification}), both of which are necessary for understanding CF-GNNExplainer. 

\subsection{Graph Neural Networks}
\label{section:gnns-general}
Graphs are structures that represent a set of entities (nodes) and their relations (edges). 
GNNs operate on graphs to produce representations that can be used in downstream tasks such as graph or node classification. The latter is the focus of this work.
We refer to the survey papers by \citet{battaglia_relational_2018} and \citet{chami2021machine} for an overview of existing GNN methods. 

Let $f(A, X; W) \rightarrow y$ be any GNN, where $y$ is the set of possible predicted classes, $A$ is an $n \times n$ adjacency matrix, $X$ is an $n \times p$ feature matrix, and $W$ is the learned weight matrix of $f$. 
In other words, $A$ and $X$ are the inputs of $f$, and $f$ is parameterized by $W$. 

A node's representation is learned by iteratively updating the node's features based on its neighbors' features.   
The number of layers in $f$ determines which neighbors are included: if there are $\ell$ layers, then the node's final representation only includes neighbors that are at most $\ell$ hops away from that node in the graph $\graph$. 
The rest of the nodes in $\graph$ are not relevant for the computation of the node's final representation.  
We define the \emph{\cgraph{}} of a node $v$ as the set of the nodes and edges relevant for the computation of $f(v)$ (i.e., those in the $\ell$-hop neighborhood of $f$), represented as a tuple: $\compgraph = (\compadj, \compnode)$, where $\compadj$ is the subgraph adjacency matrix and $X_v$ is the node feature matrix for nodes that are at most $\ell$ hops away from $v$. We then define a node $v$ as a tuple of the form $v = (\compadj, x)$, where $x$ is the feature vector for $v$. 

\subsection{Matrix Sparsification}
\label{section:matrix-sparsification}
CF-GNNExplainer uses matrix sparsification to generate counterfactual examples, inspired by~\citet{srinivas_training_2016}, who propose a method for training sparse neural networks. 
Given a weight matrix $W$, a binary sparsification matrix is learned which is multiplied element-wise with $W$ such that some of the entries in $W$ are zeroed out. 
In the work by \citet{srinivas_training_2016}, the objective is to remove entries in the weight matrix in order to reduce the number of parameters in the model. 
In our case, we want to \emph{zero out entries in the adjacency matrix} (i.e., remove edges) in order to generate counterfactual explanations for GNNs. 
That is, we want to remove the important edges -- those that are crucial for the prediction. 
%!TEX root = ../main.tex

\section{Problem Formulation}
\label{section:problem-formulation}
In general, a counterfactual example $\bar{x}$ for an instance $x$ according to a trained classifier $f$ is found by perturbing the features of $x$ such that $f(x) \neq f(\bar{x})$ \citep{wachter_counterfactual_2017}. 
An optimal counterfactual example $\bar{x}^*$ is one that minimizes the distance between the original instance and the counterfactual example, according to some distance function $d$. 
The resulting optimal counterfactual explanation is therefore $\Delta^*_{x} = \bar{x}^* - x$ \citep{lucic2020focus}. 

For graph data, it may not be enough to simply perturb node features, especially since they are not always available. 
This is why we are interested in generating counterfactual examples by perturbing the graph structure instead. 
In other words, we want to change the relationships between instances (i..e, nodes), rather than change the instances themselves. 
Therefore, a counterfactual example for graph data has the form $\bar{v} = (\bar{\compadj}, x)$, where $x$ is the feature vector and $\bar{\compadj}$ is a perturbed version of  $\compadj$, the adjacency matrix of the subgraph neighborhood of a node $v$. $\bar{\compadj}$ is obtained by removing some edges from $\compadj$, such that $f(v) \neq f(\bar{v})$. 
Following \citet{wachter_counterfactual_2017} and \citet{lucic2020focus}, we generate counterfactual examples by minimizing a loss function of the form:
\begin{align}
\label{eq:loss-graph}
    \mathcal{L} = \losspred(v, \bar{v} \mid f, g) + \beta \lossdist(v, \bar{v} \mid d),
\end{align}
where $v$ is the original node, $f$ is the original model, $g$ is the counterfactual model that generates $\bar{v}$, and $\losspred$ is a prediction loss that encourages $f(v) \neq f(\bar{v})$. 
$\lossdist$ is a distance loss that encourages $\bar{v}$ to be close to $v$, and $\beta$ controls how important $\lossdist$ is compared to $\losspred$. 
We want to find $\bar{v}^*$ that minimizes Equation~\ref{eq:loss-graph}: this is the optimal counterfactual example for $v$.

%!TEX root = ../main.tex

\section{Method: CF-GNNExplainer}
\label{section:cfgnn-method}

To solve the problem defined in Section~\ref{section:problem-formulation}, we propose CF-GNNExplainer, which generates $\bar{v} = (\bar{\compadj}, x)$ given a node $v = (\compadj, x)$.
Our method can operate on any GNN model $f$. 
To illustrate our method and avoid cluttered notation, let $f$ be a standard, one-layer Graph Convolutional Network (GCN)~\citep{kipf_semi_supervised_2017} for node classification:
\begin{align}
    \label{eq:gcn}
    f(A, X; W) = \softmax\left[\tilde{D}^{-1/2} \tilde{A} \tilde{D}^{-1/2} X W \right], 
\end{align}
where $\tilde{A} = A + I$, $I$ is the identity matrix, $\tilde{D}_{ii} = \sum_j \tilde{A}_{ij}$ are entries in the degree matrix $\tilde{D}$, $X$ is the node feature matrix, and $W$ is the weight matrix \citep{kipf_semi_supervised_2017}.

\subsection{Adjacency Matrix Perturbation}
\label{section:adjacency-perturb}
First, we define $\bar{\compadj} = \perturbb \odot \compadj$, where $\perturbb$ is a binary perturbation matrix that sparsifies $\compadj$. 
Our aim is to find $\perturbb$ for a given node $v$ such that $f(\compadj, x) \neq f(\perturbb \odot \compadj, x$). 
To find $\perturbb$, we build upon the method by \citet{srinivas_training_2016} for training sparse neural networks (see Section~\ref{section:matrix-sparsification}), where our objective is to zero out entries in the adjacency matrix (i.e., remove edges).
That is, we want to find $\perturbb$ that minimally perturbs $\compadj$, and use it to compute $\bar{\compadj} = \perturbb \odot \compadj$. 
If an element $\perturbb_{{i,j}} = 0$, this results in the deletion of the edge between node $i$ and node $j$. 
When $\perturbb$ is a matrix of ones, this indicates that all edges in $\compadj$ are used in the forward pass. 

Similar to the work by \citet{srinivas_training_2016}, we first generate an intermediate, real-valued matrix $\perturbl$ with entries in $\left[0, 1\right]$, apply a sigmoid transformation, then threshold the entries to arrive at a binary $\perturbb$: entries greater than or equal to 0.5 become 1, while those below 0.5 become 0. 
In the case of undirected graphs (i.e., those with symmetric adjacency matrices), we first generate a perturbation vector, which we then use to populate $\perturbl$ in a symmetric manner, instead of generating $\perturbl$ directly. 

\pagebreak

\subsection{Counterfactual Generating Model}
We want our perturbation matrix $\perturbb$ to only act on $\compadj$, not $\tilde{\compadj}$, in order to preserve self-loops in the message passing of $f$. This is because we always want a node representation update to include its own representation from the previous layer. 
Therefore we first rewrite Equation~\ref{eq:gcn} for our illustrative one-layer case to isolate $\compadj$: 
\begin{align}
f(\compadj, \compnode; W) = \softmax\left[(\compdeg + I)^{-1/2} (\compadj + I) (\compdeg + I)^{-1/2} \compnode W\right].
    \label{eq:gcn3}
\end{align}
To generate CFs, we propose a new function $g$, which is based on $f$, but it is parameterized by $\perturbb$ instead of $W$. 
We update the degree matrix $\compdeg$ based on $\perturbb \odot \compadj$, add the identity matrix to account for self-loops (as in $\tilde{\compdeg}$ in Equation~\ref{eq:gcn}), and call this $\bar{\compdeg}$: 
\begin{align}
    \label{eq:cf}
    g(\compadj, \compnode, W; \perturbb) = \softmax\left[\bar{\compdeg}^{-1/2} (\perturbb \odot \compadj + I) \bar{\compdeg}^{-1/2} \compnode W\right].
\end{align}
In other words, $f$ learns the weight matrix while holding the data constant, while $g$ 
%is optimized to find a perturbation matrix that is then used to 
generates new data points (i.e., counterfactual examples) while holding the weight matrix (i.e., model) constant. 
Another distinction between $f$ and $g$ is that the aim of $f$ is to find the optimal set of weights that generalizes well on an unseen test set, while the objective of $g$ is to generate an optimal counterfactual example, given a particular node (i.e., $\bar{v}$ is the output of $g$).

\subsection{Loss Function Optimization}
We generate $\perturbb$ by minimizing Equation~\ref{eq:loss-graph}, adopting the negative log-likelihood (NLL) loss  for $\losspred$:
\begin{align}
    &\losspred(v, \bar{v}|f, g) =\mathbbm{1}\left[f(v) = f(\bar{v})\right] \cdot \mathcal{L}_{NLL}(f(v), g(\bar{v})).
    \label{eq:loss-pred}
\end{align}
Since we do not want $f(\bar{v})$ to match $f(v)$, we put a negative sign in front of $\losspred$ and include an indicator function to ensure the loss is active as long as $f(\bar{v}) = f(v)$. 
Note that $f$ and $g$ have the same weight matrix $W$ -- the main difference is that $g$ also includes the perturbation matrix $\perturbb$.

$\lossdist$ can be based on any differentiable distance function. 
In our case, we take $d$ to be the element-wise difference between $v$ and $\bar{v}$, corresponding to the difference between $\compadj$ and $\bar{\compadj}$: the number of edges removed. 
For undirected graphs, we divide this value by 2 to account for the symmetry in the adjacency matrices. 
When updating $\perturbb$, we take the gradient of Equation~\ref{eq:loss-graph} with respect to the intermediate $\perturbl$, \emph{not} the binary $\perturbb$. 

\pagebreak

\begin{algorithm}[t]
    \caption{CF-GNNExplainer: given a node $v = (\compadj, x)$ where $f(v) = y$, generate the minimal perturbation, $\bar{v} = (\bar{\compadj}, x)$, such that $f(\bar{v}) \neq y$.}
    
    \label{alg:cf-gnnexplainer}

    \begin{algorithmic}
        \STATE {\bfseries Input:} node $v = (\compadj, x)$, trained GNN model $f$, CF model $g$, loss function $\loss$, learning rate $\alpha$, number of iterations $K$, distance function $d$. 
        
        \STATE
        
        % \STATE $y$ = \textsc{get\_gnn\_prediction($v$)} 
        \STATE $f(v) = y$ \qquad \textcolor{gray}{\textit{\# Get GNN prediction} } 
        
        % \STATE $\compadj, \compdeg \gets normalize(v)$

        % \STATE \textcolor{gray}{\textit{\# Initialization} }

        \STATE $\perturbl \gets J_n$ \qquad  \textcolor{gray}{\textit{\# Initialization} } 
        
        \STATE $\bar{v}^* = \left[ \:\right]$
        % \STATE $\bar{\compgraph} \gets \perturbl \odot \compadj$ 
        % \STATE $\bar{\compgraph} \gets \bar{\compgraph}$
        
        \STATE
        
        \FOR{$K$ iterations} \label{line:start_loop}
            \STATE $\bar{v}$ = \textsc{get\_cf\_example()}
            %\STATE \textsc{compute\_loss($v$, $\bar{v}$)}
            \STATE $\loss \gets \loss(v, \bar{v}, f, g)$ \qquad \textcolor{gray}{\textit{\# Eq~\ref{eq:loss-graph} \& ~\ref{eq:loss-pred}} } 
            %\STATE \textsc{update\_p()}
            \STATE $\perturbl \gets \perturbl + \alpha \nabla_{\perturbl} \loss$ \qquad \textcolor{gray}{\textit{\# Update $\perturbl$}} 
        \ENDFOR
        
        \STATE

        \STATE \textbf{Function} \textsc{get\_cf\_example()}
        
       \bindent
        
        \STATE $\perturbb \gets \text{threshold}(\sigma(\perturbl))$
        
        \STATE $\bar{\compadj} \gets \perturbb \odot \compadj$
        \STATE $\bar{v}_{cand} \gets (\bar{\compadj}, x)$
        
        \IF{$f(v) \neq f(\bar{v}_{cand})$} 
            \STATE $\bar{v} \gets \bar{v}_{cand}$
            
            \IF{not $\bar{v}^*$}
            	\STATE $\bar{v}^* \gets \bar{v}$ \qquad \textcolor{gray}{\textit{\# First CF}} 
            
            \ELSIF{$d(v, \bar{v}) \leq d(v, \bar{v}^*)$}
            
                \STATE $\bar{v}^* \gets \bar{v}$ \qquad \textcolor{gray}{\textit{\# Best CF}} 
            \ENDIF
                    
        \ENDIF
        
        \RETURN{$\bar{v}^*$}
        
        \eindent
        
        % \STATE
        
        % \RETURN $\bar{v}^*$
        
    \end{algorithmic}
\end{algorithm}

\subsection{CF-GNNExplainer }

We call our method CF-GNNExplainer and summarize its details in Algorithm~\ref{alg:cf-gnnexplainer}. 
Given an node in the test set $v$, we first obtain its original prediction from $f$ and initialize ${\perturbl}$ as a matrix of ones, $J_n$, to initially retain all edges. 
Next, we run CF-GNNExplainer for $K$ iterations.  
To find a counterfactual example, we use Equation~\ref{eq:cf}. 

First, we compute $\perturbb$ by thresholding $\perturbl$ (see Section~\ref{section:adjacency-perturb}). 
Then we use $\perturbb$ to obtain the sparsified adjacency matrix that gives us a candidate counterfactual example, $\bar{v}_{cand}$. 
This example is then fed to the original GNN, $f$, and if $f$ predicts a different output than for the original node, we have found a valid counterfactual example, $\bar{v}$. 

We keep track of the ``best'' counterfactual example (i.e., the most minimal according to $d$), and return this as the optimal counterfactual example $\bar{v}^*$ after $K$ iterations. 
Between iterations, we compute the loss following Equations~\ref{eq:loss-graph} and \ref{eq:loss-pred}, and update $\perturbl$ based on the gradient of the loss. 
In the end, we retrieve the optimal counterfactual explanation $\Delta_v^* = v - \bar{v}^* $. 

\subsection{Complexity}
\label{section:complexity}
CF-GNNExplainer has time complexity $O(KN^2)$, where $N$ is the number of nodes in the subgraph neighbourhood and $K$ is the number of iterations. We note that high complexity is common for local XAI methods (i.e., SHAP \citep{lundberg_unified_2017}, GNNExplainer \citep{ying_gnnexplainer_2019}, etc.), but in practice, one typically only generates explanations for a subset of the dataset.

%!TEX root = ../main.tex

\section{Experimental Setup}
\label{section:cfgnn-experimental-setup}

In this section, we outline our experimental setup for evaluating CF-GNNExplainer, including the datasets and models used (Section~\ref{section:cfgnn-datasets}), the baselines we compare against (Section~\ref{section:baselines}), the evaluation metrics (Section~\ref{section:cfgnn-metrics}), and the hyperparameter search method (Section~\ref{section:hyperparams}). 
In total, we run approximately 375 hours of experiments on one Nvidia TitanX Pascal GPU with access to 12GB RAM.

\subsection{Datasets and Models}
\label{section:cfgnn-datasets}
Given the challenges associated with defining and evaluating the accuracy of XAI methods~\citep{doshi-2017-towards}, we first focus on synthetic tasks where we know the ground-truth explanations. 
Although there exist real graph classification datasets with ground-truth explanations~\citep{mutag_dataset}, there do not exist any real node classification datasets with ground-truth explanations, which is the task we focus on in this chapter. 
Building such a dataset would be an excellent contribution, but is outside the scope of this paper.

In our experiments, we use the \synfour{}, \textsc{tree-grids}, \synone{} datasets from the work by \citet{ying_gnnexplainer_2019}. 
These datasets were created specifically for the task of explaining node classification predictions from GNNs. 
Each dataset consists of (i) a base graph, (ii) motifs that are attached to random nodes of the base graph, and (iii) additional edges that are randomly added to the overall graph. 
They are all undirected graphs. 
The classification task is to determine whether or not the nodes are part of the motif. 
The purpose of these datasets is to have a ground-truth for the ``correctness'' of an explanation: for nodes in the motifs, the explanation is the motif itself \citep{luo_parameterized_2020}. 
The dataset statistics are available in Table~\ref{table:stats}.

\synfour{} consists of a binary tree base graph with 6-cycle motifs, \textsc{tree-grids} also has a binary tree as its base graph, but with 3$\times$3 grids as the motifs. 
For \synone{}, the base graph is a Barabasi-Albert (BA) graph with house-shaped motifs, where each motif consists of 5 nodes (one for the top of the house, two in the middle, and two on the bottom). 
Here, there are four possible classes (not in motif, in motif: top, middle, bottom). 
We note that compared to the other two datasets, the \synone{} dataset is much more densely connected -- the node degree is more than twice as high as that of the \synfour{} or \synfive{} datasets, and the average number of nodes and edges in each node's computation graph is order(s) of magnitude larger. 
We use the same experimental setup (i.e., dataset splits, model architecture) as \citet{ying_gnnexplainer_2019} to train a 3-layer GCN (hidden size = 20) for each task. 
Our GCNs have at least 87\% accuracy on the test set.

\begin{table}[]
\caption{Dataset statistics. The \# edges in the motif indicates the size of the ground truth (GT) explanation. }
%\todo{Double check notation: use commas for bigger numbers or not.}		% NO commas!
\label{table:stats}
\centering
\begin{tabular}{lrrr}
\toprule
                          & \multicolumn{1}{c}{\textsc{Tree}}   & \multicolumn{1}{c}{\textsc{Tree}} & \multicolumn{1}{c}{\textsc{BA}}     \\
                          & \multicolumn{1}{c}{\textsc{Cycles}} & \multicolumn{1}{c}{\textsc{Grid}} & \multicolumn{1}{c}{\textsc{Shapes}} \\ 
\midrule
\# classes                 & 2                         & 2               & 4                          \\ 

\# nodes in motif                 & 6                        & 9            & 5                        \\
\# edges in motif      (GT)            & 6                       & 12             & 6                \\
\midrule
\# nodes in total                  & 871                        & 1231            & 700                        \\
\# edges in total                  & 1950                       & 3410             & 4100                \\

\midrule
Avg node degree           & 2.27                       & 2.77                     & 5.87                       \\
Avg \# nodes in $\compadj$ & 19.12                      & 30.69                    & 304.40                     \\
Avg \# edges in $\compadj$ & 18.99                      & 33.94                    & 1106.24                    \\
\bottomrule
\end{tabular}
\end{table}

\subsection{Baselines}
\label{section:baselines}
Since existing GNN XAI methods give explanations in the form of relevant subgraphs as opposed to counterfactual examples, it is not straightforward to identify baselines for our experiments that ensure a fair comparison between methods. 
To evaluate CF-GNNExplainer, we compare against 4 baselines: \baserand{}, \basekeep{}, \baserm{}, and \gnnexplainer{}.
The random perturbation is meant as a sanity check. 
We randomly initialize the entries of $\perturbl \in \left[-1, 1\right]$ and apply the same sigmoid transformation and thresholding as described in Section~\ref{section:adjacency-perturb}. 
We repeat this $K$ times and keep track of the most minimal perturbation resulting in a counterfactual example. 
Next, we compare against baselines that are based on the ego graph of $v$ (i.e., its 1-hop neighbourhood): \basekeep{} keeps all edges in the ego graph of $v$, while \baserm{} removes all edges in the ego graph of $v$. 

Our fourth baseline is based on \gnnexplainer{} by \citet{ying_gnnexplainer_2019}, which identifies the $S$ most relevant edges for the prediction (i.e., the most relevant subgraph of size $S$). 
To generate counterfactual explanations, we remove the subgraph generated by \gnnexplainer{}. 
We include this method in our experiments in order to have a baseline based on a prominent GNN XAI method, but we note that subgraph-retrieving methods like \gnnexplainer{} are not meant for generating counterfactual explanations. 
Unlike our method, \gnnexplainer{} cannot automatically find a \emph{minimal} subgraph and therefore requires the user to determine the number of edges to keep in advance (i.e., the value of $S$). 
As a result, we cannot evaluate how minimal its counterfactual explanations are, but we can compare it against our method in terms of 
\begin{inparaenum}[(i)]
	\item its ability to generate valid counterfactual examples, and 
	\item how accurate those counterfactual examples are.
\end{inparaenum}
We report results on \gnnexplainer{} for $ S \in \{1, 2, 3, 4, 5,$ GT$\}$, where GT is the size of the ground truth explanation (i.e., the number of edges in the motif, see Table~\ref{table:stats}).

\subsection{Metrics}
\label{section:cfgnn-metrics}
We generate a counterfactual example for each node in the graph separately and evaluate in terms of four metrics.

\medskip \noindent
\textbf{Fidelity:} is defined as the proportion of nodes where the original predictions match the prediction for the explanations \citep{molnar2019,ribeiro-2016-should}. Since we generate counterfactual examples, we do not want the original prediction to match the prediction for the explanation, so we want a low value for fidelity. 

\medskip \noindent
\textbf{Explanation Size:} is the number of removed edges. It corresponds to the $\lossdist$ term in Equation~\ref{eq:loss-graph}: the difference between the original $\compadj$ and the counterfactual $\bar{\compadj}$. Since we want to have \emph{minimal} explanations, we want a small value for this metric. Note that we cannot evaluate this metric for \gnnexplainer{}. 

\medskip \noindent
\textbf{Sparsity:} measures the proportion of edges in $\compadj$ that are removed \citep{yuan2020explainability}. A value of 0 indicates all edges in $\compadj$ were removed. Since we want \emph{minimal} explanations, we want a value close to 1. Note that we cannot evaluate this metric for \gnnexplainer{}.

\medskip \noindent
\textbf{Accuracy:} is the mean proportion of explanations that are ``correct''. Following the work by \citet{ying_gnnexplainer_2019, luo_parameterized_2020}, we only compute accuracy for nodes that are originally predicted as being part of the motifs, since accuracy can only be computed on instances for which we know the ground truth explanations. 
Given that we want \emph{minimal} explanations, we consider an explanation to be correct if it \emph{exclusively} involves edges that are inside the motifs (i.e., only removes edges that are within the motifs). 

%The exact calculations of all metrics can be found in the public code base at \url{https://github.com/cf-gnnexplainer}. 

\subsection{Hyperparameter Search}
\label{section:hyperparams}
We experiment with different optimizers and hyperparameter values for the number of iterations $K$, the trade-off parameter $\beta$, the learning rate $\alpha$, and the Nesterov momentum $m$ (when applicable). 
We choose the setting that produces the most counterfactual examples. 
We test the number of iterations $K \in \{100, 300, 500\}$, the trade-off parameter $\beta \in \{0.1, 0.5\}$, the learning rate $\alpha \in \{0.005, 0.01, 0.1, 1\}$, and the Nesterov momentum $m \in \{0, 0.5, 0.7, 0.9\}$. 
We test Adam, SGD and AdaDelta as optimizers. 
We find that for all three datasets, the SGD optimizer gives the best results, with $k = 500$, $\beta = 0.5$, and $\alpha = 0.1$. 
For the \synfour{} and \synfive{} datasets, we set $m = 0$, while for the \synone{} dataset, we use $m = 0.9$.

%!TEX root = ../main.tex

\section{Results}
\label{section:cfgnn-results}

We evaluate CF-GNNExplainer in terms of the metrics outlined in Section~\ref{section:cfgnn-metrics}. 
The results are shown in Table~\ref{table:results1} and Table~\ref{table:results-gnnexplainer}.  
In cases where the baselines outperform CF-GNNExplainer on a particular metric, they perform poorly on the rest of the metrics, or on other datasets.

\subsection{Main Findings}

\textbf{Fidelity:}
CF-GNNExplainer outperforms \basekeep{} across all three datasets, and outperforms \baserm{} for \synfour{} and \synfive{} in terms of fidelity. 
We find that \baserand{} has the lowest fidelity in all cases -- it is able to find counterfactual examples for every single node. 
In the following subsections, we will see that this corresponds to poor performance on the other metrics.

\begin{table*}[h]
\centering
\caption{Results comparing our method (abbreviated as \OurShort{}) to \baserand{}, \basekeep{}, and \baserm{}. Below each metric, $\blacktriangledown$ indicates a low value is desirable, while $\blacktriangle$ indicates a high value is desirable.}
\label{table:results1}
\setlength{\tabcolsep}{4pt}
\scriptsize{
\begin{tabular}{lrrrr rrrr rrrr}
\toprule
\multicolumn{1}{c}{} & \multicolumn{4}{c}{\synfour{}}                                                                                                                 & \multicolumn{4}{c}{\synfive{}}                                                                                                                   & \multicolumn{4}{c}{\synone{}}                                                                                                                  \\ 
\cmidrule(r){2-5}\cmidrule(r){6-9}\cmidrule{10-13} 
               & \multicolumn{1}{c}{Fid.} & \multicolumn{1}{c}{Size} & \multicolumn{1}{c}{Spar.} & \multicolumn{1}{c}{Acc.} & \multicolumn{1}{c}{Fid.} & \multicolumn{1}{c}{Size} & \multicolumn{1}{c}{Spar.} & \multicolumn{1}{c}{Acc.} & \multicolumn{1}{c}{Fid.} & \multicolumn{1}{c}{Size} & \multicolumn{1}{c}{Spar.} & \multicolumn{1}{c}{Acc.} \\

% & \multicolumn{1}{c}{$\downarrow$} &\multicolumn{1}{c}{$\downarrow$} &\multicolumn{1}{c}{$\uparrow$} & \multicolumn{1}{c}{$\uparrow$} & \multicolumn{1}{c}{$\downarrow$} &\multicolumn{1}{c}{$\downarrow$} &\multicolumn{1}{c}{$\uparrow$} & \multicolumn{1}{c}{$\uparrow$} & \multicolumn{1}{c}{$\downarrow$} &\multicolumn{1}{c}{$\downarrow$} &\multicolumn{1}{c}{$\uparrow$} & \multicolumn{1}{c}{$\uparrow$} \\

Method & \multicolumn{1}{c}{$\blacktriangledown$} &\multicolumn{1}{c}{$\blacktriangledown$} &\multicolumn{1}{c}{$\blacktriangle$} & \multicolumn{1}{c}{$\blacktriangle$} & \multicolumn{1}{c}{$\blacktriangledown$} &\multicolumn{1}{c}{$\blacktriangledown$} &\multicolumn{1}{c}{$\blacktriangle$} & \multicolumn{1}{c}{$\blacktriangle$} & \multicolumn{1}{c}{$\blacktriangledown$} &\multicolumn{1}{c}{$\blacktriangledown$} &\multicolumn{1}{c}{$\blacktriangle$} & \multicolumn{1}{c}{$\blacktriangle$} \\
\midrule
\baserand{}               & \textbf{0.00}                     & 4.70                              & 0.79                                & 0.63                               & \textbf{0.00}                     & 9.06                              & 0.75                                & 0.77                               & \textbf{0.00}                     & 503.31                            & 0.58                                & 0.17                              \\
\basekeep{}                 & 0.32                              & 15.64                             & 0.13                                & 0.45                               & 0.32                              & 29.30                             & 0.09                                & 0.72                               & 0.60                              & 504.18                            & 0.05                                & 0.18                              \\
\baserm{}              & 0.46                              & 2.11                              & 0.89                                & ---                                  & 0.61                              & 2.27                              & 0.92                                & ---                                  & 0.21                              & 10.56                             & 0.97                                & \textbf{0.99}                     \\

% \gnnexpshort{} ($S=$ GT) &  0.55 &	6.00 & 0.57 &	0.46 &	0.35 &	11.83 &	0.53 &	0.74 &	0.82 &	6.00 &	0.79 &	0.24    \\

\midrule
CF-GNN              & 0.21                              & \textbf{2.09}                     & \textbf{0.90}                       & \textbf{0.94}                      & 0.07                              & \textbf{1.47}                     & \textbf{0.94}                       & \textbf{0.96}                      & 0.39                              & \textbf{2.39}                     & \textbf{0.99}                       & 0.96                 \\
\bottomrule
\end{tabular}
}
\end{table*}

\begin{figure*}[]

    \centering

    \includegraphics[scale=0.27]{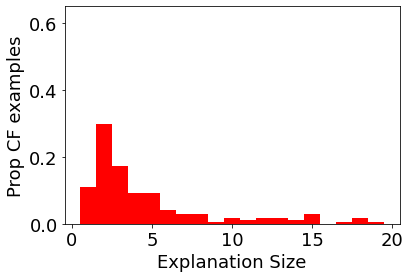}
    \includegraphics[scale=0.27]{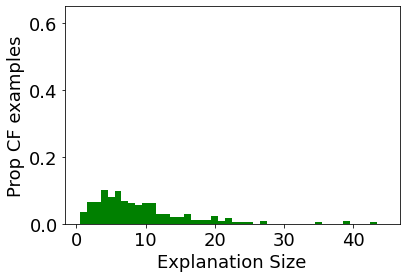}
    \includegraphics[scale=0.27]{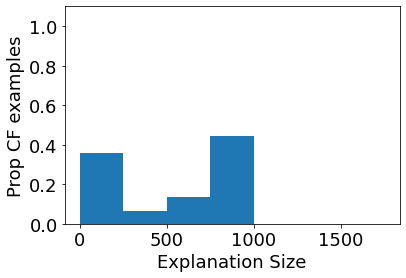}
    
        \caption{Histograms showing the proportion of counterfactual examples that have a certain explanation size from \baserand{}. Note the $x$-axis for \synone{} goes up to 1500. Left: \synfour{}, Middle: \synfive{}, Right: \synone{}.  }
        \label{fig:random-explanation-size}
        \bigskip \bigskip
        
    \includegraphics[scale=0.27]{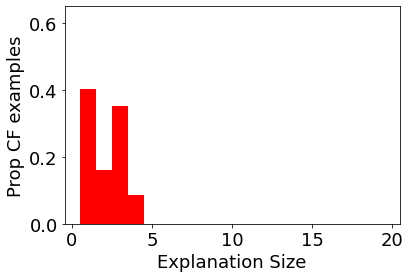}
    \includegraphics[scale=0.27]{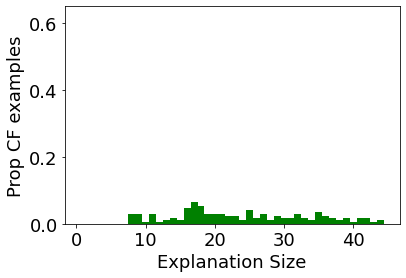}
    \includegraphics[scale=0.27]{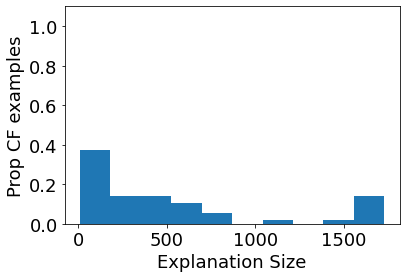}
    
        \caption{Histograms showing the proportion of counterfactual examples that have a certain explanation size from \basekeep{}. Note the $x$-axis for \synone{} goes up to 1500. Left: \synfour{}, Middle: \synfive{}, Right: \synone{}. }
        \label{fig:keep-explanation-size}
        \bigskip \bigskip

    \includegraphics[scale=0.27]{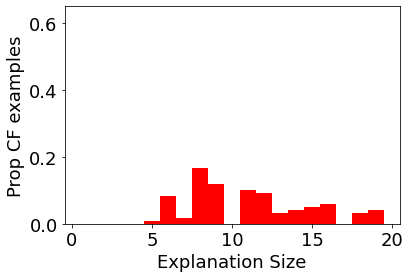}
    \includegraphics[scale=0.27]{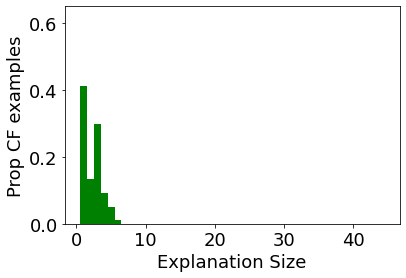}
    \includegraphics[scale=0.27]{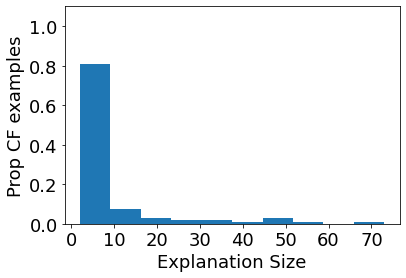}
    
        \caption{Histograms showing the proportion of counterfactual examples that have a certain explanation size from \baserm{}. Note the $x$-axis for \synone{} goes up to 70. Left: \synfour{}, Middle: \synfive{}, Right: \synone{}. }
        \label{fig:remove-explanation-size}
        \bigskip \bigskip
        
    % \includegraphics[scale=0.38]{images/tree-cycles-gnnexplainer.png}
    % \includegraphics[scale=0.38]{images/tree-grid-gnnexplainer.png}
    % \includegraphics[scale=0.38]{images/ba-shapes-gnnexplainer.png}
    
    %     \caption{Histograms showing explanation size from \gnnexplainer{} for $S=$ GT. Note that the y-axis goes up to 1. Left: \synfour{}, Middle: \synfive{}, Right: \synone{}.}
    %     \label{fig:gnnexplainer-explanation-size}

    \includegraphics[scale=0.27]{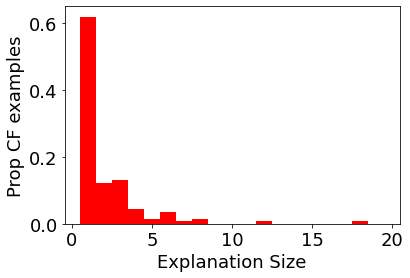}
    \includegraphics[scale=0.27]{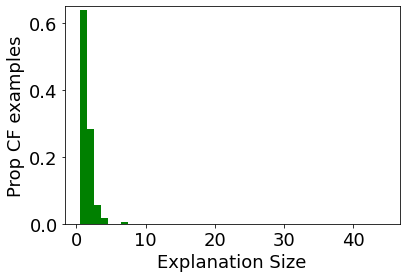}
    \includegraphics[scale=0.27]{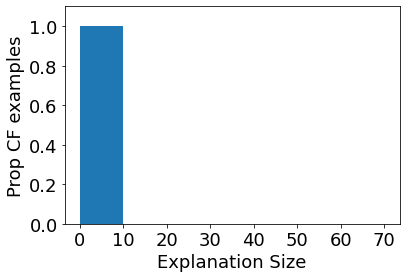}
    
        \caption{Histograms showing the proportion of counterfactual examples that have a certain explanation size from CF-GNNExplainer. Note the $x$-axis for \synone{} goes up to 70. Left: \synfour{}, Middle: \synfive{}, Right: \synone{}. }
        \label{fig:explanation-size}
        
\end{figure*}

\medskip \noindent
\textbf{Explanation Size:}
Figures~\ref{fig:random-explanation-size} to~\ref{fig:explanation-size} show histograms of the explanation size for CF-GNNExplainer and the baselines. 
We see that across all three datasets, CF-GNNExplainer has the smallest (i.e., most minimal) explanation sizes. 
This is especially true when comparing to \baserand{} and \basekeep{} for the \synone{} dataset, where we had to use a different scale for the $x$-axis due to how different the explanation sizes were. 
We postulate that this difference could be because \synone{} is a much more densely connected graph;
it has fewer nodes but more edges compared to the other two datasets, and the average number of nodes and edges in the \cgraph{} is order(s) of magnitude larger (see Table~\ref{table:stats}). 
Therefore, when performing random perturbations, there is substantial opportunity to remove edges that do not necessarily need to be removed, leading to much larger explanation sizes.
When there are many edges in the \cgraph{}, removing everything except the 1-hop neighbourhood, as is done in \basekeep{}, also results in large explanation sizes. 
In contrast, the loss function used by CF-GNNExplainer ensures that only a few edges are removed, which is the desirable behavior since we want minimal explanations. 

\pagebreak

\medskip \noindent
\textbf{Sparsity:}
CF-GNNExplainer outperforms the \baserand{}, \baserm{}, \basekeep{} baselines for all three datasets in terms of sparsity.
%\footnote{GNNExplainer cannot be evaluated on sparsity.} 
We note CF-GNNExplainer and \baserm{} perform much better on this metric in comparison to the other methods, which aligns with the results from explanation size. 

\medskip \noindent
\textbf{Accuracy:}
We observe that CF-GNNExplainer has the highest accuracy for the \synfour{} and \synfive{} datasets, whereas \baserm{} has the highest accuracy for \synone{}. 
However, we are unable to calculate the accuracy of \baserm{} for the other two datasets since it is unable to generate \emph{any} counterfactual examples for motif nodes, contributing to the low sparsity on those datasets. 
We observe accuracy levels upwards of 94\% for CF-GNNExplainer across \emph{all} datasets, indicating that it is consistent in correctly removing edges that are crucial for the initial predictions in the vast majority of cases (see Table~\ref{table:results1}).

\subsection{Comparison to \gnnexplainer{}}
Table~\ref{table:results-gnnexplainer} shows the results comparing our method to \gnnexplainer{}. We find that our method outperforms \gnnexplainer{} across all three datasets in terms of both fidelity and accuracy, for all tested values of $S$. 
However, this is not surprising since \gnnexplainer{} is not meant for generating counterfactual explanations, so we cannot expect it to perform well on a task it was not designed for. 
We cannot compare explanation size or sparsity fairly since \gnnexplainer{} requires the user to input the value of $S$.

\begin{table*}[]
\centering
\caption{Results comparing our method to \gnnexplainer{}. \gnnexplainer{} cannot find $S$ automatically, so we try varying values of $S$. GT indicates the size of the ground truth explanation for each dataset. CF-GNNExplainer finds $S$ automatically. Below each metric, $\blacktriangledown$ indicates a low value is desirable, while $\blacktriangle$ indicates a high value is desirable.}
\label{table:results-gnnexplainer}
\setlength{\tabcolsep}{4pt}
\scriptsize{
\begin{tabular}{lrrrr rrrr rrrr}
\toprule
\multicolumn{1}{c}{} & \multicolumn{4}{c}{\synfour{}}                                                                                                                 & \multicolumn{4}{c}{\synfive{}}                                                                                                                   & \multicolumn{4}{c}{\synone{}}                                                                                                                  \\ 
\cmidrule(r){2-5}\cmidrule(r){6-9}\cmidrule{10-13} 
               & \multicolumn{1}{c}{Fid.} & \multicolumn{1}{c}{Size} & \multicolumn{1}{c}{Spars.} & \multicolumn{1}{c}{Acc.} & \multicolumn{1}{c}{Fid.} & \multicolumn{1}{c}{Size} & \multicolumn{1}{c}{Spars.} & \multicolumn{1}{c}{Acc.} & \multicolumn{1}{c}{Fid.} & \multicolumn{1}{c}{Size} & \multicolumn{1}{c}{Spars.} & \multicolumn{1}{c}{Acc.} \\
\gnnexpshort{} & \multicolumn{1}{c}{$\blacktriangledown$} &\multicolumn{1}{c}{$\blacktriangledown$} &\multicolumn{1}{c}{$\blacktriangle$} & \multicolumn{1}{c}{$\blacktriangle$} & \multicolumn{1}{c}{$\blacktriangledown$} &\multicolumn{1}{c}{$\blacktriangledown$} &\multicolumn{1}{c}{$\blacktriangle$} & \multicolumn{1}{c}{$\blacktriangle$} & \multicolumn{1}{c}{$\blacktriangledown$} &\multicolumn{1}{c}{$\blacktriangledown$} &\multicolumn{1}{c}{$\blacktriangle$} & \multicolumn{1}{c}{$\blacktriangle$} \\
\midrule

 $S=1$ & 0.65 & 1.00 & 0.92 & 0.61 & 0.69 & 1.00 & 0.96 & 0.79 & 0.90 & 1.00 & 0.94 & 0.52 \\
 $S=2$ & 0.59 & 2.00 & 0.85 & 0.54 & 0.51 & 2.00 & 0.92 & 0.78 & 0.85 & 2.00 & 0.91 & 0.40  \\
 $S=3$ & 0.56 & 3.00 & 0.79 & 0.51 & 0.46 & 3.00 & 0.88 & 0.79 & 0.83 & 3.00 & 0.87 & 0.34 \\
 $S=4$ & 0.58 & 4.00 & 0.72 & 0.48 & 0.42 & 4.00 & 0.84 & 0.79 & 0.83 & 4.00 & 0.83 & 0.31 \\
 $S=5$ & 0.57 & 5.00 & 0.66 & 0.46 & 0.40 & 5.00 & 0.80  & 0.79 & 0.81 & 5.00 & 0.81 & 0.27 \\
 $S=$ GT &  0.55 &	6.00 & 0.57 &	0.46 &	0.35 &	11.83 &	0.53 &	0.74 &	0.82 &	6.00 &	0.79 &	0.24    \\

\midrule
CF-GNN               & \textbf{0.21}                              & 2.09                     & 0.90                       & \textbf{0.94}                      & \textbf{0.07}                              & 1.47                     & 0.94                       & \textbf{0.96}                      & \textbf{0.39}                              & 2.39                     & 0.99                       & \textbf{0.96}                 \\
\bottomrule
\end{tabular}
}
\end{table*}

\subsection{Summary of Results} 
Evaluating on four distinct metrics for each dataset gives us a more holistic view of the results. 
We find that across all three datasets, CF-GNNExplainer can generate counterfactual examples for the majority of nodes in the test set (i.e., low fidelity), while only removing a small number of edges (i.e., low explanation size, high sparsity). For nodes where we know the ground truth (i.e., those in the motifs) we achieve at least 94\% accuracy. 

Although \baserand{} can generate counterfactual examples for every node, they are not very minimal or accurate. 
The latter is also true for \basekeep{} -- in general, it has the worst scores for explanation size, sparsity and accuracy. 
\gnnexplainer{} performs at a similar level as \basekeep{}, indicating that although it is a prominent GNN XAI method, it is not well-suited for solving the counterfactual explanation problem. 

\baserm{} is competitive in terms of explanation size, but it performs poorly in terms of fidelity for the \synfour{} and \synfive{} datasets, and its accuracy on these datasets is unknown since it is unable to generate \emph{any} counterfactual examples for nodes in the motifs. 
These results show that our method is simple and effective in solving the counterfactual explanation task, unlike the baselines we test.

%!TEX root = ../main.tex

\section{Conclusion}
\label{section:cfgnn-conclusion}
In this chapter, we propose CF-GNNExplainer, a method for generating counterfactual explanations for any GNN. Our simple and effective method is able to generate counterfactual explanations that are (i) minimal, both in terms of the absolute number of edges removed (explanation size), as well as the proportion of the \cgraph{} that is perturbed (sparsity), and (ii) accurate, in terms of removing edges that we know to be crucial for the initial predictions. 

We evaluate our method on three commonly used datasets for GNN explanation tasks and find that these results hold across all three datasets. 
We find that existing GNN XAI methods are not well-suited to solving the counterfactual explanation task, while CF-GNNExplainer is able to reliably produce minimal, accurate counterfactual explanations. 

This answers \textbf{\ref{rq:cf-gnn}}: we can generate counterfactual explanations for graph-based models by extending the problem formalization from Chapter~\ref{chapter:research-focus} to accommodate graph data. 
We do so by introducing a perturbation matrix that acts on the adjacency matrix to remove edges in the graph, then applying similar gradient-based optimization techniques as in Chapter~\ref{chapter:research-focus} for each instance in the dataset. 
In the following chapter, we will investigate how to generate explanations that are specific to a particular real-world use case and evaluate them on real users.

\section*{Reproducibility}
To facilitate the reproducibility of the work in this chapter, our code is available at \url{https://github.com/a-lucic/cf-gnnexplainer}.

\part{Users}
% !TEX root = ../thesis-main.tex

%\chapter{Contrastive Explanations for Retail Forecasting}
\chapter{Contrastive Explanations for Forecasting Errors}
\label{chapter:research-mcbrp}

\footnote[]{This chapter was published at the ACM Conference on Fairness, Accountability, and Transparency (FAccT 2020) under the title ``Why Does My Model Fail: Contrastive Explanations for Retail Forecasting'' \citep{lucic_2020_why}, where it won a best paper award.}
\acresetall

%\todo{Add intro paragraph that connects this chapter to its RQ from chapter 1: (How) Can we develop an explanation method based on a real-world use case and evaluate it in a human-centric way?  (Yes), we took a use case from Ahold and developed an explanation method around those users' needs. We evaluate the method with a user study that has objective + subjective components.}

In this second part of the thesis, we shift our focus to the consumers of explanations: users. 
In this chapter, we address the following research question:

\medskip
\noindent
\textbf{\ref{rq:mcbrp}:} \acl{rq:mcbrp}
\medskip

\noindent
The answer to \textbf{\ref{rq:mcbrp}} is yes: we first identify a use case where users seek explanations: understanding errors in sales forecasting. 
We then design an algorithm that generates explanations for large errors in forecasting predictions. 
We evaluate our method through a user study with both researchers and practitioners to investigate the impact our explanations have on users'
\begin{inparaenum}[(i)]
	\item objective abilities to understand the model's predictions and 
	\item subjective attitudes towards the model. 
\end{inparaenum}
Our experimental results show that explanations generated by our method help both types of users understand why large errors in predictions occur, but do not have an impact on their trust or confidence in the model.

% !TEX root = ../thesis-main.tex

\section{Introduction}
\label{section:1}
As more and more decisions about humans are made by machines, it becomes imperative to understand how these outputs are produced and what drives a model to a particular prediction \citep{riberio-2016-model}. 
As a result, algorithmic interpretability has gained significant interest and traction in the machine learning (ML) community over the past few years \citep{doshi-2017-towards}.
However, there exists considerable skepticism outside of the ML community due to a perceived lack of transparency behind algorithmic predictions, especially when errors are produced \citep{dietvorst-2015-aa}. 
We aim to evaluate the effect of explaining model outputs, specifically large errors, on users' attitudes towards trusting and deploying complex, automatically learned models. 

Further motivation for explainable ML is provided by significant societal developments. 
Important examples include the recently enacted European General Data Protection Regulation (GDPR), which specifies that individuals will have the right to ``the logic involved in any automatic personal data processing'' \citep{gdpr}. 
In Canada and the United States, this right to an explanation is an integral part of financial regulations, which is why banks have not been able to use high-performing ``black-box'' models to evaluate the credit-worthiness of their customers. 
Instead, they have been confined to easily interpretable algorithms such as decision trees (for segmenting populations) and logistic regression (for building risk scorecards) \citep{khandani-2010-consumer}. 
At NeurIPS 2017, the Explainable ML Challenge was launched to combat this limitation, indicating the finance industry's interest in exploring algorithmic explanations \citep{fico_2017}. 

We use explanations as a mechanism for supporting innovation and technological development while keeping the human ``in the loop'' by focusing on predictive modeling as a tool that aids individuals with a given task. 
Specifically, our interest lies with interpretability in a scenario where users with varying degrees of ML expertise are confronted with large errors in the outcome of predictive models. 
We focus on explaining large errors because people tend to be more curious about unexpected outcomes rather than ones that confirm their prior beliefs~\citep{hilton-1986-knowledge}. 

However, \citet{dietvorst-2015-aa} show that when users are confronted with errors in algorithmic predictions, they are less likely to use the model. 
Seeing an algorithm make mistakes significantly decreases confidence in the model, and users are more likely to choose a human forecaster instead, even after seeing the algorithm outperform the human \citep{dietvorst-2015-aa}. 
This indicates that prediction mistakes have a significant impact on users' perception of the model. 
By focusing on explaining mistakes, we hope to give insight into this phenomenon of algorithm aversion while also giving users the types of explanations they are interested in seeing. 

Our work was motivated by the needs of analysts working on sales forecasting at Ahold Delhaize, a large retailer in the Netherlands. 
Current models in production are based on simple autoregressive methods, but there is an interest in exploring more complex techniques. 
However, the added complexity comes at the expense of interpretability, which is problematic for \OurCompany{}, especially when a complex model produces a forecast that is very different from the actual target value. 
This leads us to focus on explaining errors in regression predictions in this work. 
However, it should be noted that our method can be extended to classification predictions by defining ``distances'' between classes or by simply defining all errors as large errors. 

We focus on two aspects of explainability in this scenario: the \emph{generation} of explanations of large errors and the corresponding \emph{effectiveness} of these explanations. 
Prior methods for generating explanations fail at generating explanations for large errors because they produce similar explanations for predictions resulting in large errors and those resulting in reasonable predictions (see Table~\ref{table:lime} in Section~\ref{section:4} for an example). 
We propose a method for explaining large prediction errors, called Monte Carlo Bounds for Reasonable Predictions (MC-BRP), that shows users: 
\begin{enumerate}[(i)]
\item The required bounds of the most important features in order to have a prediction resulting in a reasonable prediction.
\item The relationship between each of these features and the target.
\end{enumerate}
\pagebreak

It should be noted that in this chapter, we focus on explaining errors \emph{in hindsight}, that is, we examine large errors once they have occurred and are not predicting them in advance without having access to the ground truth. We are also not using these explanations to improve the model, but rather examine the effectiveness of explaining large errors via \OurMethod{} on users' trust in the model and attitudes towards deploying it, as well as their understanding of the explanations. 
%We examine the effectiveness of explaining large errors via \OurMethod{} through a user study aimed at determining users' understanding of the explanations as well as their trust in the model and attitudes towards deploying it based on the explanations. 
We test on a wide range of users, including both practitioners and researchers, and analyze the differences in attitudes between these users.  
We also reflect on the process of conducting a user study by outlining some limitations of our study and make some recommendations for future work. 
We address the following research questions: \\
\textbf{RQ3.1:} \emph{Are the contrastive explanations generated by \OurMethod{} about large errors in predictions
\begin{inparaenum}[(i)]
\item interpretable, or
\item actionable? 
\end{inparaenum}}
More specifically, 
\begin{enumerate}[(i)]
 \item Can contrastive explanations about large errors give users enough information to simulate the model's output (forward simulation)?
 \item Can such explanations help users understand the model such that they can manipulate an observation's input values in order to change the output (counterfactual simulation)?
\end{enumerate}
\textbf{RQ3.2:} \emph{How does providing contrastive explanations generated by MC-BRP for large errors impact users' perception of the model?} Specifically, we want to investigate the following:
\begin{enumerate}[(i)]
 \item Does being provided with contrastive explanations generated by MC-BRP impact users' understanding of why the model produces errors?
 \item Does it impact their willingness to deploy the model?
 \item Does it impact their level of trust in the model?
 \item Does it impact their confidence in the model's performance?
\end{enumerate}
Consequently, we make the following contributions:
\begin{itemize}
	\item We contribute a method, \OurMethod{}, for generating contrastive explanations specifically for large errors in regression tasks. 
	\item We evaluate our explanations through a user study with \numprint{75} participants in both objective and subjective terms. 
	\item We conduct an analysis on the differences in attitudes between practitioners and researchers.  
\end{itemize}

\noindent
In Section~\ref{section:2} we discuss related work and identify how our problem relates to the current literature. 
In Section~\ref{section:3} we formally describe the methodology of explanations based on \OurMethod{} and in Section~\ref{section:4} we describe our experimental setup. 
In Section~\ref{section:5} we detail the results of the user study; we conduct further analyses in Section~\ref{section:6}. 
In Section~\ref{section:7} we conclude and make recommendations for future work.

% !TEX root = ../thesis-main.tex

\section{Related Work}
\label{section:2}
Based on the taxonomy described in Chapter~\ref{chapter:introduction}, our setting in this chapter is an \emph{local explanation} problem. Our method is model-agnostic in nature, but we evaluate it specifically on \emph{tree ensembles}. 
We use \emph{sensitivity analysis}, specifically Monte Carlo simulations, on \emph{tabular} data to generate our explanations. 

\subsection{Local Explanations for Tree Ensembles}

Existing work on generating local explanations for tree ensembles involves finding counterfactual examples \citep{tolomei_interpretable_2017}, identifying influential training samples \citep{sharchilev-2018-finding}, or identifying important features \citep{lundberg_explainable_2019}. 
Importantly, none of these publications are specifically about 
\begin{inparaenum}[(i)]
	\item explaining errors, or
	\item explaining regressions. 
\end{inparaenum} 
On the contrary, these publications are all based on binary classification tasks and the explanations do not necessarily provide insight into prediction mistakes. 

\citet{tolomei_interpretable_2017} propose a method for generating counterfactual examples by identifying decision paths of interest that would result in a different prediction, then traversing down each of these paths and perturbing the instance $x$ such that it satisfies the path in question. 
If this perturbation, $x'$,  
\begin{inparaenum}[(i)]
	\item satisfies the decision path, and
	\item changes the prediction in the overall ensemble, 
\end{inparaenum}
then it is a candidate transformation of $x$. 
After computing all possible candidate transformations by traversing over all paths of interest (i.e., those leading to a different prediction), the candidate transformation with the smallest distance from $x$ is selected as the counterfactual example. 
The explanation, then, is the difference between $x$ and $x'$. 
Although \citet{tolomei_interpretable_2017}'s method also produces contrastive explanations, our method differs from theirs since we are not aiming to identify one counterfactual example, but rather a range of feature values for which the prediction would be different. 
Another difference is that we do not assume full access to the original model. 

\citet{sharchilev-2018-finding} also generate local explanations for tree ensembles. 
Their methodology is based on finding influential training samples in order to automatically improve the model, which differs from our work since their explanations are not of a contrastive nature. 
These influential training samples help us understand why a certain class was predicted for a given instance, but they make no reference to the alternative class(es). 
It should be noted that they include a use case on identifying harmful training examples --- ones that contributed to incorrect predictions --- which can be seen as a way to explain errors. 
%\citet{koh-2017-understanding} also use influence functions to show the effect of upweighting samples or perturbing feature values on a model's parameters, but their method only applies to smooth parametric models. 

\subsection{Feature Importance Explanations}

\citet{lundberg_explainable_2019} propose a method for determining how much each feature contributes to a prediction and present a ranked list of the most important features as the explanation. 
The approach is based on the computationally intensive Shapley values \citep{lundberg_unified_2017}, for which the authors develop a tree-specific approximation. 
This differs from our method since identifying the most important features is only a preliminary step in our pipeline --- our work extends beyond this by including
\begin{inparaenum}[(i)]
\item feature bounds that result in reasonable predictions, and 
\item the relationship between the features and the target as a tool to help users inspect what goes wrong when the prediction error is large.
\end{inparaenum}

\citet{ribeiro-2016-should} also propose a method for identifying local feature importances and this is the one we use in our pipeline. 
Their method, LIME, is model-agnostic and is based on approximating the original model locally with a linear model. 
We share their objective of evaluating users' attitudes towards a model through local explanations but we further specify our task as explaining instances where there are large errors in predictions. 
Based on preliminary experiments, we find that LIME is insufficient for our task setting for two reasons: 
\begin{enumerate}[(i)]
\item For regression tasks, LIME's approximation of the original model is not exact. This ``added'' error can be quite large given that our target is typically of order $10^6$, and this convolutes our definition of a large error.
\item The features LIME deems most important are similar regardless of whether the prediction results in a large error or not, which does not provide any specific insight into why a large error occurs. These experiments are detailed in Section~\ref{section:4}.
\end{enumerate}

\noindent%
\subsection{Contrastive Explanations}
Other work on contrastive explanations includes identifying features that should be present or absent in order to justify a classification  \citep{dhurandhar_explanations_2018, ferrari_grounding_2018} or model-agnostic counterfactuals \citep{wachter_counterfactual_2017, russell_efficient_2019}. 
These all differ from our method since they are not specifically about explaining errors. 
Furthermore, the work by \citet{dhurandhar_explanations_2018} and \citet{ferrari_grounding_2018} is based on the binary presence or absence of input features, whereas our method perturbs inputs instead of removing them altogether. 

\subsection{Outlier Detection}

Our work in this chapter can also be viewed as a form of outlier detection. 
However, it differs from the standard literature outlined by \citet{pimentel-2014-review} with respect to the objective: we are not necessarily trying to identify outliers in terms of the training data but rather explain instances in the test set whose errors are so large  that they are considered to be anomalies.

%\citet{miller_ijcai_2017} perform a survey of the papers cited in the ``Related Works'' section of the call for the IJCAI 2017 Explainable AI workshop \citep{ijcai-2017-workshop} and find that the majority do not base their methods on the available research about explanations from other disciplines such as philosophy, psychology or cognitive sciences, or evaluate on real users. 
%In contrast, our method is rooted in the corresponding philosophical literature \citep{hilton-1990-conversational, lipton-1990-contrastive, hilton-1986-knowledge} and our evaluation is based on a user study. 

% !TEX root = ../thesis-main.tex

\section{Method: MC-BRP}
\label{section:3}

The intuition behind \OurMethod{} is based on identifying the unusual properties of a particular observation. 
We make the assumption that large errors occur due to unusual feature values in the test set that were not common in the training set. 

Given an observation that results in a large error, \OurMethod{} generates a set of bounds for each feature that would result in a reasonable prediction as opposed to a large error. 
We also include the trend as part of the explanation in order to help users understand the relationship between each feature and the target, and how the input should be changed in order to change the output. 

As pointed out previously, we consider our task of identifying and explaining large errors somewhat similar to that of an outlier detection problem. 
A standard definition of a statistical outlier is an instance that falls outside of a threshold based on the interquartile range. 
A widely used version of this, called Tukey's fences, is defined as follows \citep{tukey-1977-exploratory}:
\[ 
[Q_1 - 1.5(Q_3 - Q_1), Q_3 + 1.5(Q_3 - Q_1)],
\]
where $Q_1$ and $Q_3$ are the first and third quartiles, respectively. 

\begin{definition}\rm
\label{def:large-error}
Let $x$ be an observation in the test set $X$ and let $t$, $\hat{t}$ be the actual and predicted target values of $x$, respectively. Let $\epsilon$ be the corresponding prediction error for $x$, and let $E$ be the set of all errors of $X$. Then $\epsilon$ is a \emph{large error} iff 
\[
\epsilon > Q_3(E) + 1.5(Q_3(E) - Q_1(E)),
\]
where $Q_1(E), Q_3(E)$ are the first and third quartiles of the set of errors, respectively. 
We denote this threshold as $\epsilon_{large}$.
\end{definition}

\noindent%
\if0
We can view $X$ in Definition~\ref{def:large-error} as a disjoint union of two sets, $R$ and $L$, where $R$ is the set of observations that resulted in reasonable predictions, and $L$ is the set of observations that resulted in large errors. 
\fi
We can view $X$ in Definition~\ref{def:large-error} as a disjoint union of two sets:
\begin{enumerate}[(i)]
	\item $R$: the set of observations resulting in reasonable predictions, and
	\item $L$: the set of observations resulting in large errors.
\end{enumerate}
We determine the $n$ most important features based on LIME $\Phi^{(x)} = \{\phi_j^{(x)}\}_{j=1}^{n}$, for all $x \in X$. 
It should be noted there exist alternative methods for determining the most important features for a particular prediction \citep{lundberg_unified_2017}, which would also be appropriate.

Given $x \in X$, for each $\phi_j^{(x)} \in \Phi^{(x)}$, we determine two sets of characteristics through Monte Carlo simulations:
\begin{enumerate}[(i)]
 \item $[a_{\phi_j^{(x)}}, b_{\phi_j^{(x)}}]$: the bounds for values of $\phi_j^{(x)}$ such that $x \in R$, $x \not\in L$.
 \item $\rho_{\phi_j^{(x)}}$: the relationship between $\phi_j^{(x)}$ and the target, $t$.
\end{enumerate}
We perturb the feature values for $l \in L$ using Monte Carlo simulations in order to determine what feature values are required to produce a reasonable prediction. 
The algorithm for determining $R'$, the set of Monte Carlo simulations resulting in reasonable predictions, is detailed in Algorithm~\ref{mcbrp}.

Given $l \in L$, we determine Tukey's fences for each feature in $\Phi^{(l)}$ based on the feature values from $R$. 
This gives us the bounds from which we sample for our feature perturbations. 

Next, we randomly sample from these bounds for each $\phi_j^{(l)} \in \Phi^{(l)}$ $m$-times to generate $mn$ versions of our original observation, $l$. 
We call the $i$-th perturbed version $l_i'$, where $i \in \left\{1, \ldots, mn\right\}$. 

We then test the original model $f$ on each $l_i'$, obtain a new prediction, $\hat{t_i'}$, and construct $R'$, the set of perturbations resulting in reasonable predictions. 

Once $R'$ is generated, we compute the mean, standard deviation and Pearson coefficient \citep{swinscow-1997-stats} of the top $n$ features of $l \in L$, $\Phi^{(l)}$, based on this set. 
\if0
\bigskip
\begin{algorithm}[h]
    \caption{Monte Carlo simulation: creates a set of perturbed instances resulting in reasonable predictions $R'$ for each large error $l \in L$}
    \label{mcbrp}
    \begin{algorithmic}[1] 
        \Require{instance $l$}
        \Require{set of  $l$'s most important features $\Phi^{(l)}$}
        \Require{`black-box' model $f$}
        \Require{large error threshold $\epsilon_{large}$}
        \Require{number of MC perturbations per feature $m$} 
        \State $R' = \emptyset$ 
            \ForAll{$\phi_j^{(l)}$ in $\Phi^{(l)}$}
                \State $TF(\phi_j^{(l)}$) $\gets$ Tukey's fences for $\phi_j^{(l)}$ \Comment{Based on $R$}
                \For{$i$ in range (0, $m$)}
                    \State $\phi_j'^{(l)} \gets randomsample(TF(\phi_j^{(l)}$))
                    \State $l_i' \gets l_i.replace(\phi_j^{(l)}, \phi_j'^{(l)})$              
                    \State $\hat{t_i'} \gets f(l_i')$ \Comment{New prediction} 
                    \If{$|\hat{t'_i} - t_i| < \epsilon_{large}$}
                        \State {$R' \gets R' \cup l_i'$}
                   %  \EndIf                    
                \EndFor
            \EndFor
            \Return{$R'$}
    \end{algorithmic}
\end{algorithm}
\fi

% MC-BRP
\begin{algorithm}
    \caption{Monte Carlo simulation: creates a set of perturbed instances resulting in reasonable predictions $R'$ for each large error $l \in L$}
    \label{mcbrp}
    \begin{algorithmic}
     \STATE {\bfseries Input:} instance $l$, set of  $l$'s most important features $\Phi^{(l)}$, `black-box' model $f$, large error threshold $\epsilon_{large}$, number of MC perturbations per feature $m$. 
        \STATE $R' = \emptyset$ 	
        
            \FORALL{$\phi_j^{(l)}$ in $\Phi^{(l)}$}
            
                \STATE $TF(\phi_j^{(l)}$) $\gets$ \textsc{Tukeys\_fences}($\phi_j^{(l)}$) \qquad \textcolor{gray}{\textit{\# Based on $R$}} 
                
                \FOR{$i$ in range (0, $m$)}
                
                    \STATE $\phi_j'^{(l)} \gets \textsc{random\_sample}(TF(\phi_j^{(l)}$))
                    
                    \STATE $l_i' \gets l_i.\textsc{replace}(\phi_j^{(l)}, \phi_j'^{(l)})$            
                      
                    \STATE $\hat{t_i'} \gets f(l_i')$ \qquad \textcolor{gray}{\textit{\#New prediction} }
                    
                    \IF{$|\hat{t'_i} - t_i| < \epsilon_{large}$}
                    
                        \STATE {$R' \gets R' \cup l_i'$}
                   \ENDIF                    
                \ENDFOR
            \ENDFOR
            \RETURN{$R'$}
    \end{algorithmic}
\end{algorithm}

\begin{definition}\rm
The \emph{trend}, $\rho_{\phi_j^{(x)}}$, of each feature is the Pearson coefficient between each feature $\phi_j^{(x)}$ and the predictions $\hat{t_i'}$ based on the observations in $R'$. It is a measure of linear correlation between two variables \citep{swinscow-1997-stats}. 
\end{definition}

%noindent%
The set of bounds for each feature in $\Phi^{(x)}$ such that $\hat{t}$ results in a reasonable prediction are based on the mean and standard deviation of each $\phi_j^{(x)} \in \Phi^{(x)}$.

\begin{definition}\rm
The \emph{reasonable bounds} for values of each feature $\phi_j$ in $\Phi^{(x)}$, $[a_{\phi_j^{(x)}}, b_{\phi_j^{(x)}}]$, are 
\[
\left[\mu(\phi_j^{(x)})  - \sigma(\phi_j^{(x)}),    \mu(\phi_j^{(x)})  + \sigma(\phi_j^{(x)})\right],
\]
where $\mu(\phi_j^{(x)})$ and $\sigma(\phi_j^{(x)})$ are the mean and standard deviation of each feature, respectively, based on $R'$. 
\end{definition}

%\noindent%
We compute the trend and the reasonable bounds for each of the $n$ most important features and present them to the user in a table. 
Table~\ref{table:example} shows an example of an explanation generated by \OurMethod{}; the dataset used for this example is detailed in Section~\ref{section:dataset}.

\begin{table}
\caption{An example of an explanation generated by MC-BRP. Here, each of the input values is outside of the range required for a reasonable prediction, which explains why this particular prediction results in a large error. }
\label{table:example}
%\begin{tabular}{cp{5.0cm}p{5.8cm}rc}
\begin{tabular}{cccrc}
\toprule
\bf Input & \bf Definition & \bf Trend & \bf Value & \bf Reasonable\\
\bf  & \bf  & \bf  & \bf  & \bf  range\\
\midrule
A & total\_contract\_hrs & As input $\uparrow$, sales $\uparrow$
 & 9628.0 & [4140,6565] \\
B & advertising\_costs &  As input $\uparrow$, sales $\uparrow$
 & 18160.7 & \phantom{0}[8290,15322] \\
C & num\_transactions & As input $\uparrow$, sales $\uparrow$
 & 97332.0 & [51219,75600] \\
D & total\_headcount & As input $\uparrow$, sales $\uparrow$
 & 226.0 & \phantom{0}[95,153] \\
E & floor\_surface & As input $\uparrow$, sales $\uparrow$
 & 2013.6 & \phantom{0}[972,1725] \\  
\bottomrule
\end{tabular}
\end{table}

% !TEX root = ../thesis-main.tex

\section{Experimental Setup} 
\label{section:4}
Current explanation methods mostly serve individuals with ML expertise~\citep{guidotti-2018-survey,bhatt_explainable_2019}, but they should be extended to cater to users outside of the ML community~\citep{miller-2017-explanations}. 
Unlike previous work, our method, \OurMethod{}, generates contrastive explanations by framing the explanation around the prediction error, and aims to help users understand
\begin{inparaenum}[(i)]
\item what contributed to the large error, and 
\item what would need to change in order to produce a reasonable prediction. 
\end{inparaenum}
Presenting explanations in a contrastive manner helps frame the problem and narrows the user's focus regarding the possible outcomes \citep{hilton-1990-conversational,lipton-1990-contrastive}. 

Our explanations are contrastive because they display to the user what would have needed to change in the input order to obtain an alternative outcome from the model --- in other words, why this prediction results in a large error as opposed to a reasonable prediction. 

\subsection{Dataset and Model}
\label{section:dataset}
Our task is predicting monthly store sales using an internal company dataset with 45 features including financial, workforce and physical store aspects. 
Since not all of our practitioners have experience with ML, using an internal dataset with familiar features allows them to leverage some of their domain expertise. 
The dataset includes 45628 observations from 563 stores, collected at four-week intervals spanning from 2010--2015. 
We split the data by year (training: 2010--2013, test: 2014--2015) to simulate a production environment, and we treat every unique combination of store, interval and year as an independent observation. 
After preprocessing, we have 21415 and 12239 observations in our training and test sets, respectively. 
We train the gradient boosting regressor from scikit-learn with the default settings and obtain an $R^{2}$ of 0.96.

We verify our assumption that large errors are a result of unusual feature values by generating \OurMethod{} explanations for all instances in our test set using $n$ = 5 features and $m$ = 10000 Monte Carlo simulations. In our dataset, we find that 48\% of instances resulting in large errors have feature values outside the reasonable range for all of the $n$ = 5 most important features, compared to only 24\% of instances resulting in reasonable predictions. 
Although this is not perfect, it is clear that \OurMethod{} produces explanations that are at least somewhat able to distinguish between these two types of predictions.

%\subsection{Why Existing Solutions are Insufficient}
\subsection{Comparison to LIME}
\label{section:lime}
\citet{hilton-2017-social} states that explanations are selective -- it is not necessary or even useful to state all the possible causes that contributed to an outcome. 
The significant part of an explanation is what distinguishes it from the alternative outcome.
If LIME explanations were suitable for our problem, then we would expect to see different features deemed important for instances resulting in large errors compared to those resulting in acceptable errors. 
This would help the user understand why a particular prediction resulted in a large error. 

However, when generating LIME explanations for our test set using $n$ = 5 features, we do not see much of a distinction in the most important features between predictions that result in large errors and those that do not. 
For example, advertising\_costs is one of the top 5 most important features in 18.8\% of instances with large errors and 18.7\% of instances with reasonable predictions. 
These results are summarized in Table~\ref{table:lime}. 
\medskip

\begin{table}[]
\caption{The top $n=5$ features according to LIME for observations resulting in large errors vs. reasonable predictions.}
\label{table:lime}
\centering
\begin{tabular}{llll}
\toprule
\multicolumn{2}{l}{\textbf{Large errors}} & \multicolumn{2}{l}{\textbf{Reasonable Predictions}} \\
\midrule
advertising\_costs             & 0.188      & advertising\_costs                 & 0.187        \\
total\_contract\_hrs                 & 0.175      & total\_contract\_hrs                    & 0.179        \\
num\_transactions            & 0.151      & num\_transactions               & 0.156        \\
floor\_surface                & 0.124      & total\_headcount                & 0.134        \\
total\_headcount             & 0.123      & floor\_surface                      & 0.122        \\
month                     & 0.109      & month                          & 0.094        \\
mean\_tenure      & 0.046      & mean\_tenure         & 0.046        \\
earnings\_index                & 0.033      & earnings\_index                   & 0.031        \\
\bottomrule
\end{tabular}
\end{table}

%\noindent%
Furthermore, we originally tried to design our control group user study using explanations from LIME, but found that test users from \OurCompany{} could not make sense of the objective questions about prediction errors because LIME does not provide any insight about errors specifically. 
Given that we could not even ask questions about errors using LIME explanations to users without confusing them, it is clear that LIME is inappropriate for our task. 

\subsection{User Study Design}
\label{section:studydesign}
We test our method on a real dataset with real users, both from the retailer. 
We include a short tutorial about predictive modeling along with some questions to check users' understanding as a preliminary component of the study. 
This is because our users are a diverse set of individuals with a wide range of capabilities, including data scientists, human resource strategists, and senior members of the executive team. 
We also include \if0 some\fi participants from \OurUniversity{} to simulate users who could one day work in this environment. 
In total, we have 75 participants: 44 in the treatment group and 31 in the control group. 

\begin{figure}[t]
\centering
\includegraphics[clip,trim=0mm 0mm 0mm 7.3mm,scale=0.38]{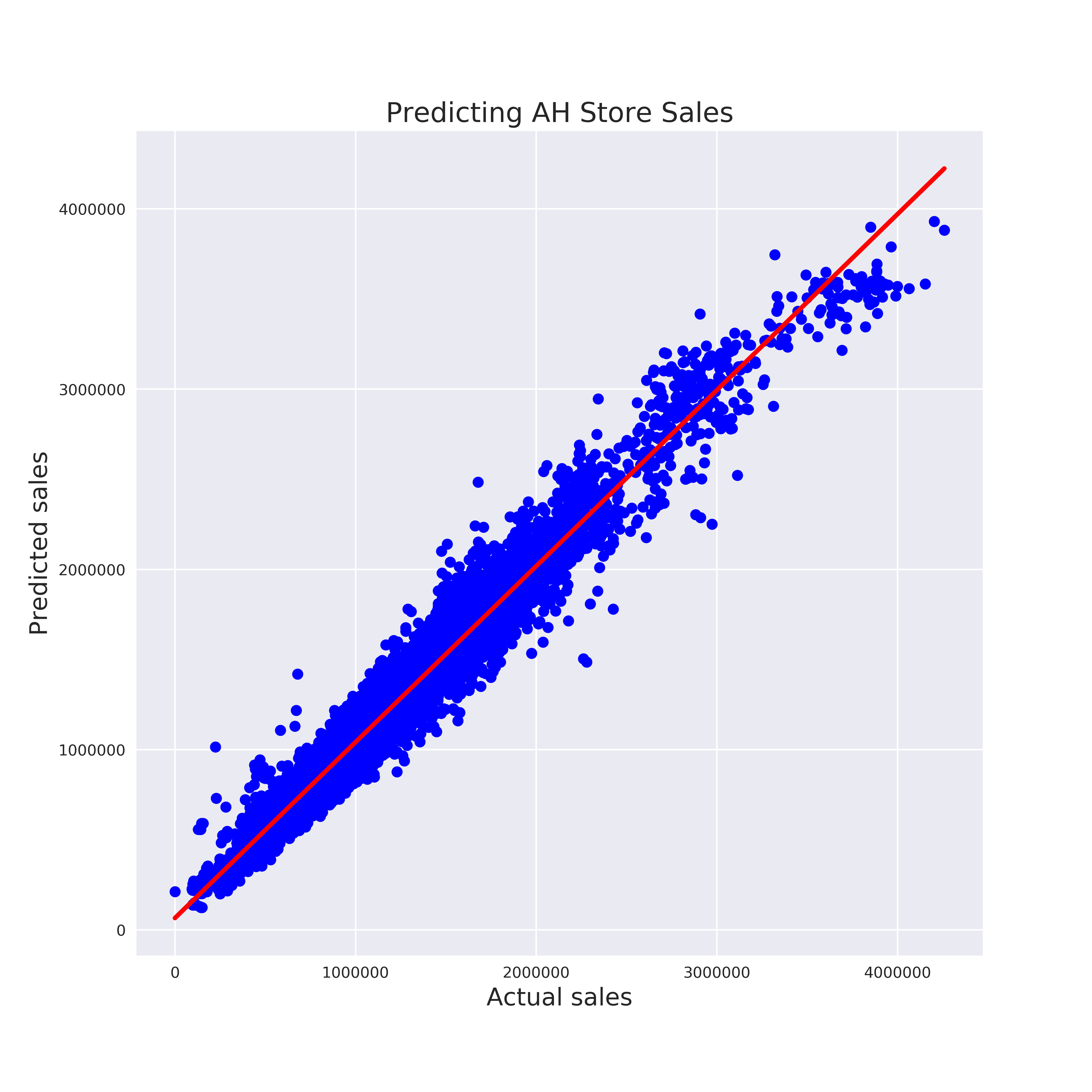}
\caption{The visual description of the model shown to the users: a graph comparing the predicted sales and actual sales based on the original model. The red line depicts perfect predictions.}
\label{fig:pred}
\vspace*{-.5\baselineskip}
\end{figure}

All users are first provided with a visual description of the model: a simple scatter plot comparing the predicted and actual sales (as shown in Figure~\ref{fig:pred}). 
We also show a pie chart depicting the proportion of predictions that result in large errors to give users a sense of how frequently these mistakes occur. 
In our case, this is 4\%. 
Since our users are diverse, we want to make our description of the model as accessible as possible while allowing them to form their own opinions about how well the model performs. 
Participants in the treatment group are shown \OurMethod{} explanations, while those in the control group are not given any explanation.

%\noindent%
The study contains two components, objective and subjective, corresponding to \textbf{RQ3.1} and \textbf{RQ3.2}, respectively. 
The objective component is meant to quantitatively evaluate whether or not users understand explanations generated by \OurMethod{}, while the subjective component assesses the effect of seeing the explanation on users' attitudes towards the model. 
We base the objective component on \textit{human-grounded metrics}, a framework proposed by \citet{doshi-2017-towards}, where the tasks conducted by users are simplified versions of the original task. 
Instead of asking users to correctly predict retail sales values of order $10^6$, we ask them to determine whether or not a prediction will result in a large error. 
%We modify the original sales prediction task into a binary classification one: we ask users to determine whether or not a prediction will result in a large error, as it seems unreasonable to expect humans to correctly predict retail sales values of order $10^6$. 

\begin{table}[h]
\caption{Summary of tasks performed in user study for the treatment and control groups. The subjective questions are asked twice. }
\label{table:study}
\centering
\begin{tabular}{ll}
\toprule
\multicolumn{1}{c}{\textbf{Treatment}} & \multicolumn{1}{c}{\textbf{Control}} \\ 
\midrule
1. Short modeling tutorial        & 1. Short modeling tutorial         \\
2. Visual model description            & 2. Visual model description             \\
3. Subjective questions                & 3. Subjective questions                 \\
4. Objective questions                 & 4. Dummy questions                      \\
5. Subjective questions       & 5. Subjective questions   \\
\bottomrule      
\end{tabular}
\end{table}

\begin{table}[h]
\caption{Summary of simulations performed in objective portion of the user study.}
\label{table:simulations}
\centering
\begin{tabular}{lll}
\toprule
\textbf{Type} & \textbf{Provide user with} & \textbf{User's task} \\ 
\midrule
Forward & (1) Input values & Simulate output  \\
 & (2) Explanation & \\ 
\midrule
Counterfactual & (1) Input values & \multirow[t]{2}{3.4cm}{Manipulate input to change output} \\
 & (2) Explanation  &  \\
 & (3) Output  &  \\ 
\bottomrule
\end{tabular}
\end{table}

To answer \textbf{RQ3.1}, we ask users in the treatment group to perform two types of simulations, both suggested by \citet{doshi-2017-towards} and summarized in Table~\ref{table:simulations}. 
The first is \textit{forward simulation}, where we provide participants with the 
\begin{inparaenum}[(i)]
\item input values, and 
\item explanation. 
\end{inparaenum}
We then ask them to simulate the output --- whether or not this prediction will result in a large error. 
The second is \textit{counterfactual simulation}, where we provide participants with the
\begin{inparaenum}[(i)]
\item input values, 
\item explanation, and 
\item output. 
\end{inparaenum}
We then ask them what they would have needed to change in the input in order to change the output. 
In other words, we want participants to determine how the input features can be changed (according to the trend) in order to produce a reasonable prediction as opposed to one that results in large error. 
These objective questions are designed to test whether or not a participant understands the explanations enough to predict or manipulate the model's output. 
We ask every participant in the treatment group to perform two forward simulations and one counterfactual simulation, and we show the same examples to all users. 

For the control group, we found that we could not ask the objective questions in the same way we did for the treatment group. 
This is because the objective component involves simulating the model based on the explanations (see Table~\ref{table:simulations}), which is not possible if the explanations are not provided. 
In fact, we initially left the objective questions in the control group study, but preliminary testing on some users from \OurCompany{} showed that this was confusing and unclear, similar to when we tried using LIME explanations. 
We were concerned this confusion would skew users' perceptions of the model and therefore convolute the results of \textbf{RQ3.2}. 
Instead, we show participants in the control group the
\noindent
\begin{inparaenum}[(i)]
\item input values, and 
\item output -- whether or not the example resulted in a large error. 
\end{inparaenum}
In this case, we ask them \emph{if they have enough information} to determine why the example does (or does not) result in a large error. 
This serves as a dummy question to engage users with the task without confusing them. 
We cannot ask users in the control group to simulate the model since they do not see the explanations, but we want to mimic the conditions of the treatment group as closely as possible. 
Therefore, \textbf{RQ3.1}, is solely evaluated on users from the treatment group. 

To answer \textbf{RQ3.2}, we contrast results from the treatment and control groups. 
We ask both groups of users the same four subjective questions twice, once towards the beginning of the study and once again at the end. 
We ask the questions at the beginning of the study to evaluate the distribution of preliminary attitudes towards the model, based solely on the visual description. 
We ask the questions at the end of the study to evaluate the effectiveness of \OurMethod{} explanations, by comparing the results from the treatment and control groups. 
The questions we devised are based on the user study by \citet{terhoeve-2017-news}. 
Table~\ref{table:study} summarizes the experimental setup for the treatment and control groups. 
Again, the treatment and control groups are treated exactly the same with the exception of the objective questions -- we only ask these to the treatment group since we cannot ask users to simulate the model without giving them the explanation. 
% !TEX root = ../thesis-main.tex

\section{Experimental Results}
\label{section:5}
%We aim to give insight into the problem of algorithmic aversion \citep{dietvorst-2015-aa} by explaining mistakes in a model's predictions. 
In this section, we evaluate the explanations generated by \OurMethod{} in terms of 
\begin{inparaenum}[(i)]
	\item objective questions, and 
	\item subjective questions. 
\end{inparaenum}

\begin{table}[b]
\caption{Results from the objective questions in the user study.}
\label{table:objective}
\centering
\begin{tabular}{ll}
\toprule
\multicolumn{2}{c}{\textbf{Human accuracy}} \\ 
\midrule
Forward simulation 1         & 84.1\%     \\
Forward simulation 2         & 84.1\%     \\
Counterfactual simulation      & 75.0\%     \\ 
\midrule
Average      & 81.1\% \\
\bottomrule
\end{tabular}
\end{table}

\subsection{Objective Questions}
%First, we evaluate users' comprehension of \OurMethod{} explanations through objective questions based on those in the treatment group. 
The results for users' objective comprehension of \OurMethod{} explanations are summarized in Table~\ref{table:objective}. 
We see that explanations generated by MC-BRP are both: 
\begin{inparaenum}[(i)]
\item interpretable and 
\item actionable, 
\end{inparaenum}
with an average accuracy of 81.1\%. 
This answers \textbf{RQ3.1}. 
When asked to perform forward simulations, the proportion of correct answers was 84.1\% for both questions. 
This indicates that the majority of users were able to interpret the explanations in order to simulate the model's output (\textbf{RQ3.1:} interpretable). 
When asked to perform counterfactual simulations, the proportion of correct answers was slightly lower at 75.0\%, but still indicates that the majority of users were able to determine how to manipulate the model's input in order to change the output (\textbf{RQ3.1:} actionable).

\begin{figure}[]
 \centering
 \includegraphics[clip,trim=0mm 0mm 0mm 0mm,scale=0.35]{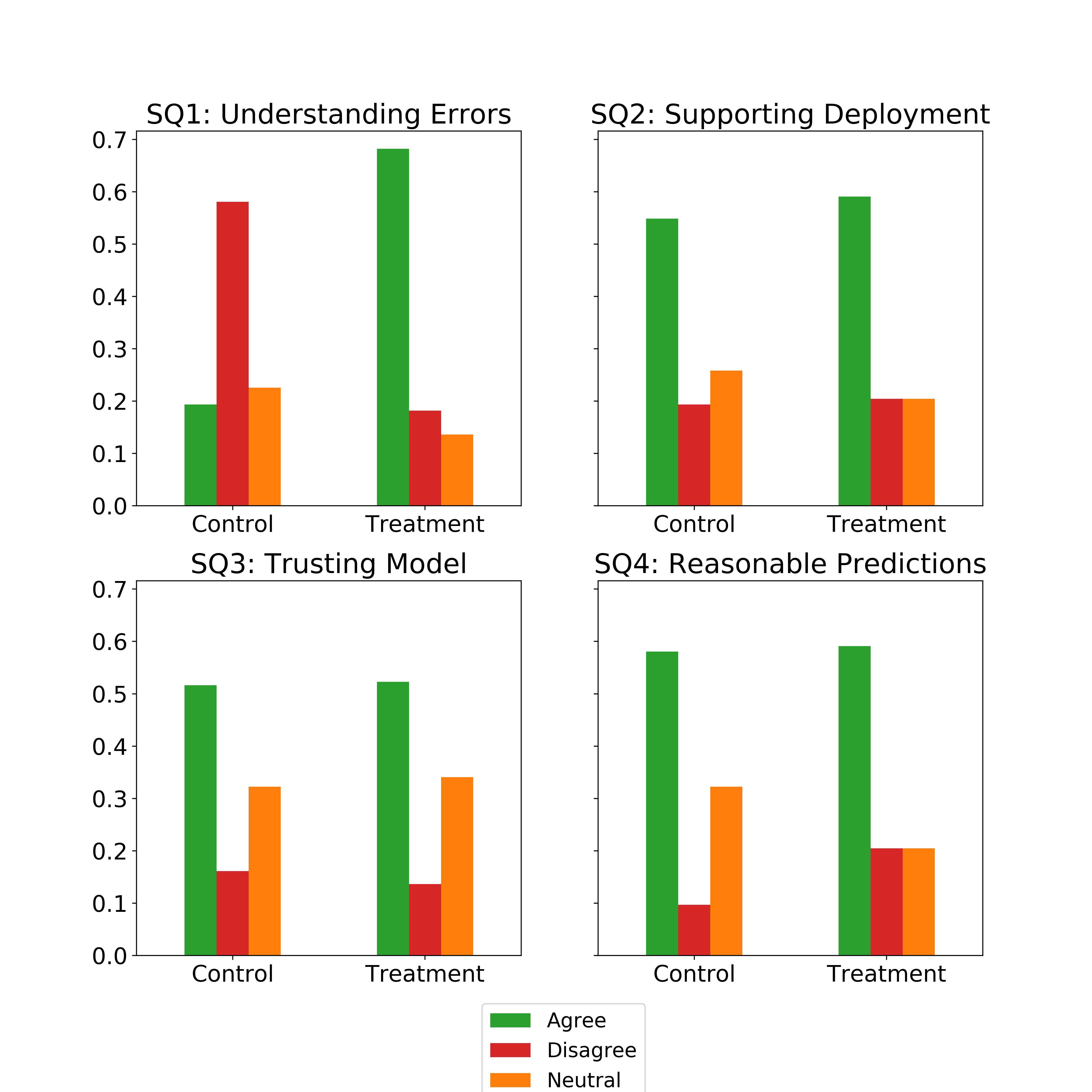}
  \caption{Results from a within-subject study comparing answers between the treatment (\OurMethod{} explanation) and control (no explanation) groups.}
 \label{fig:treat-vs-control}
\end{figure}

\subsection{Subjective Questions}
In order to understand the impact of \OurMethod{} explanations on users' attitudes towards the model, we ask them the following subjective questions:
\begin{itemize}
	\item \textbf{SQ1:} I understand why the model makes large errors in predictions.
	\item \textbf{SQ2:} I would support using this model as a forecasting tool.
	\item \textbf{SQ3:} I trust this model.
	\item \textbf{SQ4:} In my opinion, this model produces mostly reasonable outputs.
\end{itemize}

To ensure our populations did not have different initial attitudes towards the model, we compared their answers on the subjective questions after only showing a visual description of the model. 
The visual description is a graph comparing the predicted sales to the actual sales, which allows users to see the distribution of errors made by the model (see Figure~\ref{fig:pred}). 
We found no statistically significant difference ($\chi^{2}$ test, $\alpha = 0.05$) in initial attitudes towards the model, which allows us to postulate that any difference discovered between the two groups is a result of the treatment they were given (i.e., \OurMethod{} explanation vs. no explanation). 

Figure~\ref{fig:treat-vs-control} shows the distributions of answers to the four subjective questions in the treatment and control groups. 
The difference in distributions is significant for SQ1 ($\chi^{2} = 18.2$, $\alpha = 0.0001$): users in the treatment group agree with the statement more than users in the control group. 
However, we find no statistically significant difference between the two groups for the remaining questions ($\chi^{2}$ test, $\alpha = 0.05$). 
That is, \OurMethod{} explanations help users understand why the model makes large errors in predictions, but do not have an impact on users' trust or confidence in the model, or on their willingness to support its deployment. 

\begin{table}[b]
\caption{Distribution of practitioners and researchers in the treatment and control groups.}
\label{table:background}
\centering
\begin{tabular}{lcc}
\toprule
\textbf{Background} & \textbf{Practitioners} & \textbf{Researchers} \\ 
\midrule
Treatment           & 52\%                   & 48\%                 \\
Control             & 58\%                   & 42\%                 \\ 
\bottomrule
\end{tabular}
\end{table}

% !TEX root = ../thesis-main.tex

\section{Discussion}
\label{section:6}
Since our original motivation was to provide an explanation system that can be used by forecasting analysts, we conducted a more in-depth analysis of the results to determine if there was a difference in attitudes between users depending on their background (e.g., practitioners from \OurCompany{} or researchers from the university).

\subsection{Comparing Attitudes Conditioned on Background}
\label{section:conditioning}
Table~\ref{table:background} shows the distribution of practitioners and researchers in the treatment and control groups. 
Since we have a slight imbalance in background between the treatment and control groups, we test whether or not our results still hold when conditioning on background and confirm that they do.

Again, we do not find statistically significant differences in initial attitudes towards the model ($\chi^{2}$ test, $\alpha = 0.05$). 
For researchers, the distribution of answers between treatment and control groups is significantly different for SQ1 ($\chi^{2} = 14.2$, $\alpha = 0.001$), but does not differ for SQ2, SQ3, or SQ4 ($\chi^{2}$ test, $\alpha = 0.05$). 
The same holds for practitioners: the distributions are significantly different only for SQ1 ($\chi^{2} = 6.94$, $\alpha = 0.05$). 
This is consistent with our results in Section~\ref{section:5}. 
In both cases, users in the treatment group agree with SQ1 more than users in the control group, indicating that \OurMethod{} explanations help users understand why the model makes large errors in predictions, regardless of whether they are practitioners or researchers.  
Although the results are statistically significant for both groups, it should be noted that the results hold more strongly for researchers compared to those for practitioners, given the $\chi^{2}$ values. 

\begin{figure}[t]
 \centering
 \includegraphics[clip,trim=0mm 0mm 0mm 0mm,scale=0.35]{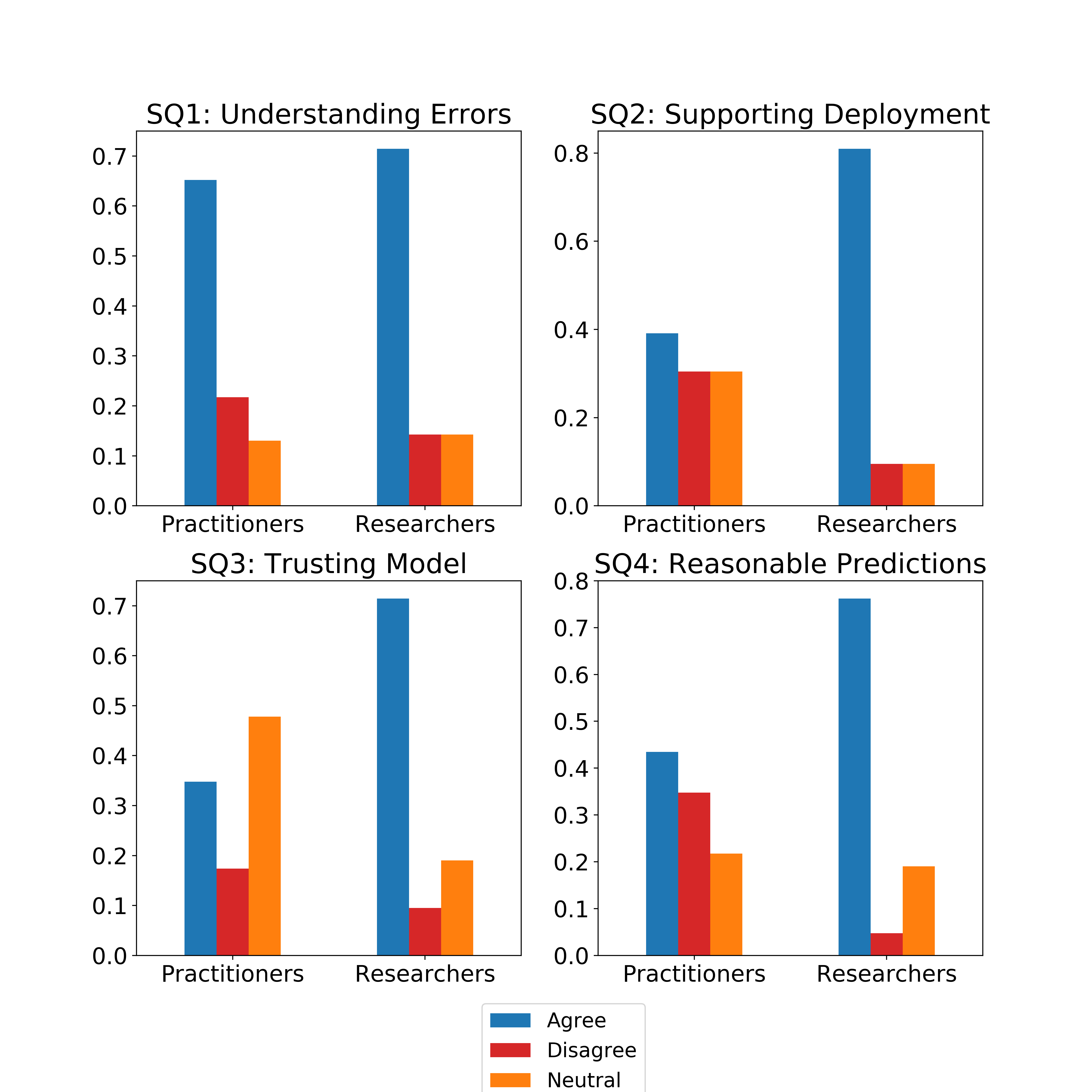}
  \caption{Results from a within-subject study comparing answers between participants who are practitioners or researchers (in the treatment group).}
 \label{fig:aca-vs-ind}
\end{figure}

\subsection{Comparing Attitudes in the Treatment Group}
Based on the users who saw the explanations, we compare the distributions of answers between practitioners and researchers in Figure~\ref{fig:aca-vs-ind} in order to understand the needs of different types of users. 
We find that there is a significant difference between practitioners and researchers for SQ2 ($\chi^{2} = 7.94$, $\alpha = 0.05$), indicating that more researchers are in favor of using the model as a forecasting tool, and less are against it or have a neutral attitude, in comparison to the practitioners. 
We also find a significant difference for SQ3 ($\chi^{2} = 5.98$, $\alpha = 0.05$): a larger proportion of researchers trust the model, while the majority of practitioners have neutral feelings. 
The results for SQ4 are significant as well ($\chi^{2} = 6.86$, $\alpha = 0.05$): 
although the majority of users in both groups believe the model produces reasonable predictions, a larger proportion of the practitioners disagree with this statement in comparison to the researchers. 

We see no significant difference between groups for SQ1 ($\chi^{2}$ test, $\alpha = 0.05$), which makes sense given that we showed that \OurMethod{} explanations have a similar effect on both practitioners and researchers when comparing users in the treatment and control groups in Section~\ref{section:conditioning}. 

Overall, these results suggest that our user study population is fairly heterogeneous, and that users from different backgrounds have different criteria for deploying or trusting a model, and varying levels of confidence regarding the accuracy of its outcomes.

\subsection{User Study Limitations}
Like any user study, ours has some limitations. 
It would have been preferable to distribute users more evenly in terms of the proportion of users in the treatment and control groups, as well as the proportion of practitioners and researchers in each of these groups. 
Unfortunately, this was not possible in our case because we recruited participants in two rounds: first for the treatment group, and then afterwards for the control group. 
One option could be to discard some practitioners in the control group in order to have a better balance in terms of background, but we felt it was more important to have as many users as possible, and it would not be clear how to choose which users to discard. 
Fortunately, we found that our results still hold when conditioning on background as mentioned in Section~\ref{section:conditioning}. 
In future work, we plan to recruit for both groups at the same time to avoid issues like these. 

We also acknowledge that not having a baseline method to compare to is a limitation of our study. 
In our case, the main issue is that there simply does not exist a method that is specifically for explaining errors in regression predictions, which would make asking questions about errors 
\begin{inparaenum}[(i)]
	\item unfair, and 
	\item confusing, as mentioned in Sections~\ref{section:lime} and~\ref{section:studydesign}. 
\end{inparaenum}
However, now that \OurMethod{}  exists, it can serve as a baseline for future work on erroneous predictions, which is another contribution of this paper.

% !TEX root = ../thesis-main.tex

\section{Conclusion}
\label{section:7}
In this chapter, we have proposed a method, MC-BRP, that provides users with contrastive explanations about predictions resulting in large errors based on:
\begin{inparaenum}[(i)]
\item the set of bounds for which reasonable predictions would be expected for each of the most important features. 
\item the trend between each of these features and the target.
\end{inparaenum}
%We evaluate \OurMethod{} explanations through a user study with objective and subjective components. 

\noindent%
Given a large error, \OurMethod{} generates a set of perturbed versions of the original instance that result in reasonable predictions. 
This is done by performing Monte Carlo simulations on each of the features deemed most important for the original prediction. 
For each of these features, we determine the bounds needed for a reasonable prediction based on the mean and standard deviation of this new set of reasonable predictions. 
We also determine the relationship between each feature and the target through the Pearson correlation, and present these to the user as the explanation. 

We evaluate \OurMethod{} both objectively (\textbf{RQ3.1}) and subjectively (\textbf{RQ3.2}) by conducting a user study with \numprint{75} real users, including both researchers and practitioners. 
We answer \textbf{RQ3.1} by conducting two types of simulations to quantify how 
\begin{inparaenum}[(i)]
\item interpretable, and 
\item actionable 
\end{inparaenum}
our explanations are. 
Through forward simulations, we show that users are able to interpret \OurMethod{} explanations by simulating the model's output with an average accuracy of 84.5\%. 
Through counterfactual simulations, 
%where users are asked to manipulate the model's input in order to change the output, 
we show that \OurMethod{} explanations are actionable with an accuracy of 76.2\%. 

We answer \textbf{RQ3.2} by conducting a between-subject experiment with subjective questions. 
The treatment group sees \OurMethod{} explanations, while the control group does not see any explanation. 
We find that explanations generated by \OurMethod{} help users understand why models make large errors in predictions (SQ1), but do not have a significant impact on support in deploying the model (SQ2), trust in the model (SQ3), or perceptions of the model's performance (SQ4). 
These results still hold when conditioning on users' background (practitioners vs. researchers). 
We also conduct an analysis on the treatment group to compare results between practitioners and researchers. 
We find significant differences for SQ2, SQ3 and SQ4, but do not find a significant difference in attitudes for SQ1. 

The answer to \textbf{\ref{rq:mcbrp}} is yes: we can create an explanation method based on a real-world use case by first identifying a use case where explanations are required and identifying what users want in the context of this use case. 
We can evaluate in a user-centric manner by conducting a user study based on the use case, which includes both objective and subjective components. 

So far, this thesis has focused on creating knowledge about responsible AI practices, specifically on developing new methods for behavior-based explanations (see Chapter~\ref{chapter:introduction}), where we explain predictions from ML models. 
In the next and final part of the thesis, we will focus more on process-based explanations, where we will investigate how to \emph{translate} knowledge about responsible AI practices to the next generation of AI researchers.

\section*{Reproducibility}
To facilitate the reproducibility of the work in this chapter, our code is available at \url{https://github.com/a-lucic/mcbrp}.

\part{Pedagogy}
% !TEX root = ../thesis-main.tex

\chapter{Teaching Responsible AI through Reproducibility}
\label{chapter:research-pedagogy}

\footnote[]{This chapter was published at the AAAI Symposium on Educational Advances in Artificial Intelligence (EAAI-AAAI 2022) under the title ``Reproducibility as a Mechanism for Teaching Fairness, Accountability, Confidentiality, and Transparency in Artificial Intelligence'' \citep{lucic2022reproducibility}.}
\acresetall

In this final part of the thesis, we investigate how to communicate responsible artificial intelligence (AI) practices to the next generation of AI researchers. 
In this chapter, we address the following research question:

\medskip
\noindent
\textbf{\ref{rq:pedagogy}:} \acl{rq:pedagogy}
\medskip

\noindent
We answer \textbf{\ref{rq:pedagogy}} by designing a course that is centered on a reproducibility project, where students work in teams to reimplement existing responsible AI algorithms from top AI conferences and reproduce the experiments reported in the papers. 
We report on our findings as we taught the course over two academic years and share recommendations for implementing similar courses in the future.

%!TEX root = ../main.tex

\section{Introduction}
\label{section:fact-introduction}

For several decades, \OurUniversity{} has offered a research-oriented Master of Science (MSc) program in AI. 
The main focus of the program is on the technical machine learning (ML) aspects of the major sub-fields of AI, such as computer vision, information retrieval, natural language processing, and reinforcement learning.
One of the most recent additions to the MSc AI curriculum is a mandatory course called \emph{\ac{FACT-AI}}. 
This course was first taught during the 2019--2020 academic year and focuses on teaching FACT-AI topics through the lens of reproducibility. 
The main project involves students working in groups to re-implement existing FACT-AI algorithms from papers in top AI venues. 
There are approximately 150 students enrolled in the course each year. 

The motivation for the course came from the MSc AI students themselves, who often play an important role in shaping the curriculum in order to meet the evolving requirements of researchers in both academia and industry. 
As the influence of AI on decision making is becoming increasingly prevalent in day-to-day life, there is a growing consensus that stakeholders who take part in the design or implementation of AI algorithms should reflect on the ethical ramifications of their work, including developers and researchers \citep{campbell2021_responsible}. 
This is especially true in situations where data-driven AI systems affect some demographic sub-groups differently than others \citep{propublica, o2017ivory}.
As a result, our students have shown an increased interest in the ethical issues surrounding AI systems and requested that the university put together a new course focusing on responsible AI.

Since our MSc AI program is characterized by a strong emphasis on understanding, developing, and building AI algorithms, we believe that a new course on responsible AI in this program should also have a hands-on approach.
The course is designed to address technical aspects of key areas in responsible AI:
\begin{enumerate*}[label=(\roman*)]
\item fairness, 
\item accountability, 
\item confidentiality, and 
\item transparency,  
\end{enumerate*}
which we operationalize through a reproducibility project. 
We believe a strong emphasis on reproducibility is important from both an educational point of view and from the point of view of the AI community, since the (lack of) reproducible results has become a major point of critique in AI \citep{hutson2018artificial}. 
Moreover, the starting point of almost any junior AI researcher (and most AI research projects in general) is re-implementing existing methods as baselines. 
The FACT-AI course is situated at a point in the program where students have learned the basics of ML and are ready to start experimenting with, and building on top of, state-of-the-art algorithms. 
Given that our MSc AI program is fairly research-oriented, it is important for students to experience the process of reproducing work done by others (and how difficult this is) at an early stage in their careers. 
We also believe reproducibility is a fundamental component of FACT-AI: the cornerstone of fair, accountable, confidential and transparent AI systems is having correct and reproducible results. 
Without reproducibility, it is unclear how to judge if a decision-making algorithm adheres to any of the FACT principles. 

In the 2019--2020 academic year, we operationalized our learning ambitions regarding reproducibility by publishing a public repository with selected code implementations and corresponding reports from the group projects.
In the 2020--2021 academic year, we took the projects one step further and encouraged students to submit to the ML Reproducibility Challenge,\footnote{\url{https://paperswithcode.com/rc2020}} a competition that solicits reproducibility reports for papers published in conferences such as NeurIPS, ICML, ICLR, ACL, EMNLP, CVPR and ECCV. 
Although the challenge broadly focuses on all papers submitted to these conferences, we focus exclusively on papers covering FACT-AI topics in our course. 
Submitting to the challenge gives students a chance to experience the whole AI research pipeline, from running experiments, to writing rebuttals, to receiving the official notifications. 
Of the 23 papers that were accepted to the ML Reproducibility Challenge in 2021, 9 came from groups in the FACT-AI course. 

In this chapter, we describe the FACT-AI course at \OurUniversity{}: a one month, full-time course based on examining ethical issues in AI using reproducibility as a pedagogical tool. 
Students work in groups to re-implement (and possibly extend) existing algorithms from top AI venues on \ac{FACT-AI} topics. 
The course also includes lectures that cover the high-level principles of FACT-AI topics, as well as paper discussion sessions where students read and digest prominent FACT-AI papers. 
In this chapter, we outline the setup for the \ac{FACT-AI} course and the experiences we had while running the course during the 2019--2020 and 2020--2021 academic years at \OurUniversity{}. 

The remainder of this chapter is structured as follows. 
In Section~\ref{section:fact-related-work}, we discuss related work, specifically other courses about responsible AI. 
In Section~\ref{section:reproducibility}, we detail ongoing reproducibility efforts in the AI community. 
In Section~\ref{section:learning-objectives}, we explain the learning objectives for our course, and explain how we realized those objectives in Section~\ref{section:coursesetup}. 
We reflect on the feedback we received about the course in Section~\ref{section:feedback}, as well as what worked (Section~\ref{section:whatworked}) and what did not (Section~\ref{section:whatdidnt}), before concluding in Section~\ref{section:fact-conclusion}. 

%!TEX root = ../main.tex

\section{Related Work}
\label{section:fact-related-work}

There have been multiple calls for introducing ethics in computer science courses in general, and in AI programs in particular \citep{leonelli2016locating, o2017ivory, singer2018tech, skirpan2018ethics, grosz2019embedded, salz2019integrating, danyluk2021computing}. 
Several surveys have investigated how existing responsible computing courses are organized \citep{peck2017,fiesler2020we, garret2020morethan, raji_you_2021}. 

\subsection{Characterizing Responsible AI Courses}
There are two primary approaches to integrating such components into the curriculum: 
\begin{enumerate*}[label=(\roman*)]
\item stand-alone courses that focus on ethical issues such as FACT-AI topics, and 
\item a holistic curriculum where ethical issues are introduced and tackled in each course. 
\end{enumerate*}
In general, the latter is rare \citep{peck2017, salz2019integrating, fiesler2020we}, and can be difficult to organize due to a lack of qualified faculty or relevant expertise \citep{bates2020integrating, raji_you_2021}. 
We opt for the first approach since our course is a new addition to an existing study program. 
%Since we were designing one new course to be added into an existing program, we opted for the first approach. 

\citet{fiesler2020we} analyze 202 courses on ``tech ethics''. 
Their survey examines 
\begin{enumerate*}[label=(\roman*)]
\item the departments the courses are taught from, as well as the home departments of the course instructors, 
\item the topics covered in the courses, organized into 15 categories, and 
\item the learning outcomes in the courses. 
\end{enumerate*}
In our case, both the FACT-AI course and its instructors are from the Informatics Institute of the Faculty of Science at \OurUniversity{}. Our learning objectives (see Section~\ref{section:learning-objectives}) correspond to the following learning objectives from \citet{fiesler2020we}: \textit{``Critique''}, \textit{``Spot Issues''}, and \textit{``Create solutions''}. 
According to their content topic categorization, our course includes \textit{``AI \& Algorithms''} and \textit{``Research Ethics''}: the former since the course deals explicitly with AI algorithms under the FACT-AI umbrella, and the latter due to its focus on reproducibility. 

We note that \textit{``AI \& Algorithms''} is only the fifth-most popular topic according to the survey, after \textit{``Law \& Policy''}, \textit{``Privacy \& Surveillance''}, \textit{``Philosophy''}, and \textit{``Inequality, Justice \& Human Rights''} \citep[see Table 2,][]{fiesler2020we}). 
Although we believe these topics are important, we also wanted to avoid the feeling that the course was a ``distraction from the real material'' \citep{lewis2021teaching}, especially since 
\begin{enumerate*}[label=(\roman*)]
\item the majority of our students are coming from a technical background into a technical MSc program, and 
\item the FACT-AI course is mandatory for all students in the MSc AI program. 
\end{enumerate*}

\subsection{Similar Responsible AI Courses}
The two courses that are the most similar to ours are those of \citet{lewis2021teaching} and \citet{yildiz2021reproducedpapers}. 

\citet{lewis2021teaching} describe a course for responsible data science. 
Similar to our course, they focus on the technical aspects of AI, involving lectures, readings, and a final project. 
However, their course differs from ours since the main project in their course is focused on examining the interpretability of an automated decision making system, while the main project in our course is focused on reproducibility. 

\citet{yildiz2021reproducedpapers} describe a course based on reproducing experiments from AI papers, focusing on ``low-barrier'' reproducibility. 
Similar to our course, this course involves replicating a paper from scratch or reproducing the experiments using existing code, performing hyperparameter sweeps, and testing with new data or with variant algorithms. 
Another similarity is that they released a public repository of re-implemented algorithms,\footnote{\url{https://reproducedpapers.org/}} which we also did for the 2019--2020 iteration of our course (see Section~\ref{section:assignment}). 
However, their course differs from ours since theirs focuses on AI papers in general, while our course focuses exclusively on FACT-AI papers. 

There are several courses that focus more on the philosophical or social science perspectives of AI ethics.
\citet{green2021aiethics} describes an undergraduate AI ethics course that teaches computer science majors to analyse issues using different ethical approaches and how to incorporate these into an \textit{explicit} ethical agent. 
\citet{shenvalue2021} introduce a toolkit in the form of ``Value Cards'' to inform students and practitioners about the social impacts of ML models through deliberation.
\citet{green2020argument} propose an approach to ethics education using ``argument schemes'' that summarize key ethical considerations for specialized domains such as healthcare or national defense.
\citet{Furey2018IntroducingET} introduce ethics concepts, primarily utilitarianism, into an existing AI course about autonomous vehicles by studying several variations of the Trolley Problem.  
\citet{burton2018} teach ethics through science fiction stories complemented with philosophy papers, allowing students to reflect and debate difficult content without emotional or personal investment since the stories are not tied to ``real'' issues. 
\citet{skirpan2018ethics} describe an undergraduate course on human-centred computing which integrates ethical thinking throughout the design of computational systems. 
Unlike these courses, our course focuses more on the technical aspects of ethical AI.  
However, incorporating such non-technical perspectives is something we would like to do in future iterations of our course, perhaps through one of the mechanisms employed by some of these courses. 

\section{Reproducibility in ML Research}
\label{section:reproducibility}

There have been several criticisms about the lack of reproducibility in AI research. 
Some have postulated that this is due to a combination of unpublished code and high sensitivity of training parameters \citep{hutson2018artificial}, while others believe the rapid rate of progress in ML research results in a lack of empirical rigor \citep{sculley2018winner}. 
Although typically well-intentioned, some papers may disguise speculation as explanation, obfuscate content behind math or language, and fail to attribute the correct sources of empirical gains \citep{lipton2019research}. 

Several efforts have been made to investigate and increase the reproducibility of AI research. 
In 2021, NeurIPS introduced a paper checklist including questions about reproducibility, along with a template for submitting source code as supplementary material \citep{neuripschecklist}. 
The Association of Computing Machinery introduced a badging system that indicates how reproducible a paper is \citep{acm-artifact}. 
Papers with Code is an organization that provides links to official code repositories and datasets in arXiv papers \citep{paperswithcode2020}. 
It also hosts an annual ML Reproducibility Challenge: a community-wide effort to investigate the reproducibility of papers accepted at top AI conferences, which we incorporated into the 2020--2021 iteration of the FACT-AI course. 

In an ideal scenario, reproducibility issues would be handled prior to publication \citep{sculley2018winner}, 
but it can be difficult to catch such shortcomings in the review process due to the increasing number of papers submitted to AI conferences. 
Therefore, we believe it is of utmost importance that the next generation of AI researchers -- including our own students -- can 
\begin{enumerate*}[label=(\roman*)]
\item identify and 
\item avoid these pitfalls while conducting their own work. 
\end{enumerate*}
This, in combination with the fact that reproducibility is a fundamental component of responsible AI research, is why we opted to teach the FACT-AI course through the lens of reproducibility. 

Our course is centered around a group project where students re-implement a recent FACT-AI algorithm from a top AI conference. 
This project has three components:
\begin{enumerate*}[label=(\roman*)]
\item a reproducibility report, 
\item an associated code base, and 
\item a group presentation. 
\end{enumerate*}
In Section~\ref{section:assignment}, we provide more details on the project and the outputs it resulted in. 

%!TEX root = ../main.tex

\section{Learning Objectives}
\label{section:learning-objectives}

In the FACT-AI course, we aim to make students aware of two types of responsibility: (i) towards society in terms of potential implications of their research, and (ii) towards the research community in terms of producing reproducible research. 
In this section, we outline the learning objectives for the FACT-AI course and explain how it fits within the context of the MSc AI program at \OurUniversity{}. 

\begin{table}[t]
\caption{The first year of the MSc AI program at \OurUniversity{}, 2019--2020.}
\label{tab:msc_program}
\centering
\setlength{\tabcolsep}{3pt}
\begin{tabular}{l c@{~}c@{~}c c@{~}c@{~}c c}
\toprule
Course & \multicolumn{3}{c}{Sem.\ 1} & \multicolumn{3}{c}{Sem.\ 2} & EC \\
\midrule
     Computer Vision 1 & $\blacksquare$ & $\square$ & $\square$ & $\square$ & $\square$ & $\square$ & 6 \\
     Machine Learning 1 & $\blacksquare$ & $\square$ & $\square$ & $\square$ & $\square$ & $\square$ & 6 \\
     Natural Language Processing 1 & $\square$ & $\blacksquare$ & $\square$ & $\square$ & $\square$ & $\square$ & 6 \\
     Deep Learning 1 & $\square$ & $\blacksquare$ & $\square$ & $\square$ & $\square$ & $\square$ & 6\\
     Fairness, Accountability, Confidentiality  & $\square$ & $\square$ & $\blacksquare$ & $\square$ & $\square$ & $\square$ & 6 \\
     and Transparency in AI \\
     Information Retrieval 1 & $\square$ & $\square$ & $\square$ & $\blacksquare$ & $\square$ & $\square$ & 6 \\
     Knowledge Representation and Reasoning & $\square$ & $\square$ & $\square$ & $\blacksquare$ & $\square$ & $\square$ & 6 \\
     Elective 1 & $\square$ & $\square$ & $\square$ & $\square$ & $\blacksquare$ & $\square$ & 6 \\
     Elective 2 & $\square$ & $\square$ & $\square$ & $\square$ & $\blacksquare$ & $\square$ & 6 \\
     Elective 3 & $\square$ & $\square$ & $\square$ & $\square$ & $\square$ & $\blacksquare$ & 6 \\
\bottomrule
\end{tabular}
\end{table}

Table~\ref{tab:msc_program} shows the setup of the first year of the 2-year MSc AI program. Each semester at \OurUniversity{} is divided into three periods: two 8-week periods followed by one 4-week period. During an 8-week period, students follow two courses in parallel. During the 4-week period, they only follow a single course. 
The FACT-AI course takes place during the 4-week period at the end of the first semester, after students have taken Computer Vision 1, Machine Learning 1, Natural Language Processing 1 and Deep Learning 1. 
It is the only course students follow during this period, so we believe it is beneficial to have them focus on one main project -- reproducing an existing FACT-AI paper. 
The learning objectives for the course are as follows: 

\begin{itemize}
	\item \textbf{LO \#1: Understanding FACT topics.} Students can explain the major notions of fairness, accountability, confidentiality, and transparency that have been proposed in the literature, along with their strengths and weaknesses.
	\item \textbf{LO \#2: Understanding algorithmic harm.} Students can explain, motivate, and distinguish the main types of algorithmic harm, both in general and in terms of concrete examples where AI is being applied.
	\item \textbf{LO \#3: Familiarity with FACT methods.} Students are familiar with recent peer-reviewed algorithmic approaches to fairness, accountability, confidentiality, and transparency in the literature. 
    \item \textbf{LO \#4: Reproducing FACT solutions.} Students can assess the degree to which recent algorithmic solutions are effective, especially with respect to the claims made in the original papers, while understanding their limitations and shortcomings. 
\end{itemize}

%!TEX root = ../main.tex

\section{Course Setup}
\label{section:coursesetup}

% The course is located right after the winter break of the second and last year of the Master in AI at \OurUniversity{}. 
% Thus, students can fully focus on it before starting to write their Master Thesis.
The FACT-AI course is organized around 
\begin{enumerate*}[label=(\roman*)]
    \item lectures, 
    \item paper discussions, and
    \item a group project. 
\end{enumerate*}
It has had two iterations so far: the 2019--2020 iteration was taught in person, while the 2020--2021 iteration was taught online due to the COVID-19 pandemic. 
In this section, we detail how we realized the learning objectives from Section~\ref{section:learning-objectives} and describe the challenges in adapting the course to an online format.

\subsection{Lectures}
To further the understanding of FACT-AI topics (LO1), we provide one general lecture for each of the 4 topics, along with a lecture specifically about reproducibility.
Lectures are an opportunity for students to familiarize themselves with algorithmic harm (LO2). Students are encouraged to ask questions that lead to discussions about potential harm done by applications of AI. 
This was more challenging in the second iteration of the course due to the online format, but we hope that facilitating such discussions will be more straightforward once we return to in-person classes. 

In addition to the general lectures, we also include some guest lectures. 
These are used to either discuss specific types of algorithmic harm (LO2), examine specific FACT-AI algorithms in depth (LO3), or expand on the non-technical aspects of FACT-AI. 
Some examples of guest lectures include a lecture on AI accountability from a legal perspective by an instructor from the law department of \OurUniversity{}, and a lecture by two former FACT-AI students who explained how they turned their group project into an ICML 2021 workshop paper \citep{neely2021order}.\footnote{This was later extended to a full paper at the International Conference on Hybrid Human-AI (HHAI 2022) \citep{neely2022song}.} 

\subsection{Paper Discussions}
The goal of the paper discussion sessions is for students to learn about prominent FACT-AI methods (LO3), and learn to think critically about the claims made in the papers we discuss (LO4). 
Students first read a seminal FACT-AI paper on their own while trying to answer the following questions: 
\begin{itemize}
\item What are the main claims of the paper?
\item What are the research questions?
\item Does the experimental setup make sense, given the research questions?
\item What are the answers to the research questions? Are these supported by experimental evidence?
\end{itemize}

\noindent%
Once students have read the papers, they participate in smaller discussion sessions with their peers about their answers to the questions above. 
After each discussion session, all the groups are brought back together for a ``dissection'' session, where an instructor goes over the same seminal paper, giving an overview of the papers' strengths and weaknesses. 
Each session was presented by a different instructor to show that there is no single way of examining a research paper, and that different researchers will bring different perspectives to their assessment of papers. 
The following papers were covered during the discussion sessions: 
\citet{hardt_equality_2016} on fairness; 
\citet{ribeiro-2016-should} on transparency; and 
\citet{Abadi_2016} on confidentiality.

\subsection{Group Project}
\label{section:assignment}

\subsubsection{Reproduction of a FACT-AI paper}
The purpose of the group project is to have students investigate the claims made by the authors of recent FACT-AI papers by diving into the details of the methods and their implementations.  
Using what they have learned from the paper discussion sessions, students work in groups to re-implement an existing FACT-AI algorithm from a top AI conference and re-run the experiments in the paper to determine the degree to which they are reproducible (LO4). 
If the code is already available, then they must extend the method in some way. 
The project consisted of three deliverables: 
\begin{enumerate*}[label=(\roman*)]
\item a reproducibility report, 
\item an associated code base, and 
\item a group presentation. 
\end{enumerate*}
In order to ensure the project is feasible, we select 10--15 papers in advance for groups to choose from. 
Our criteria for including papers is as follows:
\begin{itemize}
    \item The paper is on a FACT-AI topic. 
    \item At least one dataset in the paper is publicly available.
    \item Experiments can be run on a single GPU (which we provide access to).
    \item It is reasonable for a group of 3--4 MSc AI students to re-implement the paper within the timeframe of the course. In our case, students work on this project for one-month full-time. 
\end{itemize}

\noindent%
To ease the load for our teaching assistants (TAs), we have several groups working on the same paper. 
We assign papers to TAs based on their interests by asking them to rank the set of candidate papers in advance. 
We also encourage them to suggest alternative papers provided they fit the criteria. 
The TAs read the papers before the course starts in order to ensure they have a sufficient, in-depth understanding of the work such that they can guide students through the project. 
This also serves as an extra feasibility check, to ensure that the papers are indeed a good fit for our course.

Each group writes a report about their efforts following the structure of a standard research paper (i.e., introduction, methodology, experiments, results, conclusion). 
They also include aspects specific to reproducibility such as explaining the difficulties of implementing certain components, as well as describing any communication they had with the original authors.
In addition to the source code, students provide all results in a Jupyter notebook along with a file to install the required environment. 

\subsubsection{First Iteration: Contributing to an Open Source Repository}
In the 2019--2020 iteration of the course, we created a public repository on GitHub, which contains a selection of the implementations done by our students: \url{https://github.com/uva-fact-ai-course/uva-fact-ai-course}. 
The TAs who assisted with the course decided which implementations to include and cleaned up the code so it all fit into one cohesive repository. 
This had multiple motivations.
First, it taught students how to improve the reproducibility of their own work by releasing the code, while also giving them a sense of contributing to the open-source community. 
Second, a public repository can serve as a starting point for personal development in their future careers; companies often ask to see existing code or projects that prospective employees have worked on. 
Some students added the project to their CVs, while others wrote blog posts about their experiences,\footnote{\url{https://omarelb.github.io/self-explaining-neural-networks/}}  linking back to the repository.

\subsubsection{Second Iteration: The ML Reproducibility Challenge}
In the 2020--2021 iteration of the course, we formally participated in the annual ML Reproducibility Challenge \citep{paperswithcode2020} in order to expose our students to the peer-review process. 
This gave students something to strive towards and offered perspectives beyond simply getting a grade for the project. 
Most importantly, it gave them the opportunity to experience the full research pipeline: 
\begin{enumerate*}[label=(\roman*)]
	\item reading a technical paper, 
	\item understanding a paper's strength and weaknesses, 
	\item implementing (and perhaps also extending) the paper, 
	\item writing up the findings, 
	\item submitting to a venue with a deadline, 
	\item obtaining feedback, 
	\item writing a rebuttal, and
	\item receiving the official notification. 
\end{enumerate*}
To encourage students to formally submit to the ML Reproducibility Challenge, we offered a 5\% boost to their final grades if they submitted. 
Of the 32 groups in the FACT-AI course, 30 (94\%) groups submitted their reproducibility reports to the ML Reproducibility Challenge, of which 9 groups had their papers accepted.

\subsection{Online Course Format}
The second iteration of the course was taught in January 2021, when the COVID-19 pandemic forced us to move classes and interactions online. 
Students made use of various channels to communicate: WhatsApp, Discord, and Slack.  
Canvas was the primary mode of communication between the instructors and the students, allowing students to ask questions and instructors to communicate various announcements. 

Lectures were live, with frequent Q\&A breaks to stimulate interactivity. 
Paper discussion sessions were organized in different online meeting subrooms where students discussed the papers together.  
This proved to be a challenge: while some subrooms had productive discussions, others struggled to get the conversation going. 

The reproducibility project was more difficult to launch remotely. 
Since students had done online classes for their entire first semester, some struggled to find a group of fellow students to team up with, especially those coming from outside the MSc AI program. 
Overall, while we had various communication means, the lack of physical interaction due to COVID-19 slowed down our course organization. 

%!TEX root = ../main.tex

\section{Feedback}
\label{section:feedback}

In this section, we discuss the feedback we received about the course from the perspective of participating students (Section~\ref{section:feedback-students}) and from the ML Reproducibility Challenge reviews (Section~\ref{section:feedback-mlrc}). 
 
\subsection{Feedback from Students}
\label{section:feedback-students}
Both iterations of the course were evaluated using the standard evaluation procedure provided by \OurUniversity. 
However, only 16\% of students filled out the evaluation form (23 out of 144) in the 2020--2021 iteration, potentially because the evaluation forms were administered online instead of in-person. 
According to the evaluation procedure at our university, this is not enough for a reliable quantitative estimate of student satisfaction.
Therefore, we focus on the 2019--2020 iteration when reporting student satisfaction statistics, since 53\% of students filled out the form (79 out of 149) that year. 
The vast majority of students were (very) satisfied with the course overall (67.8\%). 
More specifically, students expressed satisfaction with the following dimensions: 
\begin{itemize}
\setlength\itemsep{0.1em}
    \item Academic challenge: 75.2\% were (very) satisfied
    \item Supervision: 76.9\% were (very) satisfied
    \item Feedback: 81.3\% were (very) satisfied
    \item Workload: 91.3\% were (very) satisfied
    \item Level of the course: 79.7\% were (very) satisfied
    \item Level of the report: 94.8\% were (very) satisfied
    \item Level of the presentation: 96.6\% were (very) satisfied
\end{itemize}

\begin{table}[htp]
\caption{Feedback about the course.}
\centering
\begin{tabular}{@{}l@{}}
\toprule
(a) Feedback from students \\
\midrule
\begin{minipage}[t]{\columnwidth}
\begin{itemize}[leftmargin=*]
    \item ``Reproducing an article was hard and intensive but a really good experience.''
     \item ``Replicating another study, seeing how (poorly) other research is performed was really eye-opening.''
    \item ``Reproducing a paper: I believe this is a good thing to do and is an important part of academia.''
    \item  ``Gave good insights into the trustworthiness of research papers, which is apparently not great.''
    \item ``I appreciate the critical view I have developed on papers as a result of this course. Normally I would easily accept the content of a paper, but I will be more critical from now on, as many papers are not reproducible.''
    \item ``I think it's really good that we get some practical insights into reproducing results from other papers, not all papers are as good as they seem to be.''
    \item ``I really appreciated that this was the first course where students are judging state-of-the-art AI-models. In other words, students were able to experience the scientific workfield of AI.''
\end{itemize}
\end{minipage}
\\ \\
\midrule
(b) Feedback from the ML Reproducibility Challenge \\
\midrule
\begin{minipage}[t]{\columnwidth}
\begin{itemize}[leftmargin=*]
\item ``The report reveals a lot of dark spots of the original paper.''
\item ``Good reviews, strong reproducibility report, provides code reimplementation from scratch which is a strong contribution.''
\item ``The discussion section is a great reference point for future work.''
\item ``The additional experimentation is rather impressive and the report reflects  an intuitive understanding of concepts such as  coverage, correctness, and counterfactual explanations.''
\item ``The report provides good insights on how the experiments in the original paper actually work, while also generating new hypothesis to be tested for future research, which is a positive outcome.''
\item ``My main concern is that it remains unclear why some of the results are so far off from the original paper? I would have expected the authors to dig deeper on that.''
%\item ``The paper is generally difficult to follow. The paper reads closer to an outline than a finished report.''
\item ``It doesn't go above and beyond the reproduction and does not offer novel insights into the workings of the original paper.''
\item ``The submission failed at reproducing the original results. It is unclear whether this is due to a difference in the experimental setup or due to implementation errors. ''
%\item ``The paper reads closer to an outline than a finished report. I would encourage the authors to spend some additional time on organization, making sure that the key takeaways are made plain and that the report reads fluidly throughout.''
\end{itemize}
\end{minipage}
\end{tabular}
\label{tab:feedback}
\end{table}

Table~\ref{tab:feedback}(a) shows some of the qualitative feedback we received from students. 
Based on this, we believe these high scores are mostly the result of the reproducibility project. 
Students enjoyed doing the project, especially due to the intensive supervision from our experienced TAs. 
The dimensions where we received the lowest scores were on the lectures and the final presentation, where only 30.6\% and 30.2\% were (very) satisfied with these aspects, respectively. 
This may be because we only provided four (high-level) lectures on each of the four topics, in order to give students as much time as possible to focus on the reproducibility project. 
However, it should be noted that the overall scores for these components were not poor, but average: 3.1/5 for lectures and 3.0/5 for the presentation. 

\subsection{Feedback from the ML Reproducibility Challenge}
\label{section:feedback-mlrc}

Of the 30 reproducibility reports submitted to the ML Reproducibility Challenge in the 2020--2021 iteration, 9 were accepted for publication in the ReScience Journal.  
In total, the ML Reproducibility Challenge accepted 23 reports, meaning that almost 40\% of the reports accepted to the challenge were from \OurUniversity{}.\footnote{\url{https://openreview.net/group?id=ML_Reproducibility_Challenge/2020}}
 
The reviews were mostly positive, with the general consensus being that most teams had gone beyond the general expectation of simply re-implementing the algorithm and re-running the experiments. 
Our TAs encouraged students to examine the generalizability of the work that was reproduced, either by trying new datasets or hyperparameters, or by performing ablation studies. 
Multiple reproducibility reports managed to question the results of the original papers with experimentally-supported claims. 
Importantly, some reviewers emphasized that these reproducibility studies were solid starting points for future research projects. 
For the reports that were rejected, the main critiques were that
\begin{enumerate*}[label=(\roman*)]
    \item only a fraction of the original work was reproduced, or 
    \item no new insights were given.
\end{enumerate*}
Some projects also had flaws in the experimental setup. 
See Table~\ref{tab:feedback}(b) for quotes from the ML Reproducibility Challenge reviews.

%!TEX root = ../main.tex

\section{Factors Contributing to a Successful Course}
\label{section:whatworked}

Understanding and re-implementing the work of other researchers is not a trivial task, especially for first-year MSc students. There were several aspects of the setup that we believe were beneficial for the students, which we organize along three dimensions: 
\begin{enumerate*}[label=(\roman*)]
    \item general, 
    \item concerning FACT-AI, and 
    \item concerning reproducibility.
\end{enumerate*}
We believe each of these factors is important for a successful implementation of this course, or other similar courses. 

\subsection{General}

\subsubsection{Timing of the course}
It is important that students have prior knowledge of ML theory as well as some programming experience before completing a project-based course in groups. At \OurUniversity{}, the FACT-AI course takes place after students have completed 4 ML-focused courses (see Table~\ref{tab:msc_program}). 
We believe it is extremely important that students have access to adequate preparation, especially in terms of programming experience, before setting off to reproduce experiments from prominent AI papers. 
Without this prior knowledge, we believe such a project would not be feasible in the allotted time frame. 

\subsubsection{Regular contact with experienced TAs}
The TAs are there to help with two main components: 
\begin{enumerate*}[label=(\roman*)]
\item understanding the paper, and 
\item debugging the implementation process. 
\end{enumerate*}
In practice, we found that it is extremely important for the TAs to have excellent programming experience since this is the main aspect students need help with. 
%We found it even more beneficial if the TAs had taken the course themselves previously, so we plan to recruit future TAs from the pool of students that completed the course in the previous year. 
We also had a dedicated Slack workspace for the TAs and course instructors to keep in touch regularly. 

Since our course is only four weeks long, we found it was important for students to have regular contact with their TAs to ensure no one got stuck in the process. 
For the first (in-person) iteration of the course, groups had one-hour tutorials with their TAs twice per week, where all groups that were working on the same paper (and therefore had the same TA) were in the same tutorial. 
Since they were all working on the same paper, there were many overlapping questions, and students found it beneficial to be able to share their experiences with one another. 
For the second (online) iteration of the course, we thought it would be challenging to ensure each group got the attention they needed if everyone was in one large online tutorial, so the TAs met with each group separately for 30 minutes, twice per week. 

\subsubsection{Early feedback on the reports}
Approximately halfway through the course, we asked students to submit a draft report to their TAs in order to get feedback. We found this significantly increased the quality of the final reports. 

\subsection{Concerning FACT-AI}

\subsubsection{Emphasizing the technical perspective of FACT-AI}
Given that the FACT-AI course is situated in the context of a technical, research-oriented MSc, having students re-implement research papers from top AI conferences was an effective way to teach FACT-AI topics for our students. 
Teaching FACT-AI from a primarily technical perspective 
aligns well with what students expect from the MSc AI program at \OurUniversity{}.
Although we believe a technical focus makes sense for our MSc program, we also believe it is important to incorporate non-technical perspectives into the course -- see Section~\ref{section:non-technical}. 

\subsubsection{Creating resources for the FACT-AI community}
We believe a significant motivating factor for students was creating concrete output that extended beyond simply completing a project for a course: creating resources for the FACT community. 
In the 2019--2020 iteration, this was done by creating a public repository with the best FACT-AI algorithm implementations, as selected by the TAs. 
In the 2020--2021 iteration, this was done by publicly submitting their reproducibility reports about FACT-AI algorithms to the ML Reproducibility Challenge, where the accepted reports were published in the ReScience Journal. 
In the future, we plan to continue aligning our course with the ML Reproducibility Challenge since we found the process extremely beneficial for our students. 

\subsection{Concerning Reproducibility}

\subsubsection{Including extension as part of reproducibility}
If source code was already available for the paper -- which is fortunately becoming increasingly common for AI research papers -- we asked students to think about how to extend the paper since the implementation was already available. This resulted in some creative and interesting ideas in the reports, and we believe this is why our students performed well at the ML Reproducibility Challenge.

\subsubsection{Simple grading setup}
For a 4-week, project-based course, we found it was beneficial for students to focus on one main deliverable consisting of three components: 
\begin{enumerate*}[label=(\roman*)]
\item the reproducibility report, 
\item the associated code base, and
\item the group presentation. 
\end{enumerate*}
The report that students submitted for the course was the same one they submitted to the ML Reproducibility Challenge. This way, participating in the ML Reproducibility Challenge was not an extra task but rather an integral part of the course.

\section{Areas of Improvement}
\label{section:whatdidnt}

Although we believe both iterations of the course went well, there are several aspects of the setup that we believe could use some improvement and other instructors should consider if they plan to implement a similar course. 

\subsection{General}

Given that this is the first time most students are formally submitting a paper, it is not surprising that there were some logistical issues. 
Some groups made minor mistakes such as forgetting to submit their work double-blind or slightly missing the submission deadline. 
We also had some groups who wrote the introduction sections of their papers as an introduction to the FACT-AI course, rather than an introduction to the topic they were working on. 
In future iterations, we will explicitly state the standard procedures of writing and submitting a paper and provide some examples. 

\subsection{Concerning FACT-AI}
\label{section:non-technical}

Although focusing primarily on the technical aspects of FACT-AI is an effective way to engage our technical students in socially-relevant AI problems, we also believe that they would benefit from additional non-technical perspectives on FACT-AI topics. 
In the future, we plan to include perspective from outside of computer science through 
\begin{enumerate*}[label=(\roman*)]
\item additional guest lectures, 
\item workshop sessions \citep{skirpan2018ethics,shenvalue2021}, and 
\item broader impact statements \citep{campbell2021_responsible} in the reproducibility reports. 
\end{enumerate*}

\subsection{Concerning Reproducibility}
In future iterations, we believe it would be useful to show students more examples of what a high-quality reproducibility paper looks like and explain in-depth why it is high-quality. 
These could be papers that were previously accepted to the ML Reproducibility Challenge, or papers from other reproducibility efforts outlined in Section~\ref{section:reproducibility}. 
We want the students to understand what makes a paper a good (reproducibility) paper, that is, it has a set of (reproducibility) claims, it argues for these claims, and shows evidence to support these claims.  
%!TEX root = ../main.tex

\section{Conclusion}
\label{section:fact-conclusion}

In this chapter, we share our setup for the FACT-AI course at \OurUniversity{}, which teaches FACT-AI topics through reproducibility. 
The course set out to give students 
\begin{enumerate*}[label=(\roman*)]
    \item an understanding of FACT-AI topics, 
    \item an understanding of algorithmic harm, 
    \item familiarity with recent FACT-AI methods, and
    \item an opportunity to reproduce FACT-AI solutions, 
\end{enumerate*}
through a combination of lectures, paper discussion sessions and a reproducibility project. 
Through their projects, our students engaged with the open-source community by creating a public code repository (in the 2019--2020 iteration), as well as with the research community via successful submissions to the ML Reproducibility Challenge (in the 2020--2021 iteration). 
We also detail how the 2020--2021 iteration brought about its own unique set of challenges due to the COVID-19 pandemic. 

In this course, we illustrate that reproducibility should be viewed as a fundamental component of FACT-AI. 
We received very positive feedback on teaching FACT-AI topics through reproducibility. We believe this was an excellent fit for our students, which not only helped motivate them for the duration of the course, but also helped them develop skills that will be essential in their future research careers, whether in the private or public sector.

With this final chapter, we answer \textbf{\ref{rq:pedagogy}}: we can use reproducibility as a mechanism for teaching responsible AI concepts to a technical, research-oriented audience. 
Structuring the course around a reproducibility project gives students the opportunity to learn about responsible AI concepts, such as explainability, in a hands-on manner. 
Since the publication of the paper on which this chapter is based \citep{lucic2022reproducibility}, we ran another iteration of the FACT-AI course in 2021--2022 under the same setup as the previous year, where students submitted their reports to the 2022 edition of the ML Reproducibility Challenge. 
21 of the 43 papers accepted to the ML Reproducibility Challenge in 2022 were from students in the FACT-AI course. 
These also included some awards: the Best Paper Award was awarded to a group from the FACT-AI course as well as 2 of the 4 Outstanding Paper Awards. 
We believe this indicates that our course setup can serve as a starting point for effective participation in the broader ML research community.

%\fi

\bookmarksetup{startatroot} % Lift from parts
%\addtocontents{toc}{\bigskip} % skip
%\addtocontents{toc}{\vspace{-50ex}}

% !TEX root = ../thesis-main.tex

\chapter{Conclusions}
\label{chapter:conclusions}

\acresetall
In this thesis, we have investigated explainability in ML from three viewpoints: 
\begin{inparaenum}[(i)]
	\item algorithms, 
	\item users, and
	\item pedagogy. 
\end{inparaenum}
In this final chapter of the thesis, we revisit the research questions from Chapter~\ref{chapter:introduction}, state our main findings in Section~\ref{section:conclusion-findings}, and identify directions for future work in Section~\ref{section:conclusion-futurework}.

\section{Main Findings}
\label{section:conclusion-findings}
In this section, we describe our main findings across the three parts of the thesis. 

\subsection{Algorithms}

The first part of this thesis focused on investigating behavior-based explanations in order to explain individual predictions from specific types of ML models. 
In Chapter~\ref{chapter:research-focus}, we asked our first research question:

\begin{description}\item[\acs{rq:focus}]\acl{rq:focus}\end{description}

\noindent
The answer to \textbf{RQ1} is yes: we are able to explain predictions for tree ensembles in a counterfactual manner by including differentiable approximations of tree-based models within a standard gradient-based optimization framework. 
In the majority of experimental settings, our method outperforms existing baselines in terms of
\begin{inparaenum}[(i)]
	\item the number of counterfactual examples produced, 
	\item the average distance between the counterfactual examples and the original examples, 
	and
	\item the proportion of counterfactual examples that are closer to the original examples. 
\end{inparaenum}
Our method is flexible since it can produce different types of counterfactual explanations depending on which distance function we choose to include in the loss function. 
In practice, this allows the user to customize the explanations depending on the use case. 

We then turned to our next research question:

\begin{description}\item[\acs{rq:cf-gnn}]\acl{rq:cf-gnn}\end{description}

\noindent
The answer to \textbf{RQ2} is yes: we can adapt our method from \textbf{RQ1} to the graph setting by introducing a binary perturbation matrix that is multiplied element-wise with the adjacency matrix in order to remove edges from the graph. 
Since this was one of the first methods for generating counterfactual explanations for GNNs, we also had to design a corresponding experimental setup to evaluate it. 
In the majority of experimental settings, we found that our method outperformed the baselines in terms of
\begin{inparaenum}[(i)]
	\item accuracy, 
	\item the number of counterfactual examples produced, 
	\item the number of edges removed, and
	\item the proportion of the subgraph neighborhood that was perturbed.
\end{inparaenum}

\subsection{Users}
In the second part of this thesis, we continued our work on behavior-based explanations, but shifted our predominantly algorithmic focus to also account for the users who consume the explanations. 
We investigated the following research question:

\begin{description}\item[\acs{rq:mcbrp}]\acl{rq:mcbrp}\end{description}

\noindent
The answer to \textbf{RQ3} is yes: we developed an explanation method based on the needs of real-world analysts in order to help them understand large errors in sales forecasting predictions. 
We designed a user study to evaluate our method and found that for the vast majority of users, our explanations were both interpretable and actionable. 
We also found that most users believed the explanations helped them understand large errors in predictions, but they did not have an impact on other aspects such as trust or confidence in the model.

\subsection{Pedagogy}
In the third part of this thesis, we transitioned from creating knowledge about ML model predictions to communicating knowledge about responsible ML practices. 
Process-based explanations play an important role here, specifically in the context of documentation practices which help ensure research is conducted in a responsible and reproducible manner. 
We asked our final research question:

\begin{description}\item[\acs{rq:pedagogy}]\acl{rq:pedagogy}\end{description}

\noindent
We answered \textbf{RQ4} by developing a course that was centered on a reproducibility project, where students worked in groups to reimplement algorithms from top-AI conferences on responsible AI topics. 
We shared our experiences with teaching the course over two academic years and suggested best practices for implementing similar reproducibility courses in the future. 

\section{Future Directions}
\label{section:conclusion-futurework}
In this section, we describe some limitations of the methods proposed in this thesis and identify potential avenues for future work.

\subsubsection{Limitations of counterfactual explanations}
Researchers have raised concerns about the hidden assumptions behind the use of counterfactual examples~\citep{barocas_hidden_2019}, as well as potentials for misuse~\citep{kasirzadeh2021use}. 
When explaining ML models using counterfactual examples, it is important to account for the context in which the systems are deployed. 
Counterfactual explanations are not a guarantee to achieving recourse~\citep{ustun_actionable_2019} -- changes suggested should be seen as candidate changes, not absolute solutions, since what is pragmatically actionable differs depending on the end user and context. 
While existing research from the cognitive sciences has shown that humans are able to interpret counterfactual explanations, the notion of what constitutes a \emph{minimal} perturbation is not clear \citep{byrne-2016-counterfactual}. 
Further research into the interpretability and cognitive efficacy of counterfactual explanations could help the field better understand the appropriate criteria to optimize for.

\subsubsection{Accommodating different types of perturbations}
In Chapter~\ref{chapter:research-cfgnn}, we proposed a method for generating counterfactual explanations for GNNs.  
In its current form, our method is limited to performing edge deletions for node classification tasks. 
Given that many graph datasets also include node features, future work should involve incorporating node feature perturbations in our framework. 
We could also extend our method to accommodate graph classification tasks. 
%We also plan to investigate adapting graph attack methods for generating counterfactual explanations, 

\subsubsection{Including additional criteria in loss functions}
In Chapter~\ref{chapter:research-focus}, we proposed a method for generating flexible counterfactual explanations for tree-based models using gradient optimization techniques. 
The flexibility comes from varying the distance function used in the loss function, which results in different types of counterfactual explanations depending on which distance function is chosen. 
Future work could involve trying alternative distance functions or including additional criteria in the loss function, such as proximity to other points in the dataset or stability of the counterfactual example. 
This could also be applied to the method proposed in Chapter~\ref{chapter:research-cfgnn}.

\subsubsection{Evaluating with users}
Although the counterfactual explanations proposed in Chapters~\ref{chapter:research-focus} and~\ref{chapter:research-cfgnn} perform well on various distance metrics, we should conduct a user study to evaluate how useful they are in practice. 
We could build on our existing user study design from Chapter~\ref{chapter:research-mcbrp} to test how
\begin{inparaenum}[(i)]
	\item varying the distance functions, and 
	\item introducing new components into the loss functions
\end{inparaenum}
impacts user preferences for counterfactual explanations.

\pagebreak
%\subsubsection*{Abstaining from prediction}
\subsubsection{Improving trust in explanations}
In Chapter~\ref{chapter:research-mcbrp}, we proposed a method for explaining errors in forecasting predictions based on identifying unusual feature values.
We find that although explanations from our method help users understand large errors in predictions, they do not have an impact on users' trust, deployment support, or perception of the model's performance. 
Future work could place more emphasis on trying to improve these aspects, for example by allowing a predictive model to abstain from prediction when a particular instance has unusual feature values beyond a certain threshold. 
%We also plan to compile a more comprehensive set of subjective questions by using multiple questions to evaluate users' impressions on the same topic.

\subsubsection{Developing robust protocols for XAI evaluation}
In general, we believe it is crucial for the ML community to invest in developing more rigorous evaluation protocols for XAI methods, both in terms of user studies as well as formal metrics. 
The XAI community could pursue collaborations with researchers from human-computer interaction to design human-centered user studies about evaluating the utility of XAI methods in practice. 
To design metrics, the XAI community could try borrowing ideas from information theory or collaborating with ML evaluation researchers in order to ensure that the explanations we generate are truly representative of the model's behavior.

%\subsubsection{Promoting standardized protocols for reproducibility.}
\subsubsection{Teaching reproducibility as a fundamental component of ML research}
Reproducibility mechanisms such as checklists and challenges can help promote reproducible research practices, but we do not believe they alone are enough to cause a shift in the ML community. 
We believe the key to fostering reproducible research starts in the classroom. 
It is important to teach the next generation of ML researchers that reproducibility is not an afterthought, but rather a fundamental component of conducting ML research responsibly. 
In addition to conducting reproducibility projects, we could also introduce reproducibility components in programming assignments across all courses within an ML study program.

\subsubsection{Identifying consistent terms for explainability}
Due to growing collection of XAI literature, there are many definitions for various distinct but related concepts such as explainability, interpretability, transparency, and intelligibility. 
As a community, we should make an effort to standardize the terms we use in order to facilitate easier communication, especially with researchers from non-ML disciplines. 
Developing XAI that is useful in practice requires interdisciplinary collaboration, which is more straightforward if we can all speak the same language. 
This could be achieved through a workshop-style event with researchers working on XAI to consolidate a standardized terminology. 

\pagebreak

\subsubsection{Final thoughts} 
Overall, our main advice for future work is to continue prioritizing explainability in the ML community, whether it is \emph{behavior-based} or \emph{process-based}. 
For both types of explanations, we should explore developing explainability techniques that cater to different types of users with varying levels of granularity, as well as robust mechanisms for evaluation. 
As a community, we need to prioritize both correctness and interpretability of explanations -- incorrect explanations that are interpretable do not provide the user with any concrete information, and correct explanations that are uninterpretable are not useful to the user. 
To promote correctness, we need to first identify what it means for an explanation to be ``correct'' and create datasets that allow us to explore this task explicitly. 
To promote interpretability, we need to approach the explainability problem from an interdisciplinary perspective, and suggest that XAI researchers spend more time connecting to the communities they are designing the explanations for.

%\fi

% Start the back matter
% Numbering stays the same
% Add 1) bibliography, 2) abstract (Dutch/English)
\backmatter
% !TEX root = thesis-main.tex

% Include list of notation
%\chapter*{List of Notation}
%\addcontentsline{toc}{chapter}{List of Notation}
%\markboth{List of Notation}{List of Notation}

% Include the bibliography

% Do some formatting of the chapter title and page headers
% Set the item separation to 0 (saves a few pages)
\renewcommand{\bibsection}{\chapter*{Bibliography}}
\renewcommand{\bibname}{Bibliography}
\markboth{Bibliography}{Bibliography}
\renewcommand{\bibfont}{\footnotesize}
\setlength{\bibsep}{0pt}

% Include the actual bib file
% Add the chapter to the ToC (it doesn't happen automatically if you loose the chapter number)
\bibliographystyle{plainnat}
\bibliography{thesis}

% Include abstract(s)

% Add at least a Dutch one, the English one could also go on the back cover
% Again, add the chapter to the ToC
\chapter*{Summary}
Model explainability has become an important problem in artificial intelligence (AI) due to the increased effect that algorithmic predictions have on humans. 
Explanations can help users understand not only why AI models make certain predictions, but also how these predictions can be changed. 
In the first part of this thesis, we investigate counterfactual explanations: given a data point and a trained model, we want to find the minimal perturbation to the input such that the prediction changes. 
We frame the problem of finding counterfactual explanations as a gradient-based optimization task and first focus on tree ensembles. 
We extend previous work that could only be applied to differentiable models by incorporating probabilistic model approximations in the optimization framework, and find that our counterfactual examples are significantly closer to the original instances than those produced by other methods specifically designed for tree-based models. 

We then extend our method for generating counterfactual explanations for tree ensembles to accommodate graph neural networks (GNNs), given the increasing promise of GNNs in real-world applications such as fake news detection and molecular simulation. 
We do so by introducing a perturbation matrix that acts on the adjacency matrix in order to iteratively remove edges from the graph, and find that our method primarily removes edges that are crucial for the original predictions, resulting in minimal counterfactual explanations. 

In the second part of this thesis, we investigate explanations in the context of a real-world use case: sales forecasting. 
We propose an algorithm that generates explanations for large errors in forecasting predictions based on Monte Carlo simulations. 
To evaluate, we conduct a user study with 75 users and find that
%by conducting a user study to determine if our explanations help users understand why large prediction errors occur and if this promotes trust in an AI model. 
the majority of users are able to accurately answer objective questions about the model's predictions when provided with our explanations, and that users who saw our explanations understand why the model makes large errors in predictions significantly more than users in the control group. 
We also conduct an in-depth analysis of the difference in attitudes between practitioners and researchers, and confirm that our results hold when conditioning on the users' background. 

In the final part of the thesis, we explain the setup for a technical, graduate-level course on responsible AI topics at \OurUniversity{}, which teaches responsible AI concepts through the lens of reproducibility. 
The focal point of the course is a group project based on reproducing existing responsible AI algorithms from top AI conferences and writing a corresponding report. 
We reflect on our experiences teaching the course over two years and propose guidelines for incorporating reproducibility in graduate-level AI study programs.

\chapter*{Samenvatting}
\markboth{Samenvatting}{Samenvatting}

De uitlegbaarheid van voorspellende modellen is een belangrijk probleem geworden in kunstmatige intelligentie (KI) vanwege het toegenomen effect dat algoritmische voorspellingen hebben op mensen.
Een uitleg kan gebruikers niet alleen helpen om te begrijpen waarom KI-modellen bepaalde voorspellingen doen, maar ook hoe deze voorspellingen be\"{i}nvloed kunnen worden. 
In het eerste deel van dit proefschrift onderzoeken we contrafeitelijke verklaringen: we willen, gegeven een datapunt en een getraind model, de minimale verandering van de input vinden die de voorspelling verandert. 
We formuleren het probleem van het vinden van contrafeitelijke verklaring als een op gradi\"enten gebaseerde optimalisatietaak en richten ons eerst op \textit{tree ensembles}. 
We bouwen voort op eerder werk dat alleen kon worden toegepast op differentieerbare modellen, door probabilistische modelbenaderingen op te nemen in het optimalisatiekader, en komen tot de bevinding dat onze contrafeitelijke voorbeelden significant dichter bij het oorspronkelijke datapunt liggen dan de voorbeelden die geproduceerd worden door andere methoden, die specifiek zijn ontworpen voor modellen die op bomen zijn gebaseerd.
 
Vervolgens breiden we onze methode voor het genereren van contrafeitelijke verklaringen uit voor \emph{tree ensembles} zodat die ook werkt voor \textit{graph neural networks} (GNNs), gezien de toenemende belofte van GNNs voor toepassingen in de echte wereld, zoals de detectie van nepnieuws en moleculaire simulatie.
We bereiken dit door een perturbatiematrix te introduceren die de elementen uit de bogenmatrix vermenigvuldigt om iteratief zijden van de graaf te verwijderen, en komen tot de bevinding dat onze methode voornamelijk zijden verwijdert die cruciaal zijn voor de oorspronkelijke voorspelling, wat resulteert in een minimale contrafeitelijke verklaring.
 
In het tweede deel van dit proefschrift onderzoeken we verklaringen in de context van een praktijkvoorbeeld: verkoopprognoses. We introduceren een algoritme dat verklaringen genereert voor grote fouten bij het doen van regressievoorspellingen op basis van Monte Carlo-simulaties. We evalueren door middel van een gebruikersonderzoek met 75 deelnemers. We komen tot de bevinding dat de meerderheid van de gebruikers in staat is accuraat antwoord te geven op meerkeuzevragen over de voorspellingen van het model wanneer zij voorzien zijn van onze uitleg. Daarnaast komen we tot de bevinding dat gebruikers die onze uitleg zagen, significant vaker dan gebruikers uit de controlegroep begrijpen waarom het model grote fouten maakt in voorspellingen. We voeren ook een analyse uit van het verschil in attitudes tussen praktijkbeoefenaars en onderzoekers, en bevestigen dat onze resultaten stand houden gegeven de achtergrond van de gebruiker. 
 
In het laatste deel van dit proefschrift behandelen we de opzet van een technisch vak op masterniveau over verantwoord gebruik van KI, gegeven aan de Universiteit van Amsterdam. In dit vak worden verantwoorde KI-concepten vanuit het oogpunt van reproduceerbaarheid gedoceerd. Het speerpunt van de cursus is een groepsproject dat gebaseerd is op het reproduceren van bestaande verantwoorde KI-algoritmen van vooraanstaande KI-conferenties en het schrijven van een bijbehorend rapport. We reflecteren op onze ervaringen met het geven van de cursus gedurende twee jaar en stellen richtlijnen voor voor het opnemen van reproduceerbaarheid in KI-studieprogramma's op masterniveau.

%% Close
\end{document}